\documentclass[final,5p,times,twocolumn]{elsarticle}

\usepackage{amssymb}
\usepackage{amsthm}

\usepackage{amsmath}
\usepackage{amsfonts}
\usepackage{amssymb}
\usepackage{epsfig}
\usepackage{epstopdf}
\usepackage{subfig}
\usepackage{xcolor}
\usepackage{caption}
\usepackage{grffile}
\usepackage{algorithmic}
\usepackage{algorithm}
\usepackage{xcolor}
\usepackage[pdftex]{hyperref}

\theoremstyle{plain}
\newtheorem{theorem}{Theorem}
\theoremstyle{plain}
\newtheorem*{theorem*}{Theorem}
\theoremstyle{plain}
\newtheorem{corollary}{Corollary}
\theoremstyle{plain}
\newtheorem*{corollary*}{Corollary}
\newenvironment{proofsketch}{%
  \proof}{\endproof}
\theoremstyle{definition}
\newtheorem{defn}{Definition}
\allowdisplaybreaks

\newcommand{\myalgo}{DANTE~}

\newcommand\blfootnote[1]{%
  \begingroup
  \renewcommand\thefootnote{}\footnote{#1}%
  \addtocounter{footnote}{-1}%
  \endgroup
}

\journal{Elsevier Neural Networks}

\begin{document}

\begin{frontmatter}


\title{DANTE: Deep AlterNations for Training nEural networks}

\title{}

\author[add_iith]{Vaibhav~B~Sinha\corref{cor1}\fnref{equal,ut}}
\ead{cs15btech11034@iith.ac.in}
\author[add_iith]{Sneha Kudugunta\fnref{equal}}
\ead{snehakudugunta@google.com}
\fntext[equal]{Authors contributed equally}
\fntext[ut]{Currently at Department of Computer Science, University of Texas at Austin, US}
\cortext[cor1]{Corresponding author}

\author[add_iith]{Adepu Ravi Sankar}
\ead{cs14resch11001@iith.ac.in}

\author[add_iith]{Surya Teja Chavali}
\ead{chavali2@wisc.edu}

\author[add_iitk]{Purushottam Kar}
\ead{purushot@cse.iitk.ac.in}

\author[add_iith]{Vineeth N Balasubramanian}
\ead{vineethnb@iith.ac.in}

\address[add_iith]{Department of Computer Science and  Engineering, Indian Institute of Technology Hyderabad, India}
\address[add_iitk]{Department of Computer  Science  and  Engineering, Indian Institute of Technology Kanpur, India}

\author{}

\address{}

\begin{abstract}
We present DANTE, a novel method for training neural networks using the alternating minimization principle. DANTE provides an alternate perspective to traditional gradient-based backpropagation techniques commonly used to train deep networks. It utilizes an adaptation of quasi-convexity to cast training a neural network as a bi-quasi-convex optimization problem. We show that for neural network configurations with both differentiable (e.g. sigmoid) and non-differentiable (e.g. ReLU) activation functions, we can perform the alternations effectively in this formulation. DANTE can also be extended to networks with multiple hidden layers. In experiments on standard datasets, neural networks trained using the proposed method were found to be  promising and competitive to traditional backpropagation techniques, both in terms of quality of the solution, as well as training speed.
\end{abstract}




\end{frontmatter}



\section{Introduction}
\label{sec:introduction}

For much of the recent march of deep learning, gradient-based backpropagation methods, e.g. Stochastic Gradient Descent (SGD) and its variants, have been the mainstay of practitioners. The use of these methods, especially on vast amounts of data, has led to unprecedented progress in several areas of artificial intelligence in recent years. The intense focus on these techniques has led to an intimate understanding of hardware requirements and code optimizations needed to execute these routines on large datasets in a scalable manner. Today, myriad off-the-shelf and highly optimized packages exist that can churn reasonably large datasets on GPU architectures with relatively mild human involvement and little bootstrap effort.\blfootnote{\textcopyright 2020. This manuscript version is made available under the CC-BY-NC-ND 4.0 license \url{http://creativecommons.org/licenses/by-nc-nd/4.0/}\\DOI: \url{https://doi.org/10.1016/j.neunet.2020.07.026}}

However, this surge of success of backpropagation-based methods in recent years has somewhat overshadowed the need to continue to look for options beyond backpropagation to train deep networks. Despite several advancements in deep learning with respect to novel architectures such as encoder-decoder networks, generative adversarial models and transformer networks, the reliance on backpropagation methods remains. 
Several works have studied the limitations of SGD-based backpropagation, whether it be vanishing gradients, especially for certain activation functions \cite{HochreiterS1997}; the tendency of SGD to face difficulties with saddle points \cite{dauphin2014identifying} - even for simple architectures \cite{tian2016symmetry}; or even more subtle issues such as significant difference in training time for networks having same expressive power as seen in \cite{pmlr-v70-shalev-shwartz17a}. Importantly, while existing backpropagation-based methods work, it is essential to continuously look for alternative methods that can help train neural networks effectively.

Complementarily, there has been marked progress in recent years in the broader area of non-convex optimization. Several alternate algorithms with provable guarantees, such as iterative hard thresholding \cite{BlumensathD2009}, alternating minimization \cite{JainNS2013}, \cite{AnandkumarG2016} and \cite{malach2018provably}. 
In this work, we leverage recent developments in optimization (quasi-convex, to be precise) to propose a non-backpropagation strategy to train neural networks. Our method is called \myalgo (Deep AlterNations for Training nEural networks), and it offers an alternating minimization-based technique for training neural networks. There have been a few related efforts of late, which we review in Section \ref{sec:background}.

\myalgo is based on the simple but useful observation that the problem of training a single hidden-layer neural network can be cast as a bi-quasiconvex optimization problem (described in Section \ref{formulation}). This observation allows us to use an alternating optimization strategy to train the neural network, where each step involves solving relatively simpler quasi-convex problems. \myalgo then uses efficient solvers for quasi-convex problems such as stochastic normalized gradient descent \cite{sngd} to train the neural network using alternating minimization. The key original contributions of this work can be summarized as:

\vspace{-5pt}
\begin{itemize}
\setlength\itemsep{0em}
	\item We show that the error in each layer of a neural network can, in fact, be viewed as a quasi-convex function, thus allowing us to treat a single hidden-layer neural network as a bi-quasi-convex optimization problem. Motivated by recent work \cite{sngd}, this allows us to propose an alternating minimization strategy, \myalgo, where each quasi-convex optimization problem can be solved effectively (using Stochastic Normalized Gradient Descent (SNGD)).
    \item While earlier results on the effectiveness of SNGD for solving a quasi-convex problem was restricted to a simple sigmoid Generalized Linear Model (GLM) with the squared loss, we show that SNGD can converge in high probability to an $\epsilon$-suboptimal solution even in case of layers of a neural network. We also expand the scope to include Rectified Linear Units (ReLU) activation functions and its variants by introducing a Generalized ReLU activation function. We also provide theoretical results in case of networks using Cross Entropy as loss (all of which has not been done before).
    \item We show \myalgo can be extended to train deep neural networks with multiple hidden layers.
	\item We empirically validate \myalgo with both the generalized ReLU and sigmoid activations and establish that \myalgo provides competitive or better performance on several standard datasets, when compared to standard mini-batch SGD-based backpropagation.
	\item While the high level idea of using the definition to prove the quasi-convexity of functions is inspired by \cite{sngd}, we would like to highlight that our proofs are more involved, and use techniques that were not in \cite{sngd} (to adapt to newer activation functions, handle hidden layers, as well as multiple output neurons). 
\end{itemize}
\vspace{-5pt}
We now review earlier related efforts, before presenting details of the proposed methodology.

\section{Related Work}
\label{sec:background}

Backpropagation-based techniques date back to the early days of neural network research \cite{rumelhart1986learning,chauvin1995backpropagation} but remain to this day, the most commonly used methods for training a variety of neural networks including multi-layer perceptrons, convolutional neural networks, autoencoders, recurrent networks and the like. 

In recent years, Taylor \textit{et al.} \cite{taylor2016training} and Choromanska \textit{et al.} \cite{pmlr-v97-choromanska19a} proposed methods to train neural networks which belong to the broad framework of alternating-minimization. Although both of these approaches use alternating-minimization, they are fundamentally different from ours. Both of them use auxiliary variables and the minimization is done on these auxiliary variables too, while our algorithm does not have any auxiliary variables and only minimizes on the weights of the network, giving it significant advantages in space complexity and training-time. Moreover, Taylor's algorithm does not use gradients unlike ours. Jaderberg proposed the idea of `synthetic gradients' in \cite{jaderberg2016decoupled}. Although interesting, this work is more focused towards a more efficient way to carry out gradient-based parameter updates in a neural network. More recently, Jagatap and Hegde \cite{jagpratap2018relu} proposed a method to train single hidden layer ReLU networks using an alternating minimization technique. Unlike our method, this method alternates between updating weights, and state variables which indicate which ReLU activations are on, and so is very specific to ReLU activations. 

In this work, we focus on an entirely new approach to training neural networks using alternating optimization inspired by quasi-convexity (different from the abovementioned methods), and show that this approach shows promising results to train neural networks of different depths on a range of datasets. Although alternating minimization has found much appeal in areas such as matrix factorization (\cite{JainNS2013}), to the best of our knowledge, this is the one of the early efforts in using alternating principles to train feedforward neural networks effectively.

Other efforts that are related to this work include target propagation based methods, such as in \cite{bengio2014auto}, Difference Target Propagation \cite{lee2015difference} and target propagation in a Bayesian setting \cite{bengio2015towards}. There are also efforts that use random feedback weights such as feedback-alignment \cite{lillicrap2014random} and direct/indirect feedback-alignment\cite{nokland2016direct} where the weights used for propagation need not be symmetric with the weights used for forward propagation. We however do not focus on credit assignment in this work. One could view the proposed method however as carrying out `implicit' credit assignment using partial derivatives, but there is no defined model for credit assignment which is not the focus of this work. We now describe our methodology.
\vspace{-3mm}

\section{Deep AlterNations for Training nEural networks (\myalgo)}
\label{sec:altmin}
We begin by presenting the overall problem formulation.
\subsection{Problem Formulation}
\label{formulation}
Consider a neural network with $L$ layers. Each layer $l \in \{1, 2,\ldots,L\}$ has $n_l$ nodes and is characterized by a linear operator $W_l \in \mathbb R^{n_{l-1} \times n_l}$ and a non-linear activation function defined as $\phi_l: \mathbb{R}^{n_l} \rightarrow \mathbb{R}^{n_l}$. The activations generated by layer $l$ are denoted by $\mathbf{a}_l \in \mathbb R^{n_l}$. We denote by $\mathbf{a}_0$, the input activations and $n_0$ to be the number of input activations i.e. $\mathbf{a}_0 \in \mathbb R^{n_0}$. Each layer uses activations being fed into it to compute its own activations as $\mathbf{a}_l = \phi_l\langle W_l,\mathbf{a}_{l-1} \rangle \in \mathbb{R}^{n_l}$, where $\phi \langle .,. \rangle$ denotes $\phi (\langle .,. \rangle)$ for simplicity of notation. A multi-layer neural network is formed by nesting such layers to form a composite function $f$ given as follows:
\vspace{-3pt}
\begin{equation}
\label{eq_mlnn}
f(\mathbf{W};\mathbf{x}) = \phi_L\langle W_L,\phi_{L-1}\langle W_{L-1},\cdots, \phi_1 \langle W_1,\mathbf{x} \rangle \rangle \rangle
\end{equation}
\noindent where $\mathbf W = \{ W_l\}$ is the collection of all the weights through the network, and $\mathbf{x} = \mathbf{a}_0$ contains the input activations for each training sample. 

Given $m$ data samples $\left\{ \left(\mathbf{\mathbf{x}_{i}},y_{i}\right)\right\} _{i=1}^{m}$ from a distribution $\cal D$ and a loss function $J$, the network is trained by tuning the weights $\mathbf{W}$ to minimize the empirical risk associated with:
\vspace{-5pt}
\begin{equation}
\label{eq_mlnn_loss}
\min_{\mathbf{W}}\ \mathbb E_{(\mathbf x, y) \sim \cal D}[J(f(\mathbf W;\mathbf{x}), y)]
\end{equation}
For purpose of simplicity and convenience, we first consider the case of a single hidden layer neural network, represented as $f(\mathbf{W};\mathbf{x}) = \phi_2 \langle W_2,\phi_{1} \langle W_{1}, \mathbf{x} \rangle \rangle$ to describe our methodology. We later describe how this can be extended to  multi-layer neural networks.
A common loss function used to train neural networks is the squared loss function which yields the following objective (we later also study changing the loss function in this work):
\vspace{-5pt}
\begin{equation}
\label{eq:form}
\min_{\mathbf{W}} \ \mathbb E_{(\mathbf x, \mathbf{y}) \sim \cal D} \|f(\mathbf W;\mathbf{x}) - \mathbf y\|_2^2
\end{equation}
\noindent where:
\vspace{-5pt}
\begin{equation}
\label{eq:form_2}
\|f(\mathbf W;\mathbf{x}) - \mathbf y\|_2^2 = \|\phi_2 \langle W_2,\phi_{1} \langle W_{1}, \mathbf{x} \rangle \rangle - \mathbf y\|_2^2
\end{equation}
An important observation here is that if we fix $W_1$, then Eq. \eqref{eq:form_2} turns into a set of Generalized Linear Model (GLM) problems with $\phi_2$ as the activation function, i.e.
\vspace{-5pt}
\begin{equation}
\label{eqn_W2_problem}
\min_{W_2} \ \mathbb E_{(\mathbf x, \mathbf{y}) \sim \cal D} \|\phi_2 \langle W_2, \mathbf z \rangle - \mathbf y\|_2^2
\end{equation}
where $\mathbf z = \phi_{1} \langle W_{1}, \mathbf{x} \rangle$.  In particular, a GLM with differentiable activation functions such as sigmoid 
satisfy a property called \emph{Strict Locally Quasi-Convexity} (SLQC), which allows techniques such as SNGD to solve the GLM problem effectively, as pointed out in \cite{sngd}. We exploit this observation, and generalize this result in many ways: (i) we firstly show that a GLM with non-differentiable activation functions such as ReLUs (and its variants) also satisfy the SLQC property; (ii) we show that a set of GLMs, such as in a layer of a neural network, also satisfy the SLQC property; and (iii) we leverage these generalizations to develop an alternating minimization methodology to train neural networks; and (iv) we also move away from the restriction of square loss and provide extensions for Cross-Entropy Loss. Sections \ref{subsec_rationale} and \ref{subsec_rationale_extension} describe these generalizations further. 

Optimizing for $W_1$ in Equation \ref{eq:form_2}, unfortunately, cannot be directly viewed as a set of SLQC GLMs. To this end, we provide a generalization of local quasi-convexity in Section \ref{subsec_rationale} and show that fixing $W_2$ does indeed turn the problem below into yet another SLQC problem, this time with $W_1$ as the parameter (note that $\phi_{W_2} \langle \cdot \rangle = \phi_2 \langle W_2,\phi_{1} \langle \cdot \rangle \rangle$):
\vspace{-5pt}
\begin{equation}
\label{eqn_W1_problem}
\min_{W_1} \ \mathbb E_{(\mathbf x, \mathbf{y}) \sim \cal D} \|\phi_{W_2} \langle W_{1}, \mathbf{x} \rangle - \mathbf y\|_2^2
\end{equation}
Putting Equations \ref{eqn_W2_problem} and \ref{eqn_W1_problem} together gives us a single-layer neural network setup, where each layer is individually SLQC, and can be efficiently solved using SNGD. This allows us to propose our alternating minimization strategy to train a neural network in an effective manner. We now describe in detail each of the above steps, beginning with the background and preliminaries required to set up this problem. 
\vspace{-5pt}

\subsection{Background and Preliminaries}
\label{subsec_rationale}

Let $\| \cdot \|$  denote the $L_2$ (Euclidean) norm for vectors, and $\| \cdot \|_F$ denote the Frobenius norm of a matrix. We sometimes drop the subscript $F$ from $\| \cdot \|_F$ for ease of reading (the appropriate norm can be identified from the context). $\mathbb{B}(\mathbf{x},r)$ denotes a Euclidean ball of radius $r$ with $\mathbf{x}$ as center and $\mathbb B$ denotes $\mathbb{B}(0,1)$. We begin with the formal definitions of Local Quasi-Convexity and Generalized Linear Model (GLM).
\vspace{-5pt}
\begin{defn}[\textbf{\textit{Local-Quasi-Convexity}} \cite{sngd}]
\label{defn:lqc}
Let $\mathbf{x},\mathbf{z}\in \mathbb R^{d},\kappa,\epsilon>0$ and $f: \mathbb{R}^{d}\to \mathbb{R}$ be a differentiable function. Then $f$ is said to be $(\epsilon,\kappa,\mathbf z)$-Strictly-Locally-Quasi-Convex (SLQC) in $\mathbf{x}$, if at least one of the following applies: 
\vspace{-5pt}
\begin{enumerate}
\setlength\itemsep{0em}
\item $f(\mathbf{x})-f(\mathbf{z})\le\epsilon$
\item $\| \nabla f(\mathbf{x})\|>0$, and \textrm{$\forall \mathbf{y} \in \mathbb{B}\left(\mathbf{z},\epsilon/\kappa\right)$, }$\langle \nabla f(\mathbf{x}), \mathbf{y}-\mathbf{x}\rangle\le0$ 
\end{enumerate}
\end{defn}
\vspace{-8pt}
\noindent where $\mathbb{B}\left(\mathbf{z},\epsilon/\kappa\right)$ is a ball centered at $\mathbf{z}$ with radius $\epsilon/\kappa$.
\vspace{-5pt}
\begin{defn}[\textbf{\textit{Idealized and Noisy Generalized Linear Model}} \cite{sngd}]
\label{defn:nglm}
In the idealized GLM setting, we are given $m$ samples $\left\{ \left(\mathbf{\mathbf{x}_{i}},y_{i}\right)\right\} _{i=1}^{m} \in \mathbb{B} \times [0,1]$  and an activation function $\phi:\mathbb{R}\to \mathbb{R}$. There exists $\mathbf{w^\ast} \in \mathbb{R}^d$  such that   $y_i = \phi \langle \mathbf{w}^\ast, \mathbf{x}_i \rangle \forall i \in \{1,\cdots,m\}$ where $\mathbf{w^\ast}$ is the global minimizer of the empirical error function:
\vspace{-5pt}
\[ \hat{err}(\mathbf{w})= \frac{1}{m} \sum_{i=1}^m \left(y_i-\phi(\langle\mathbf{\mathbf{w},x_{i}\rangle)}\right)^{2} \]
\vspace{-1pt}
\noindent In the noisy GLM setting, we are given $m$ samples $\left\{ \left(\mathbf{\mathbf{x}_{i}},y_{i}\right)\right\} _{i=1}^{m} \in \mathbb{B}_d \times [0,1]$ drawn i.i.d. from an unknown distribution $\cal D$. There exists a $\mathbf{w^\ast} \in \mathbb{R}^d$ such that $\mathbb{E}_{(\mathbf{x},y) \sim \mathcal{D}}[y|\,\mathbf{x}] = \phi(\langle \mathbf{w}^*,\textbf{x}\rangle)$, and $\mathbf{w^\ast}$ is the global minimizer of :
\vspace{-5pt}
\[ err(\mathbf{w})=\mathbb{E}_{(\mathbf{x},y) \sim \mathcal{D}}\left(y-\phi(\langle\mathbf{\mathbf{w},x\rangle)}\right)^{2} \]
\end{defn}
\vspace{-5pt}
\noindent Hazan et al. showed in \cite{sngd} that the idealized GLM problem with the sigmoid activation function is $(\epsilon,e^{\|\mathbf w^\ast\|},\mathbf w^\ast)$-SLQC in $\mathbf{w}$, $\forall \mathbf{w} \in \mathbb{B}(0,\|\mathbf w^\ast\|)$ and $\forall \epsilon > 0$; and that if we draw $m \geq \Omega\left( \frac{\exp(2\|\mathbf w^\ast\|)}{\epsilon^2}\log\frac{1}{\delta}\right)$ i.i.d. samples from $\cal D$, the empirical error function $\hat{err}$ with sigmoid activation is $(\epsilon,e^{\|\mathbf w^\ast\|},\mathbf w^\ast)$-SLQC in $\mathbf w$ for any $\mathbf{w} \in \mathbb{B}(0,\|\mathbf w^\ast\|)$ with probability at least $1-\delta$. However, these results by themselves are not directly useful, considering they are proved only for a single GLM (which can be viewed as a neural network with no hidden layers and a single output neuron), and which are non-trivial to extend to a traditional multi-layer/feedforward neural network. Besides, their proofs rely on properties of the sigmoid function, which restricts us from using these (and any following) results to contemporary neural networks which use other activation functions such as the ReLU. We overcome all of these restrictions in this work, and provide a new mechanism to use such a theoretical result in practice.\\

\subsubsection{SLQC-ness of a GLM with Non-Linear Activations}
Before presenting our approach with multi-layer neural networks, we begin our description of the proposed methodology by showing that a GLM with a ReLU activation function is also SLQC (~\cite{sngd} already showed this for a GLM with sigmoid activation). This will later allow us to seamlessly extend our results to both sigmoid and ReLU multi-layer neural networks. (We note that $\tanh$ - which is simply a rescaled sigmoid - is also subsumed in these definitions.) To this end, we introduce a new \textit{generalized ReLU} activation function, defined as follows:
\vspace{-5pt}
\begin{defn}\textit{\textbf{(Generalized ReLU)}}
\label{defn_gen_relu}
The generalized ReLU function $f:\mathbb{R} \to \mathbb{R}$, $0 < a \le b $, $a, b \in \mathbb{R}$ is defined as:
\vspace{-10pt}
\begin{center}
    $
f(x) = \left\{
        \begin{array}{ll}
            ax & \quad x \leq 0 \\
            bx & \quad x > 0
        \end{array}
    \right.
$
\end{center}
\end{defn}
\vspace{-3pt}

Note that this definition subsumes variants of ReLU such as the Leaky ReLU \cite{maas2013rectifier} or PReLU \cite{he2015delving}. This function is differentiable at every point except 0. We define the function $g$ that provides a valid subgradient for the generalized ReLU at all $x$ to be: 
\vspace{-8pt}
\begin{center}
$
g(x) = \left\{
        \begin{array}{ll}
            a & \quad x < 0 \\
            b & \quad x \ge 0 
        \end{array}
    \right.
$
\end{center}

We now prove our first results using the above definition of the generalized ReLU in idealized and noisy GLMs below.
\begin{theorem}
In the idealized GLM with generalized ReLU activation, assuming $\|\mathbf{w^{*}}\|\le W$, 
$\hat{err}(\mathbf{w})$ is $\left(\epsilon,\frac{2b^{3}W}{a},\mathbf{w^{*}}\right)-SLQC$
in $\mathbf{w}, \forall \mathbf{w}\in \mathbb{B}(0,W)$ and $\forall \epsilon > 0$.
\label{theorem_iglm_relu}
\end{theorem}
\vspace{-13pt}
\begin{proofsketch} We use Definition \ref{defn:lqc} to show this result. Consider a point $\textbf{v}$, $\epsilon/\kappa$-close to minima $\textbf{w}^*$ with $\kappa = \frac{2b^{3}W}{a}$. Throughout the paper, we measure closeness in L2-norm. Let $G$ represent the subgradient of $\hat{err}_m(\textbf{w})$. We show that $\langle G(\mathbf{w}), \mathbf{w} - \mathbf{v} \rangle \ge 0$, which proves the result. To show this inequality, we exploit the Lipschitzness of ReLU function, the bound on its derivative and the fact that $\phi\langle\mathbf{w}^*, \mathbf{x}_i \rangle = y_i$. The complete proof is presented in Section \ref{appendix_idealglm} for ease of reading further at this time.
\end{proofsketch}

\begin{theorem}
\label{theorem_noisyglm_relu}
In the noisy GLM with generalized ReLU activation, assuming $\|\mathbf{w^{*}}\|\le W$, given  $\mathbf{w} \in \mathbb{B}(0,W)$, then with probability $\ge 1 - \delta$ after $m \ge \mathcal{O} (log(1/\delta)/ \epsilon^2)  $ samples, 
$\hat{err}(\mathbf{w})$ is $\left(\epsilon,\frac{2b^{3}W}{a},\mathbf{w^{*}}\right)-SLQC$
in $\mathbf{w}$.
\end{theorem}
\vspace{-10pt}
\begin{proof} Please see Section \ref{appendix_noisyglm} for the proof.
\end{proof}
\vspace{-3pt}
The above theorems, in combination with the results in \cite{sngd}, allow us to conclude that for a single-output no-hidden-layer neural network with sigmoid or ReLU activation, the error function, $\hat{err}$, is SLQC in $\mathbf{w}$. This is however not directly useful for neural networks, as stated earlier. To this end, we propose a new extension of SLQC relevant to a set of GLMs, such as in a layer of a neural network. We note that all of the following sections are novel contributions, which did not exist earlier.

\subsubsection{SLQC-ness of a Multi-Output Neural Network with No Hidden Layers}
We begin with a revised definition of Local Quasi-Convexity for matrices using the Frobenius inner product.

\begin{defn}[\textbf{\textit{Local Quasi-Convexity for Matrices}}]
\label{def:lqcm}
Let $\mathbf{x},\mathbf{z}\in \mathbb R^{d \times d'},\kappa,\epsilon>0$ and $f: \mathbb{R}^{d \times d'}\to \mathbb{R}$ be a differentiable function. Then $f$ is $(\epsilon,\kappa,\mathbf z)$-Strictly Locally Quasi-Convex (SLQC) in $\mathbf{x}$, if at least one of the following applies:
\vspace{-5pt}
\begin{enumerate}
\setlength\itemsep{0em}
\item $f(\mathbf{x})-f(\mathbf{z})\le\epsilon$
\item $\| \nabla f(\mathbf{x})\|>0$, and \textrm{$\forall \mathbf{y} \in \mathbb{B}\left(\mathbf{z},\epsilon/\kappa\right)$,} $\langle \nabla f(\mathbf{x}), \mathbf{y}-\mathbf{x} \rangle_{F} \le 0$ 
\end{enumerate}
\end{defn}
\vspace{-5pt}
\noindent where $\mathbb{B}\left(\mathbf{z},\epsilon/\kappa\right)$ is a ball centered at $\mathbf{z}$ with radius $\epsilon/\kappa$. ($\langle \cdot, \cdot \rangle_{F}$ denotes the Frobenius inner product.)\\ 


We now show that the error, $\hat{err}(\mathbf{W})$, of a multi-output no-hidden-layer neural network is also SLQC in $\textbf{W}$. (Note that w.r.t. our problem setup in Equation \ref{eq:form_2}, this is equivalent to showing that the one-hidden layer neural network problem is SLQC in $W_2$ alone.) The empirical error function is now given by:
\vspace{-8pt}
\begin{equation*}
    \hat{err}(\mathbf{W})= \frac{1}{m} \sum_{i=1}^m \|\mathbf{y_i}-\phi(\langle\mathbf{W,x_{i}\rangle)}\|^{2}
\end{equation*}
where $\mathbf{x_{i}} \in \mathbb{R}^d$ is the input, $\mathbf{y_{i}} \in \mathbb{R}^{d'}$ is the corresponding correct output, $\mathbf{W} \in \mathbb{R}^{d \times d'}$ is the matrix of weights and $\phi$ is applied element-wise in the multi-output no-hidden-layer neural network. Let the global minimizer of $\hat{err}$ be $\mathbf{W^*}$.
\begin{theorem}
\label{theorem_relu_outer_layer}
Let an idealized single-layer multi-output neural network be characterized by a linear operator $\mathbf{W} \in \mathbb{R}^{d \times d'} = [\mathbf{w}_1 \enskip \mathbf{w}_2 \enskip \cdots \enskip \mathbf{w}_{d'}]$ and a generalized ReLU activation function be applied element-wise $\phi:\mathbb{R}^{d'} \rightarrow \mathbb{R}^{d'}$. Let the output of the layer be $\phi \langle \textbf{W}, \textbf{x} \rangle \in \mathbb{R}^{d'}$ where $\textbf{x} \in \mathbb{R}^d$ is the input. Assuming $\|\mathbf{W^{*}}\|_F\le W$, $\hat{err}(\mathbf{W})$ is $\left(\epsilon,\frac{2b^{3}W}{a},\mathbf{W^{*}}\right)-SLQC$ in $\mathbf{W}$ for all $\mathbf{W}\in \mathbb{B}_{d}(0,W)$ and $\epsilon > 0$.
\end{theorem}
\vspace{-10pt}
\begin{proofsketch} To show this result, we use Definition \ref{def:lqcm}. Let $\textbf{V} = [\textbf{v}_1 \enskip \textbf{v}_2 \cdots \enskip \textbf{v}_{d'}]$ be a point $\epsilon/\kappa$-close to minima $\textbf{W}^*$ with $\kappa = \frac{2b^{3}W}{a}$. Let  $G(\textbf{W})$ be the subgradient of $\hat{err}_m(\textbf{W})$. Then we show that $\langle G(\textbf{W}),\textbf{W}-\textbf{V}\rangle_F \ge 0$, thus proving the result. Section \ref{appendix_relu_outer_layer} presents the complete proof.
\end{proofsketch}

\subsubsection{One Hidden Layer Networks with Single Output Neurons}
Taking this further, we next consider a single-hidden-layer neural network. While the outer layer (layer 2) of a single-hidden-layer neural network can be directly viewed as a set of GLMs (see section \ref{sec_appendix_set}), the inner layer cannot be viewed the same way (due to lack of expected outputs in the hidden layer). We hence need to show that given a fixed $\mathbf{w_2}$, the error $\hat{err}(\mathbf{W_1}, \mathbf{w_2})$ is also SLQC in $\mathbf{W_1}$. In this case, the empirical error function is:
\[ \hat{err}(\mathbf{W_1}, \mathbf{w_2}) = \frac{1}{m} \sum_{i=1}^m \| y_i - \phi_2 \langle \mathbf{w_2}, \phi_1 \langle \mathbf{W_1}, \mathbf{x_i} \rangle \rangle \|^2\]
where $\mathbf{x_{i}} \in \mathbb{R}^d$ is the input; $y_{i} \in \mathbb{R}$ is the corresponding correct output; and $\mathbf{W_1} \in \mathbb{R}^{d \times d'}$, $\mathbf{w_2} \in \mathbb{R}^{d'}$ are the weights of the inner and outer layers respectively. Let the global minimizer of $\hat{err}$ be $(\mathbf{W_1^*}, \mathbf{w_2^*})$. This setting corresponds to the inner layer of single-output single-hidden-layer neural network.
\begin{theorem}
\label{theorem_relu_inner_layer_single_op}
Let an idealized two-layer  neural network be characterized by linear operators $\mathbf{W_1} \in \mathbb{R}^{d \times d'}$, $\mathbf{w_2} \in \mathbb{R}^{d'}$ and generalized ReLU activation functions $\phi_1:\mathbb{R}^{d'} \rightarrow \mathbb{R}^{d'}$, $\phi_2:\mathbb{R} \rightarrow \mathbb{R}$. Assuming $\|\mathbf{W^{*}_1}\|_F\le W_1$, $\|\mathbf{w^{*}_2}\|\le W_2$, $\hat{err}(\mathbf{W_1}, \mathbf{w_2})$ is $\left(\epsilon,\left( \frac{a}{4b^5W_2^2W_1} - \frac{W_1}{\epsilon}\right)^{-1},\mathbf{W^{*}_1}\right)-SLQC$ in $\mathbf{W_1}, \forall \mathbf{W_1}\in \mathbb{B}(0,W_1)$ and $\forall \epsilon > 0$.
\end{theorem}
\vspace{-10pt}
\begin{proofsketch} We again use Definition \ref{def:lqcm} to prove the result. Let $\mathbf{V_1}$ be a point $\frac{\epsilon}{\kappa}$ close to minima. We show that $\langle \nabla_{\mathbf{W_1}} \hat{err}(\mathbf{W_1}, \mathbf{w_2}),\textbf{W}_1-\textbf{V}_1\rangle_F \ge 0$. As a consequence of Definition \ref{def:lqcm}, this proves the result. See Section \ref{appendix_relu_inner_layer_single_op} for the complete proof.
\end{proofsketch}
\vspace{-3pt}

\subsubsection{One Hidden Layer Networks with Multiple Output Neurons}
The above results together postulate that a single-hidden-layer neural network is layer-wise SLQC. We use the above result to show that error $\hat{err}(\mathbf W_1, \mathbf W_2)$ is SLQC in $\mathbf W_1$ for single-hidden-layer neural network, even with multiple outputs. The empirical error function in this setting is:
\vspace{-8pt}
\begin{equation*}
    \hat{err}(\mathbf{W_1}, \mathbf{W_2}) = \frac{1}{m} \sum_{i=1}^m \| \mathbf{y_i} - \phi_2 \langle \mathbf{W_2}, \phi_1 \langle \mathbf{W_1}, \mathbf{x_i} \rangle \rangle \|^2
\end{equation*}
where $\mathbf{x_{i}} \in \mathbb{R}^d$ is the input, $\mathbf{y_{i}} \in \mathbb{R}^{d^{''}}$ is the corresponding correct output, $\mathbf{W_1} \in \mathbb{R}^{d \times d'}$, $\mathbf{W_2} \in \mathbb{R}^{d^{'} \times d^{''}}$ are the weights of the inner and outer layers respectively. Let the global minimizer of $\hat{err}$ be $(\mathbf{W_1^*}, \mathbf{W_2^*})$. This setting corresponds to the inner layer of a multi-output single-hidden-layer neural network.
\begin{theorem}
\label{theorem_relu_inner_layer_multi_op}
Let an idealized two-layer  neural network be characterized by linear operators $\mathbf{W_1} \in \mathbb{R}^{d \times d'}$, $\mathbf{W_2} \in \mathbb{R}^{d' \times d^{''}}$ and generalized ReLU activation functions $\phi_1:\mathbb{R}^{d'} \rightarrow \mathbb{R}^{d'}$, $\phi_2:\mathbb{R}^{d''} \rightarrow \mathbb{R}^{d''}$. Assuming $\|\mathbf{W^{*}_1}\|_F\le W_1$, $\|\mathbf{W^{*}_2}\|_F\le W_2$, $\hat{err}(\mathbf{W_1}, \mathbf{W_2})$ is $\left(\epsilon,\left( \frac{a}{4b^5W_2^2W_1} - \frac{W_1}{\epsilon}\right)^{-1},\mathbf{W^{*}_1}\right)-SLQC$ in $\mathbf{W_1}, \forall \mathbf{W_1}\in \mathbb{B}(0,W_1)$ and $\forall \epsilon > 0$.
\end{theorem}
\vspace{-10pt}
\begin{proofsketch} We use Theorem \ref{theorem_relu_inner_layer_single_op} to prove this result. The error $\hat{err}(\mathbf{W_1}, \mathbf{W_2})$ of a multi-output single-hidden layer network can be seen as the sum of errors of $d''$ single-output single-hidden layer networks. This observation combined with Theorem \ref{theorem_relu_inner_layer_single_op} is used to prove the result. See Section \ref{appendix_relu_inner_layer_multi_op} for the complete proof.
\end{proofsketch}
\vspace{-3pt}

While the above results have been shown with the generalized ReLU, each of these results also holds for sigmoid activation functions. Moreover, most other widely used error functions such as cross-entropy loss are convex (and thus SLQC) as well as Lipschitz. We believe that our results can be extended to most commonly used error functions, and we present an extension to cross-entropy loss below in Section \ref{subsec_rationale_extension}. 

\subsubsection{The ReLU Case}
In an earlier subsection, we defined the Generalized ReLU in Defn \ref{defn_gen_relu}. Note that this definition does not cover the standard ReLU $(a=0)$. We call the standard ReLU as ReLU in this subsection. We hence now provide additional results for networks having ReLU as the activation function. These results are, naturally, quite similar to the ones for Generalized ReLU. Note that for ReLU as the activation function, the subgradient $g$ would be $0$ for $x < 0$ and $b$ otherwise.

As in the previous subsection, consider first the case of a network having no hidden layers and only one output neuron. In this case we have the following corollary (to Theorem \ref{theorem_iglm_relu}):

\begin{corollary}
In the idealized GLM with ReLU activation, assuming $\|\mathbf{w^{*}}\|\le W$, 
$\hat{err}(\mathbf{w})$ is $\left(\epsilon,2b^{2}W,\mathbf{w^{*}}\right)-SLQC$
in $\mathbf{w}, \forall \mathbf{w}\in \mathbb{B}(0,W)$ and $\forall \epsilon > 0$.
\label{cor_iglm_relu}
\end{corollary}
\vspace{-13pt}
\begin{proofsketch} The proof is similar to the proof of Theorem \ref{theorem_iglm_relu}. See Section \ref{appendix_idealglm_cor} for the complete proof.
\end{proofsketch}

Now consider the case with no hidden layer but multiple output neurons. For this case, we have the following corollary (to Theorem \ref{theorem_relu_outer_layer}).

\begin{corollary}
\label{cor_relu_outer_layer}
Let an idealized single-layer multi-output neural network be characterized by a linear operator $\mathbf{W} \in \mathbb{R}^{d \times d'} = [\mathbf{w}_1 \enskip \mathbf{w}_2 \enskip \cdots \enskip \mathbf{w}_{d'}]$ and a standard ReLU activation function  applied element-wise $\phi:\mathbb{R}^{d'} \rightarrow \mathbb{R}^{d'}$. Let the output of the layer be $\phi \langle \textbf{W}, \textbf{x} \rangle \in \mathbb{R}^{d'}$ where $\textbf{x} \in \mathbb{R}^d$ is the input. Assuming $\|\mathbf{W^{*}}\|_F\le W$, $\hat{err}(\mathbf{W})$ is $\left(\epsilon,2b^{2}W,\mathbf{W^{*}}\right)-SLQC$ in $\mathbf{W}$ for all $\mathbf{W}\in \mathbb{B}_{d}(0,W)$ and $\epsilon > 0$.
\end{corollary}
\vspace{-10pt}
\begin{proofsketch} The proof is similar to the proof of Theorem \ref{theorem_relu_outer_layer}, Section \ref{appendix_cor_relu_outer_layer} presents the complete proof.
\end{proofsketch}

The above results allow us to extend our results to the standard ReLU activation function. We leave the specifics of extending to the one-hidden layer network to the reader. We note, however, that all of our experiments in this work are conducted with the Leaky ReLU which satisfies the Generalized ReLU definition in Defn \ref{defn_gen_relu}, and hence is in line with the theorems proved in earlier subsections.


\subsection{Extension to Cross-Entropy Loss}
\label{subsec_rationale_extension}

We now present results for neural networks trained using the Cross Entropy Error, $- \sum_i (y_i\log(p_i))$ (where $p_i$ is the predicted probability for the $i$th output class from the neural network, obtained using the sigmoid function on a single output neuron, or softmax function when there are multiple output neurons), as loss instead of Mean Square Error. We first consider the case of a network having no hidden layer and only neuron in the output layer. This corresponds to use of $J$ as binary cross-entropy loss in Equation \ref{eq_mlnn_loss}, with a sigmoid activation in the output layer. (We wish to highlight that the proof techniques used in the results below are very different from those in \cite{sngd}.)

\begin{theorem}
\label{thm_iglm_ce}
In the idealized GLM like setting with sigmoid activation and binary cross entropy loss, assuming $\|\mathbf{w^{*}}\|\le W$, 
$\hat{err}(\mathbf{w})$ is $\left(\epsilon,\frac{\epsilon}{(1 - e^{-\epsilon})^2},\mathbf{w^{*}}\right)-SLQC$
in $\mathbf{w}, \forall \mathbf{w}\in \mathbb{B}(0,W)$ and $\forall \epsilon > 0$.
\end{theorem}
\begin{proofsketch}
To show this, we use Definition \ref{defn:lqc}. With a point $\mathbf{v}$, $\epsilon / \kappa$ close to $\mathbf{w}^*$, we show that $\langle G(\mathbf{w}), \mathbf{w} - \mathbf{v} \rangle \ge 0$. Section \ref{appendix_iglm_ce} presents the complete proof.
\end{proofsketch}

We next consider the case of no hidden layers but multiple output neurons. In this case, given cross-entropy error as the loss function, the following result holds:

\begin{theorem}
\label{theorem_outer_layer_ce}
Let an idealized single-layer multi-output neural network be characterized by a linear operator $\mathbf{W} \in \mathbb{R}^{d \times d'} = [\mathbf{w}_1 \enskip \mathbf{w}_2 \enskip \cdots \enskip \mathbf{w}_{d'}]$ and softmax activation function applied $\phi:\mathbb{R}^{d'} \rightarrow \mathbb{R}^{d'}$. Let the output of the layer be $\phi \langle \textbf{W}, \textbf{x} \rangle \in \mathbb{R}^{d'}$ where $\textbf{x} \in \mathbb{R}^d$ is the input and the loss function, $\hat{err}$, used is Cross-Entropy Error. Assuming $\|\mathbf{W^{*}}\|_F\le W$, $\hat{err}(\mathbf{W})$ is $\left(\epsilon,\frac{\epsilon d'}{(1 - e^{-\epsilon})^2},\mathbf{W^{*}}\right)-SLQC$ in $\mathbf{W}$ for all $\mathbf{W}\in \mathbb{B}_{d}(0,W)$ and $\epsilon > 0$.
\end{theorem}
\begin{proof}
See Section \ref{appendix_outer_layer_ce} for the proof.
\end{proof}

The above results allow us to extend our results in Section \ref{subsec_rationale} to the cross-entropy loss. We leave the specific statement of the above result for one hidden-layer networks to the reader. We however do experimentally study the use of cross-entropy loss in our methodology, even for deep neural networks, in Section \ref{sec_expts_new}. Also, while the cross-entropy loss is more closely related to sigmoid-activated neurons, we empirically also study the use of leaky ReLU as activations in hidden layers in our experiments (Section \ref{sec_expts_new}).

Given the above background of results, we now present our methodology to train a multi-layer neural network effectively using these results.



\subsection{Methodology}
\label{sec_methodology} 
As stated earlier, \cite{sngd} showed that Stochastic Normalized Gradient Descent (SNGD) converges with high probability to the optimum for SLQC functions. We leverage this result to arrive at a formal procedure to train neural networks effectively. 
To this end, we begin by briefly reviewing the Stochastic Normalized Gradient Descent (SNGD) method, and state the relevant result.
\begin{algorithm}[t]
 \caption{Stochastic Normalized Gradient Descent (SNGD)}
    \label{alg_sngd}
    \begin{algorithmic}
    \STATE {\bfseries Input:} Number of iterations $T$, training data $S = \{(\textbf{x}_{i}, y_{i})\}_{i=1}^{m}\in \mathbb{R}^{d} \times \mathbb R$, learning rate $\eta$, minibatch size $b$, Initialization parameters $\mathbf{w}_{0}$
    	\FOR{$t=1$ to $T$}
	\STATE Select a random mini-batch of training points by sampling $\{(\mathbf x_i, y_i)\}_{i=1}^{b}\sim \text{\text{Uniform}}(S)$ \hfill
		
	\STATE Let $f_{t}(\mathbf{w})=\frac{1}{b}\sum_{i=1}^{b}(y_i - \phi\langle \mathbf w,\mathbf x_i \rangle)^2$ 
		
	\STATE Let $\mathbf g_{t}=\nabla f_{t}(\mathbf{w}_{t})$, and $\hat{\mathbf g}(t)=\frac{\mathbf g_t}{\|\mathbf g_t\|}$ 
    
    \STATE $\mathbf w_{t+1} = \mathbf w_{t}-\eta\cdot\hat{\mathbf g_{t}}$
        \ENDFOR
        
    \STATE {\bfseries Output:} Model given by $\mathbf{w}_{T}$
    \end{algorithmic}    
\end{algorithm}
\vspace{-5pt}
\subsubsection{Stochastic Normalized Gradient Descent (SNGD)} 
Normalized Gradient Descent (NGD) is an adaptation of traditional Gradient Descent, where the updates in each iteration are based only on the direction of the gradients. This is achieved by normalizing the gradients. SNGD is the stochastic version of NGD, where weight updates are performed using individual (randomly chosen) training samples, instead of the complete set of samples. Mini-batch SNGD generalizes this by applying updates to the parameters at the end of every mini-batch of samples, as does mini-batch Stochastic Gradient Descent (SGD). In the remainder of this paper, we refer to mini-batch SNGD as SNGD itself, as is common for SGD. Algorithm \ref{alg_sngd} describes the SNGD methodology for a generic problem.

We now state the result showing the effectiveness of SNGD for SLQC functions.
\begin{theorem}[~\cite{sngd}]
\label{theorem_sngd_hazan}
Let $\epsilon,\delta,G,M,\kappa>0$, let $f:\mathbb{R}^{d}\to\mathbb{R}$
and $\mathbf{w}^{*} = \arg\min_{\mathbf{w}}f(\mathbf{w})$. Assume
that for $b\geq b_{0}(\epsilon,\delta,T)$, with probability $\ge1-\delta$,
$f_t$ defined in Algorithm \ref{alg_sngd} is $(\epsilon,\kappa,\mathbf{w}^{*})$-SLQC $\forall\textbf{w}$,
and $|f_t|\le M \forall t\in\{1,\cdots,T\}$ . If we run SNGD with
$T\ge\frac{\kappa^{2}||\mathbf{w}_{1}-\mathbf{w}^{*}||^{2}}{\epsilon^{2}}$
and $\eta=\frac{\epsilon}{\kappa}$, and $b\ge\max\left\{ \frac{M^{2}log\left(\frac{4T}{\delta}\right)}{2\epsilon^{2}},b_{0}(\epsilon,\delta,T)\right\} $,
with probability $1-2\delta$, $f(\mathbf{w})-f(\mathbf{w}^{*})\le3\epsilon$\footnote{Replacing inner product with Frobenius inner product in the proof for this result in \cite{sngd} allows us to extend this result to our definition of SLQC for matrices.}.
\end{theorem}
Importantly, note that the convergence rate of SNGD depends on the $\kappa$ parameter. While the GLM error function with sigmoid activation has $\kappa = e^W$ (stated earlier in the section), the generalized ReLU setting introduced in this work has $\kappa = \frac{2b^{3}W}{a}$ (i.e. linear in $W$) for both GLMs and layers, which is an exponential improvement for the SNGD procedure's effectiveness. This is significant as the number of iterations $T$ in Theorem~\ref{theorem_sngd_hazan} depends on $\kappa^2$. In other words, SNGD offers accelerated convergence with the proposed generalized ReLU layers as compared to sigmoid GLMs proposed earlier.


\subsubsection{DANTE}
We have thus far shown that each layer of the considered one-hidden-layer neural network comprises of a set of SLQC problems, each independent in its parameters. Also, SNGD provides an effective method for each such SLQC problem to converge to its respective $\epsilon$-suboptimal solution with high probability, as shown in Theorem \ref{theorem_sngd_hazan}. This allows us to propose an alternating strategy, \myalgo, where each individual SLQC problem is effectively solved (or each individual layer is effectively trained) using SNGD, which we now present. We note that although \myalgo uses stochastic gradient-style methods internally (such as SNGD), the overall strategy adopted by \myalgo is not necessarily a descent-based strategy, but an alternating-minimization strategy.




Consider the optimization problem below for a single hidden layer network:
\vspace{-3pt}
\[
\min_{\mathbf{W}} \ f(\mathbf{W}_1, \mathbf{W}_2) = \mathbb E_{\mathbf x \sim \cal D} \|\phi_2 \langle \mathbf{W}_2,\phi_{1} \langle \mathbf{W}_{1}, \mathbf{x} \rangle \rangle - \mathbf y\|_2^2
\]
\noindent As seen in Section \ref{subsec_rationale}, on fixing each of $W_1$ and $W_2$, we have an SLQC problem. On fixing $W_1$, we have the SLQC problem:
\vspace{-3pt}
\[
\min_{\mathbf{W}} \ \mathbb E_{\mathbf x \sim \cal D} \|\phi_2 \langle \mathbf{W}_2, \mathbf z \rangle - \mathbf y\|_2^2,
\]
\noindent where $\mathbf z = \phi_{1} \langle \mathbf{W}_{1}, \mathbf{x} \rangle$. On fixing $\mathbf{W}_2$, we have the following SLQC problem:
\vspace{-3pt}
\[
\min_{\mathbf{W}} \ \mathbb E_{\mathbf x \sim \cal D} \|\phi_{\mathbf{W}_2} \langle \mathbf{W}_1, \mathbf x \rangle - \mathbf y\|_2^2,
\]
\myalgo optimizes the empirical risk associated with each of these intermediate problems using SNGD steps by sampling several mini-batches of data points and performing updates as in Algorithm \ref{alg_sngd}. Algorithm~\ref{alg_am} provides the complete algorithm for the proposed method. Note that the results from the last two subsections hold for any weights and are not limited to the initialized weights. For example if the network is initialized with $\mathbf{W}^0_1$ and $\mathbf{W}^0_2$ and after training the layers once, we obtain weights $\mathbf{W}^1_1$ and $\mathbf{W}^1_2$. The SNGD conditions would still hold for this pair of weights, justifying the applicability of the algorithm.

\begin{algorithm}[t]
    \caption{Deep AlterNations for Training nEural networks (\myalgo)}
    \label{alg_am}
    \begin{algorithmic}
    \STATE  {\bfseries Input:} Stopping threshold $\epsilon$, Number of iterations of alternating minimization $T_{AM}$, Number of iterations for SNGD $T_{SNGD}$, initial values $W_1^0, W_2^0$, learning rate $\eta$, minibatch size $b$
    \STATE  $t := 1$
    \WHILE{$|f(W_1^t,W_2^t) - f(W_1^{t-1},W_2^{t-1})| \ge \epsilon $ \textbf{or} $t < T_{AM}$}
    \STATE $W_2^t \leftarrow \underset{W}{\arg\min} \; \mathbb{E}_{\mathbf x \sim \cal D} \|\phi_2 \langle W,\phi_{1} \langle W_{1}^{t-1}, \mathbf{x} \rangle \rangle - \mathbf y\|_2^2$ \hfill \texttt{//use SNGD}
		
	\STATE		$W_1^t \leftarrow \underset{W}{\arg\min} \; \mathbb{E}_{\mathbf x \sim \cal D} \|\phi_2 \langle W_2^t,\phi_{1} \langle W, \mathbf{x} \rangle \rangle - \mathbf y\|_2^2$ \hfill \texttt{//use SNGD}
       
	\STATE     $t := t + 1$
    \ENDWHILE
    \STATE  {\bfseries Output:} $W_{1}^{t-1}, W_{2}^{t-1}$
    \end{algorithmic}

\end{algorithm}
\begin{algorithm}[t]
    \caption{\myalgo for a multi-layer auto-encoder}
    \label{alg_multi}
\begin{algorithmic}
\STATE {\bfseries Input:} Network with $2n - 1$ hidden layers and weights $\mathbf{W}_1, \dots \mathbf{W}_{2n}$
\FOR {$l=1$ to $n$}
\STATE Consider the one-hidden layer network formed by $\mathbf{W}_l$ and $\mathbf{W}_{2n-l+1}$.
\STATE Train $\mathbf{W}_l$ and $\mathbf{W}_{2n-l+1}$ using Algorithm \ref{alg_am}
\ENDFOR
\STATE {\bfseries Output:} Trained $\mathbf{W}_1, \dots \mathbf{W}_{2n}$
\end{algorithmic}
\end{algorithm}

\subsection{Extending to a Multi-Layer Neural Network}
\label{sec_multi}

In the previous sections, we illustrated how a single hidden-layer neural network can be cast as a set of SLQC problems and proposed an alternating minimization method, DANTE. This approach can be generalized to deep auto-encoders by considering a greedy layer-wise approach to training a neural network \cite{bengio2007greedy}. Unlike earlier layer-wise training efforts where such training is used only as a pretraining step, no further finetuning is necessary in our methodology; the layer-wise training directly results in the final model. 

\begin{figure}[h]%
    \centering
    \subfloat[Phase - 1]{\includegraphics[width=4.5cm]{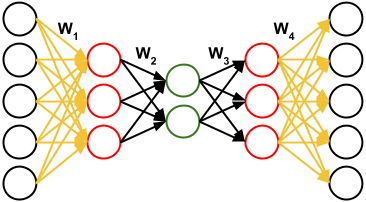} }%
    \qquad
    \subfloat[Phase - 2]{\includegraphics[width=4.5cm]{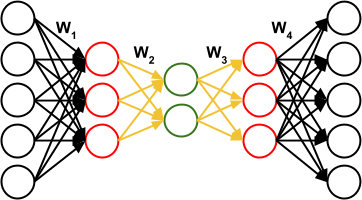} }%
    \caption{An illustration of the proposed multi-layer \myalgo (best viewed in color). In training phase 1, the outer pairs of weights (shaded in gold) are treated as a single-hidden-layer neural network and trained using single-layer \myalgo. In phase 2, the inner pair of weights (shaded in gold) are treated as a single-hidden-layer neural network and trained using single-layer \myalgo.}%
    \label{fig:multi_phases}%
		\vspace*{-2ex}
\end{figure}

We now describe our approach. Consider for example a three-hidden layer autoencoder as pictured in figure \ref{fig:multi_phases}. Say the weights in the network are $\mathbf{W}_1, \mathbf{W}_2, \mathbf{W}_3$ and $\mathbf{W}_4$ respectively from the leftmost to the rightmost layer. In the first phase we consider the one-hidden layer network obtained by the weights $\mathbf{W}_1$ and $\mathbf{W}_4$ (the network of layer dimensions $5 \rightarrow{} 3 \rightarrow{} 5$). We train these two weights using our one-hidden layer DANTE algorithm (section \ref{sec_methodology}). Once these layers are trained, in the second phase, we consider the one-hidden layer network obtained by the weights $\mathbf{W}_2$ and $\mathbf{W}_3$ (a network of layer dimensions $3 \rightarrow{} 2 \rightarrow{} 3$). We train these weights by the one-hidden layer DANTE algorithm with the input and output being $\phi_1 \langle \mathbf{W}_1, \mathbf{x} \rangle$ (the activations of the first hidden layer). This example demonstrated the overall idea behind deep autoencoder training. For a general deep autoencoder, we take pairs of weights symmetric from the center and train them moving from the farthest pair to the one formed by the center layers. Algorithm \ref{alg_multi} summarizes the proposed approach to use DANTE for a deep neural network, and Figure \ref{fig:multi_phases} illustrates the approach.

Note that it is possible to use other schemes to use \myalgo for multi-layer neural networks such as a round-robin scheme, where each layer is trained separately one after the other in the sequence in which the layers appear in the network. Our experiments found that both of these approaches (Algorithm \ref{alg_multi} and round-robin scheme) work equally well for autoencoders. To train multi-layer neural networks we use the round-robin scheme.

Following earlier efforts on alternating optimization for neural networks \cite{taylor2016training}\cite{jagpratap2018relu}, we note that proving convergence for alternating minimization methods that train neural networks is not straightforward and a significant effort by itself, and hence is left as an important direction of future work. We focus this work on identifying this alternating minimization procedure, which is derived from a sound understanding of the individual problems underneath (we believe this is a contribution by itself when looking for alternatives to backpropagation), and showcase its empirical effectiveness.


\section{Experiments and Results}
\label{sec_expts_new}
We validated \myalgo by training feedforward neural networks, as well as autoencoders, on standard datasets including MNIST, Kuzushiji-MNIST (KMNIST) \cite{clanuwat2018deep}, SVHN, CIFAR-10 and Tiny ImageNet. The Tiny-Imagenet dataset has 200 classes while the others have 10 classes each. We followed the benchmark training and evaluation protocols established for each of these datasets. We studied the training and test loss as well as the test accuracy on all our experiments. We used vanilla SGD-based backpropagation (henceforth, called SGD in the experiments) as the baseline method. In order to ensure fair comparison between SGD and \myalgo, we tried different learning rates and picked the best ones individually for both methods. We show the comparative results with these best learning rates. We also show results later in this section using adaptive learning rate methods on both learning schemes.

Note that in all the presented results (unless explicitly stated otherwise), the X-axis is the number of weights updated. We choose this as the reference instead of number of epochs to be fair to DANTE as it updates fewer weights than SGD in any given epoch (where only one pair of layers is updated). In graphs comparing SGD and DANTE, the blue curve is always DANTE and the green one is SGD.

\subsection{Feedforward Neural Networks}
This subsection presents the comparative performance of SGD and \myalgo on feedforward neural networks. To ensure an exhaustive comparison, we used multiple datasets and varied network widths and depths in our experiments. Our initial experiments use Leaky ReLU as the activation function, with $a=0.01$ and $b=1$, as well as sigmoid activation, and Mean Square Error as the loss function (we later show results with cross-entropy error). 
\begin{figure*}[]
    \setlength\tabcolsep{0pt}
    \centering
    \begin{tabular}{cccccc}
        \hspace*{-3pt} \includegraphics[width=.16\textwidth]{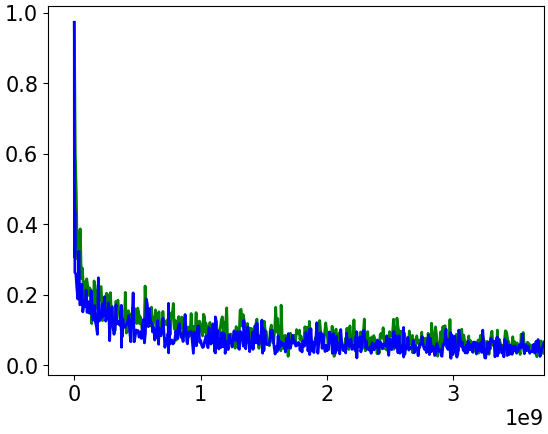} & 
        \hspace*{-3pt} \includegraphics[width=.16\textwidth]{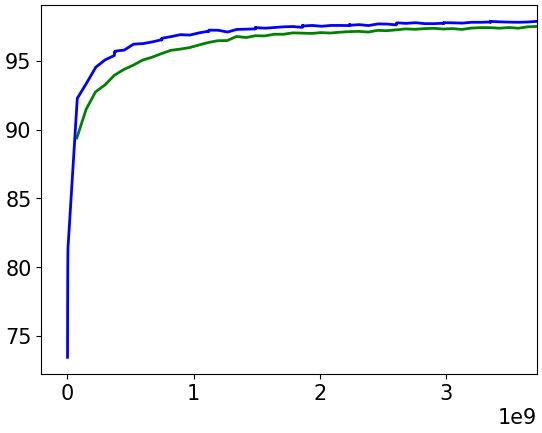} & 
        \hspace*{-3pt} \includegraphics[width=.16\textwidth]{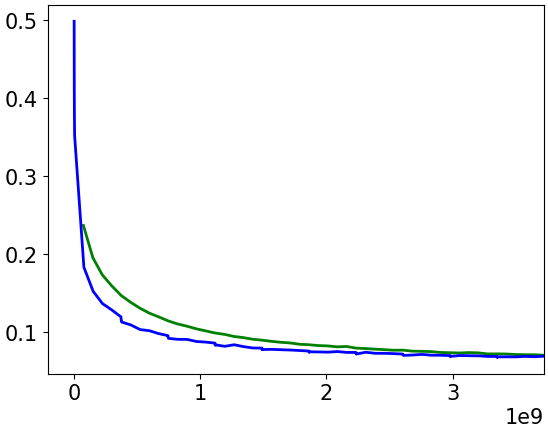} &
        \hspace*{-3pt} \includegraphics[width=.16\textwidth]{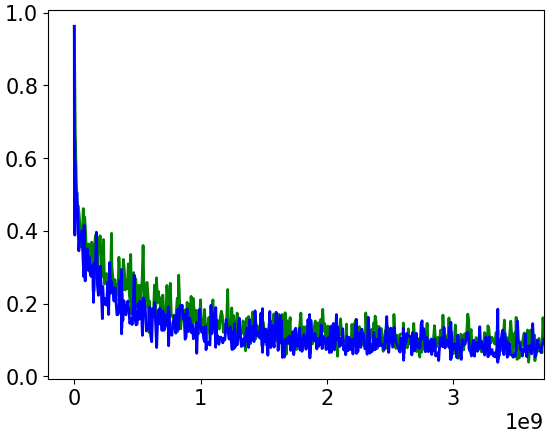} & 
        \hspace*{-3pt} \includegraphics[width=.16\textwidth]{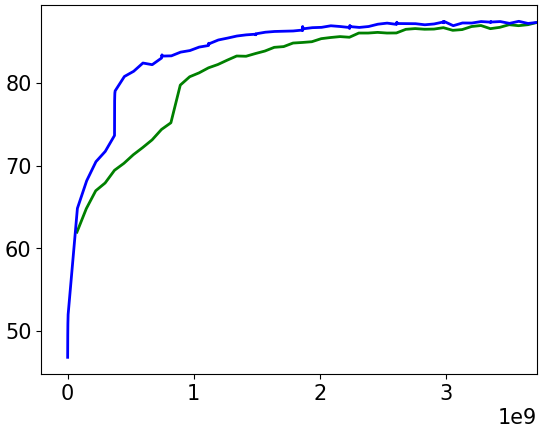} & 
        \hspace*{-3pt} \includegraphics[width=.16\textwidth]{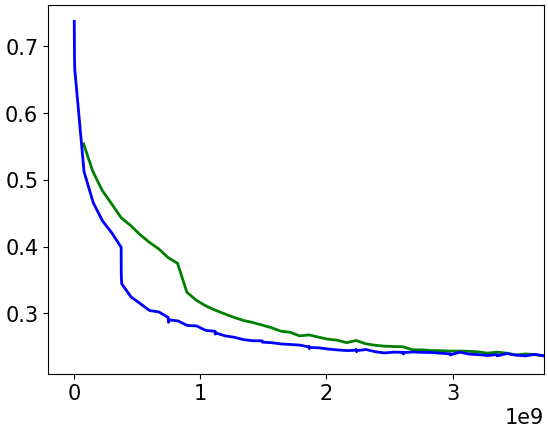} \\
        \hspace*{-3pt} \includegraphics[width=.16\textwidth]{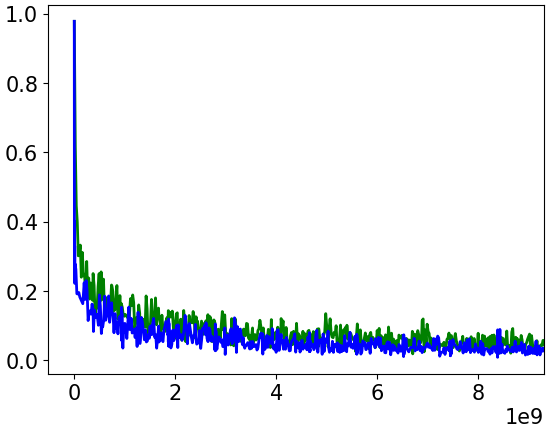} & 
        \hspace*{-3pt} \includegraphics[width=.16\textwidth]{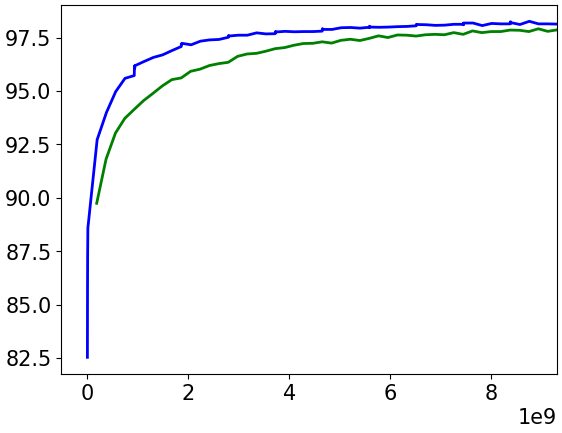} & 
        \hspace*{-3pt} \includegraphics[width=.16\textwidth]{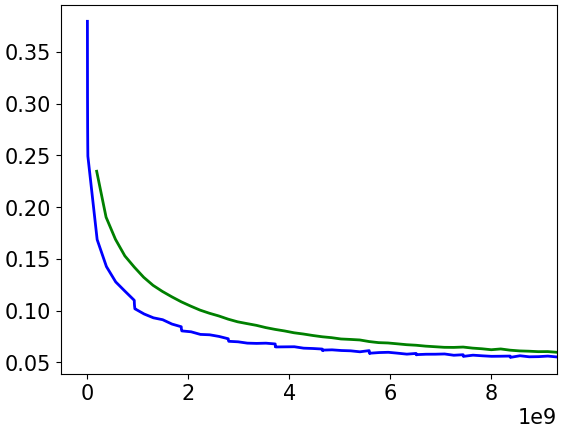} &
        \hspace*{-3pt} \includegraphics[width=.16\textwidth]{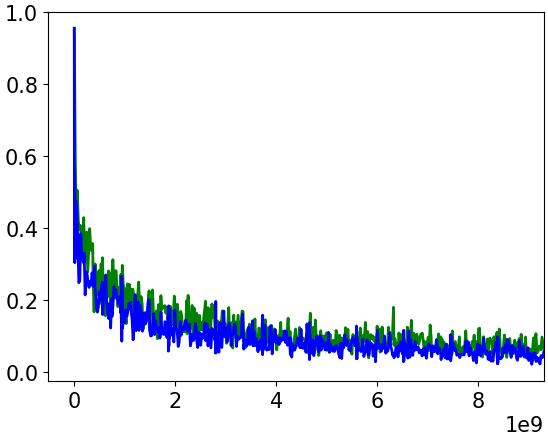} & 
        \hspace*{-3pt} \includegraphics[width=.16\textwidth]{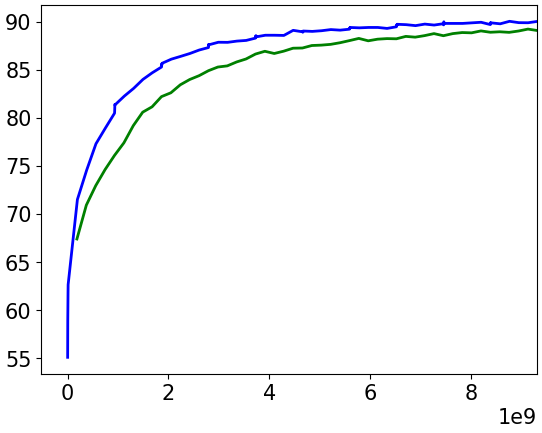} & 
        \hspace*{-3pt} \includegraphics[width=.16\textwidth]{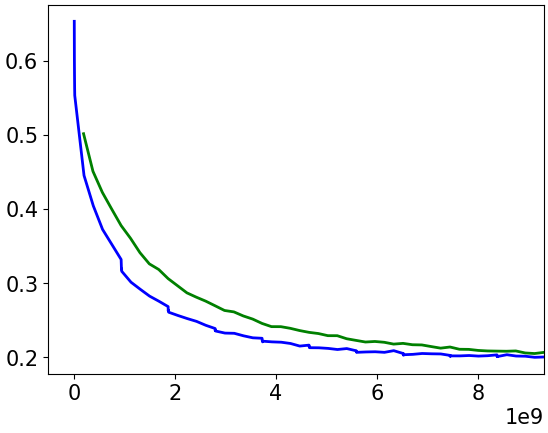} \\
        \hspace*{-3pt} \includegraphics[width=.16\textwidth]{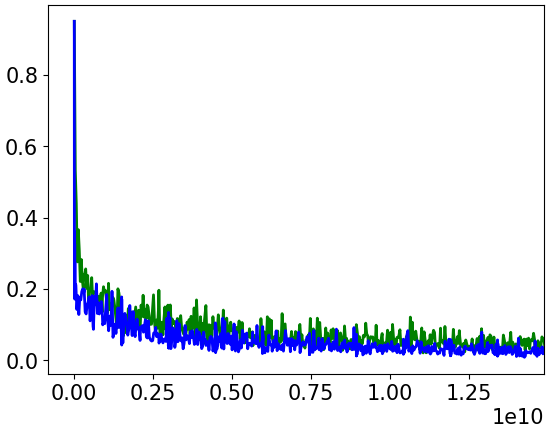} & 
        \hspace*{-3pt} \includegraphics[width=.16\textwidth]{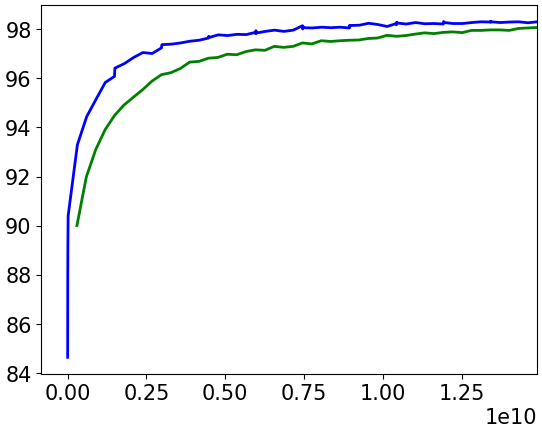} & 
        \hspace*{-3pt} \includegraphics[width=.16\textwidth]{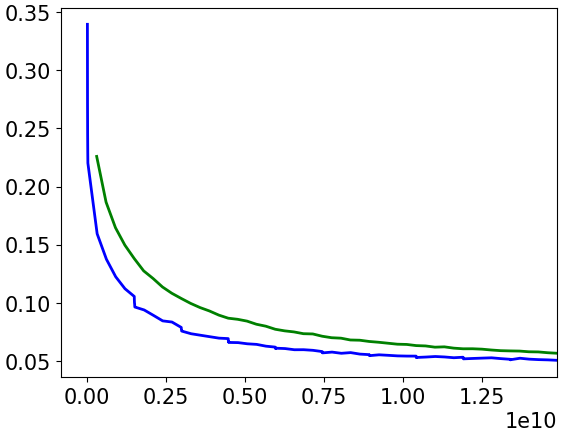} &
        \hspace*{-3pt} \includegraphics[width=.16\textwidth]{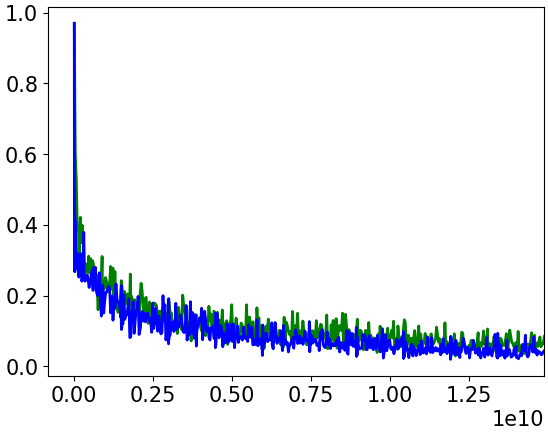} & 
        \hspace*{-3pt} \includegraphics[width=.16\textwidth]{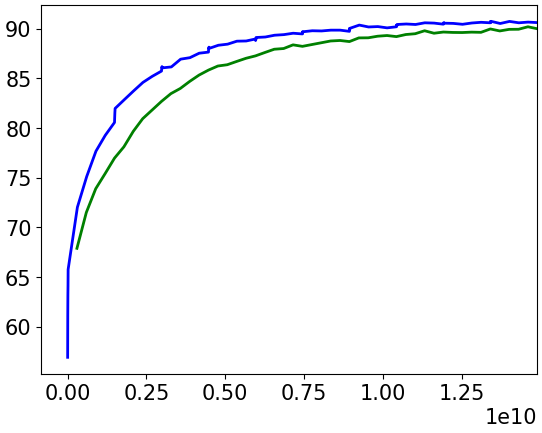} & 
        \hspace*{-3pt} \includegraphics[width=.16\textwidth]{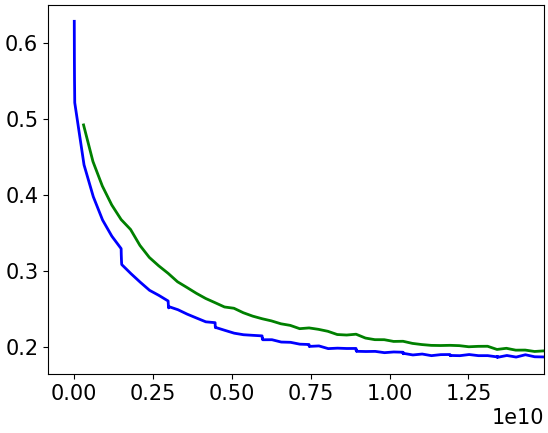} \\
        \hspace*{-3pt} \includegraphics[width=.16\textwidth]{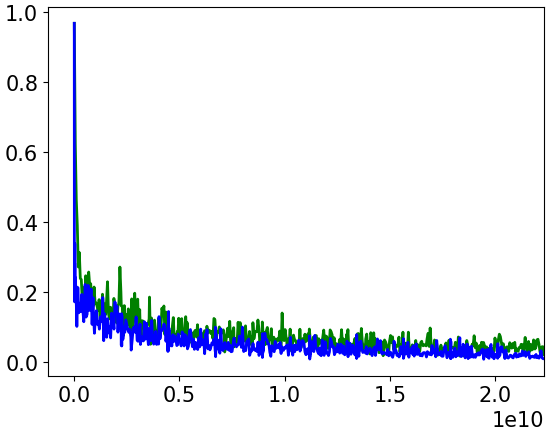} & 
        \hspace*{-3pt} \includegraphics[width=.16\textwidth]{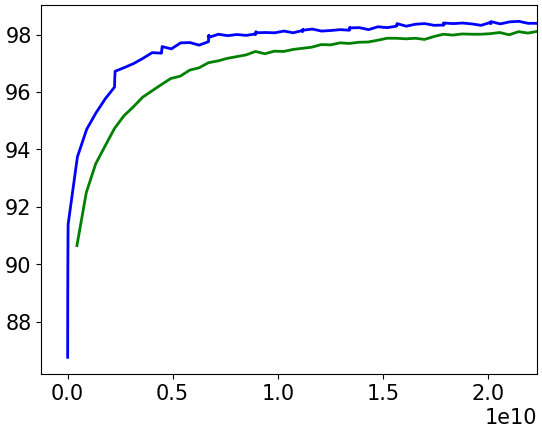} & 
        \hspace*{-3pt} \includegraphics[width=.16\textwidth]{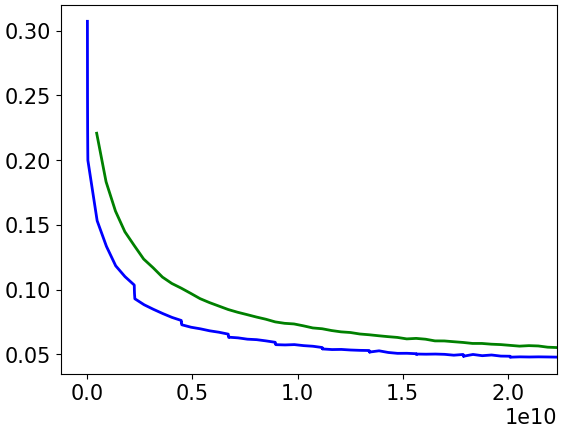} &
        \hspace*{-3pt} \includegraphics[width=.16\textwidth]{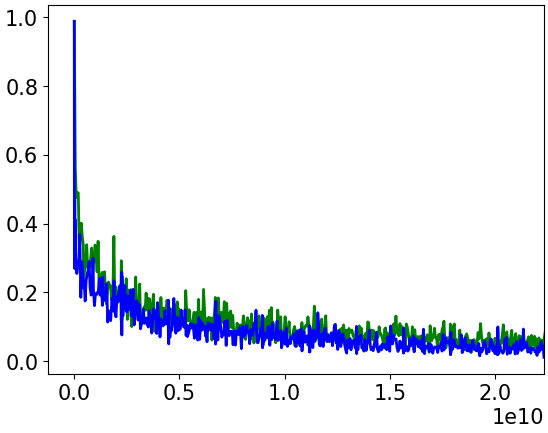} & 
        \hspace*{-3pt} \includegraphics[width=.16\textwidth]{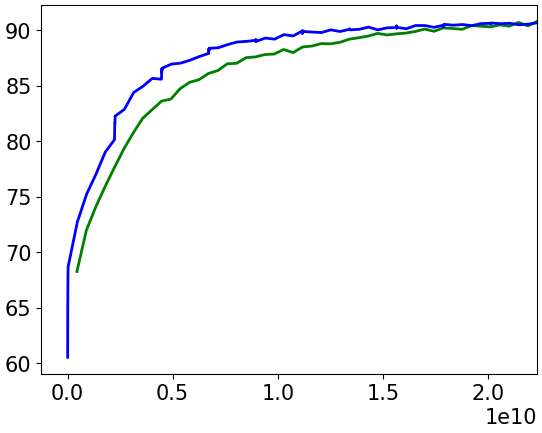} & 
        \hspace*{-3pt} \includegraphics[width=.16\textwidth]{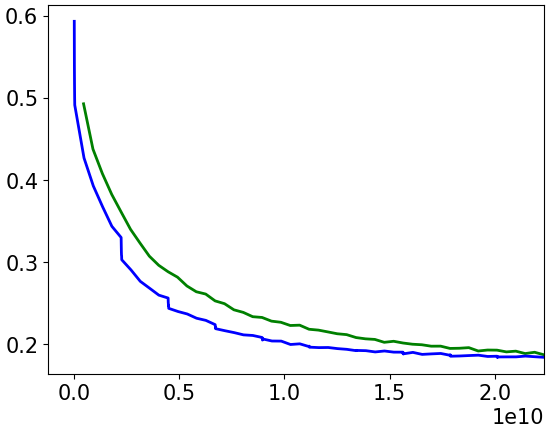} \\
        \hspace*{-3pt} \includegraphics[width=.16\textwidth]{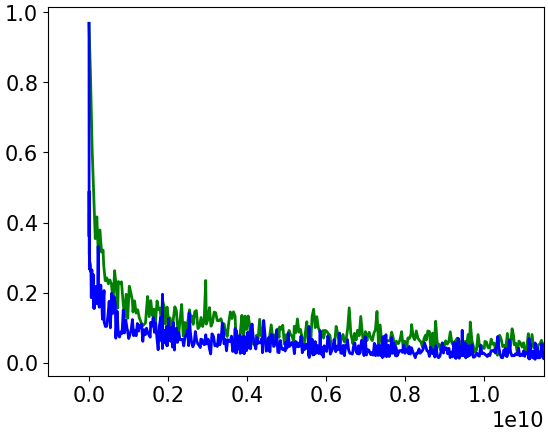} & 
        \hspace*{-3pt} \includegraphics[width=.16\textwidth]{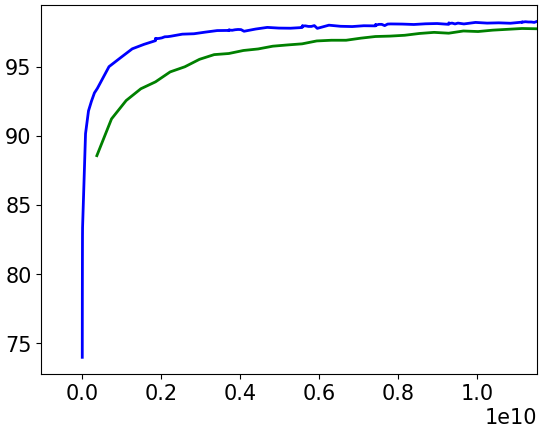} & 
        \hspace*{-3pt} \includegraphics[width=.16\textwidth]{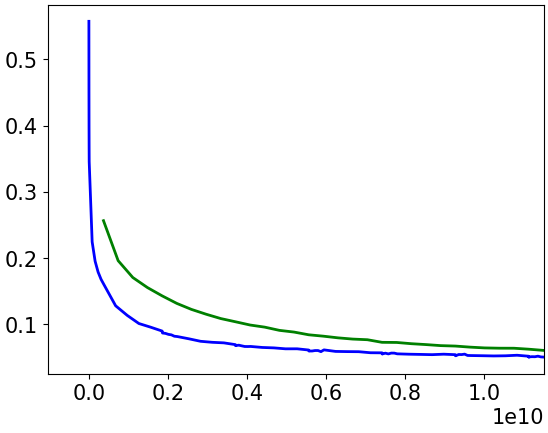}&
        \hspace*{-3pt} \includegraphics[width=.16\textwidth]{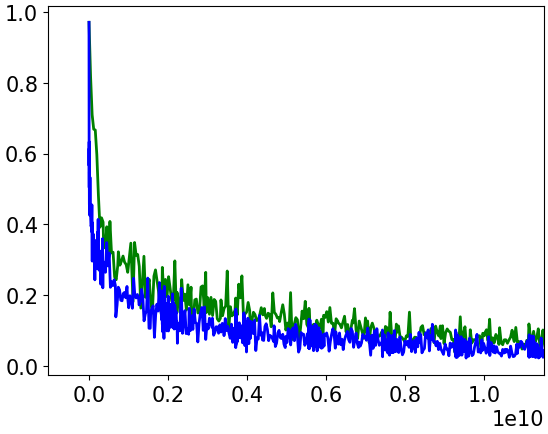} & 
        \hspace*{-3pt} \includegraphics[width=.16\textwidth]{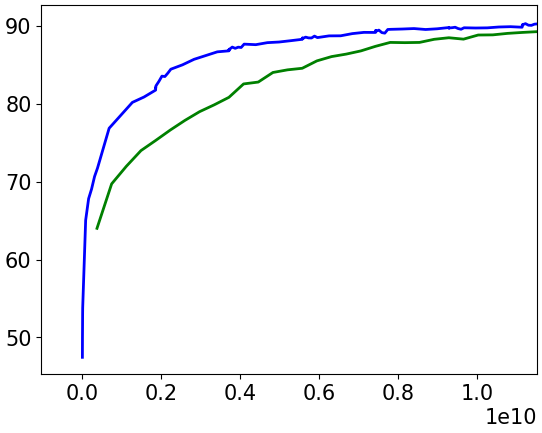} & 
        \hspace*{-3pt} \includegraphics[width=.16\textwidth]{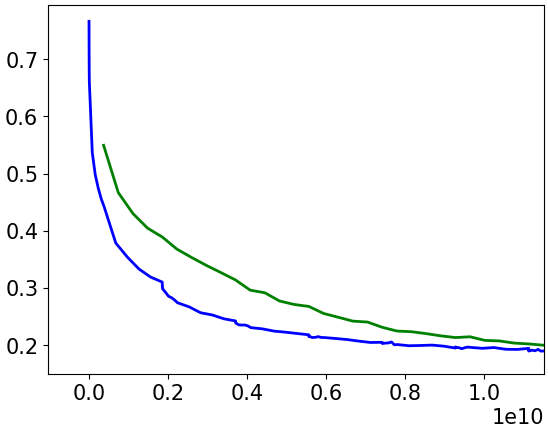} \\
        \hspace*{-3pt} \includegraphics[width=.16\textwidth]{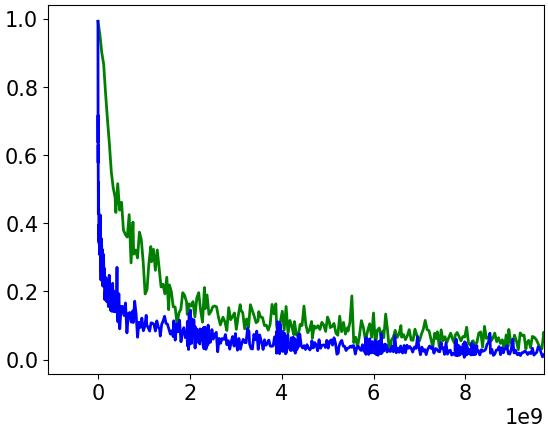} & 
        \hspace*{-3pt} \includegraphics[width=.16\textwidth]{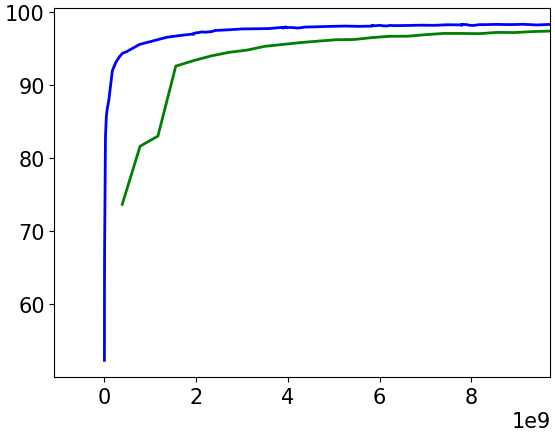} & 
        \hspace*{-3pt} \includegraphics[width=.16\textwidth]{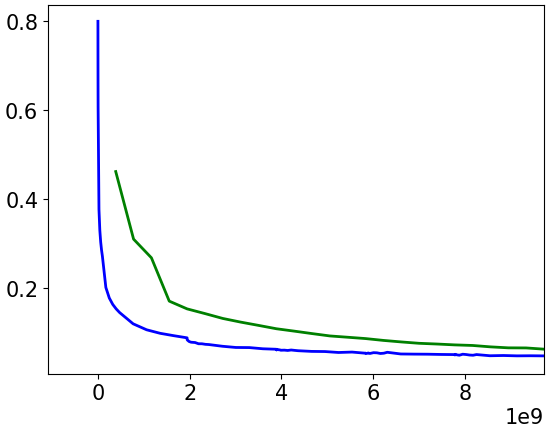} &
        \hspace*{-3pt} \includegraphics[width=.16\textwidth]{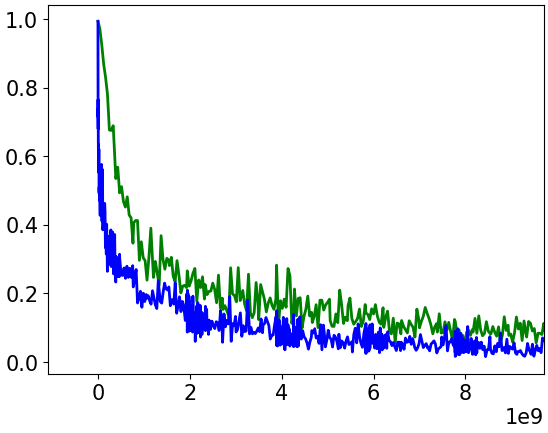} & 
        \hspace*{-3pt} \includegraphics[width=.16\textwidth]{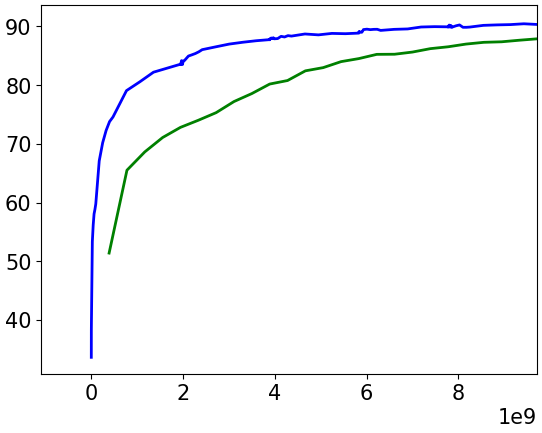} & 
        \hspace*{-3pt} \includegraphics[width=.16\textwidth]{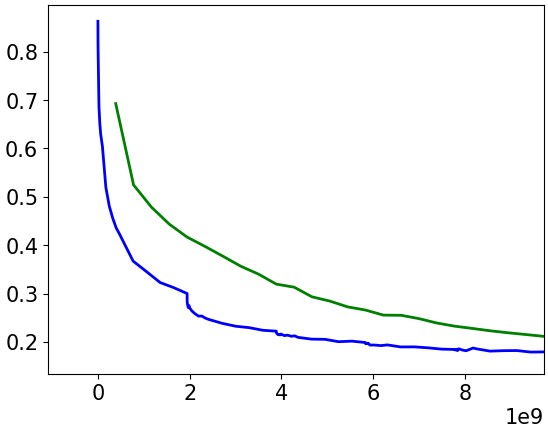} \\
        \hspace*{-3pt} \includegraphics[width=.16\textwidth]{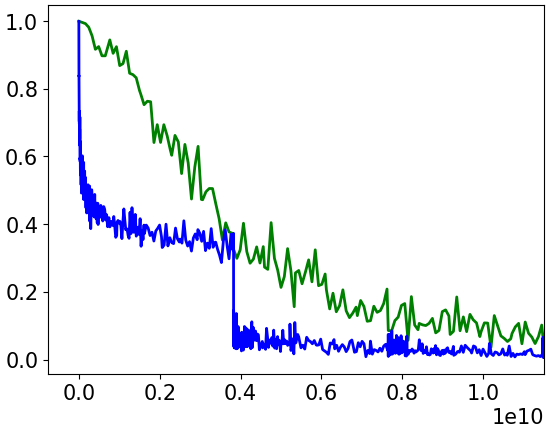} & 
        \hspace*{-3pt} \includegraphics[width=.16\textwidth]{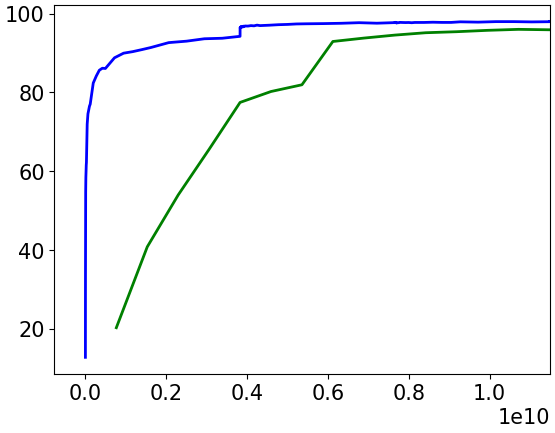} & 
        \hspace*{-3pt} \includegraphics[width=.16\textwidth]{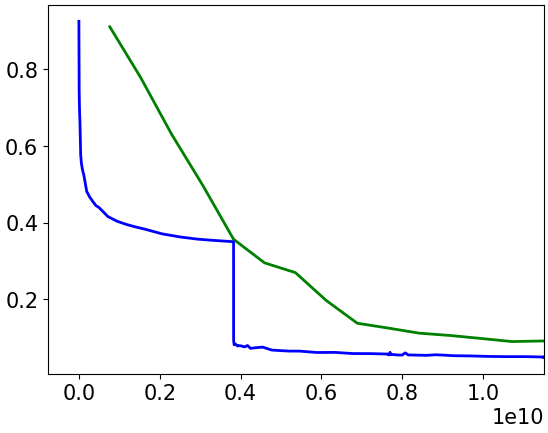} &
        \hspace*{-3pt} \includegraphics[width=.16\textwidth]{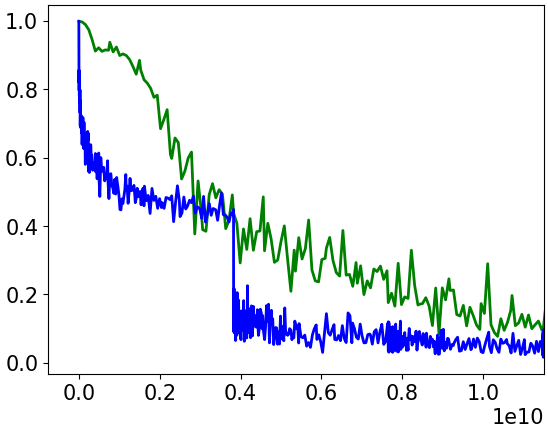} & 
        \hspace*{-3pt} \includegraphics[width=.16\textwidth]{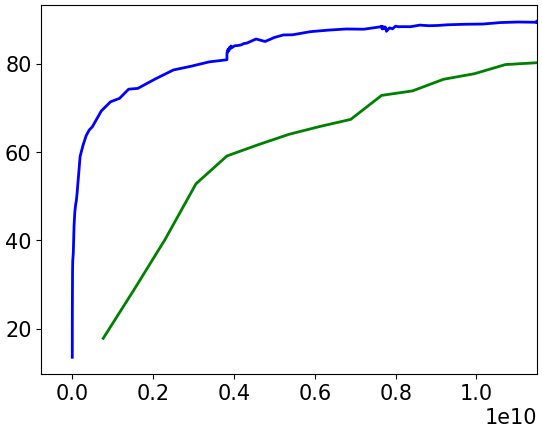} & 
        \hspace*{-3pt} \includegraphics[width=.16\textwidth]{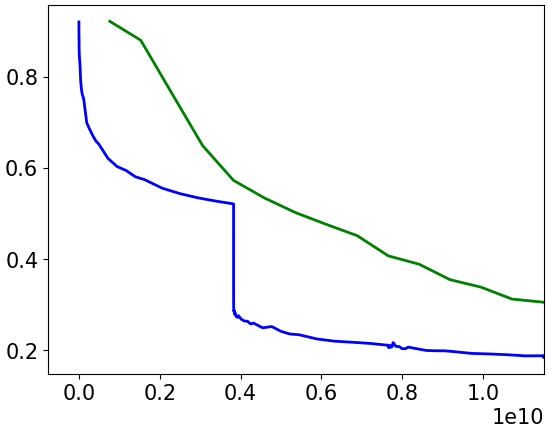} \\
        \end{tabular}
    \caption{\textit{(Best viewed in color)} Comparative Performance of SGD (Green) and DANTE (Blue) on MNIST and KMNIST datasets. The rows correspond to networks $(784 \xrightarrow{} 100 \xrightarrow{} 10)$, $(784 \xrightarrow{} 250 \xrightarrow{} 10)$, $(784 \xrightarrow{} 400 \xrightarrow{} 10)$, $(784 \xrightarrow{} 600 \xrightarrow{} 10)$, $(784 \xrightarrow{} 400 \xrightarrow{} 200 \xrightarrow{} 10)$, $(784 \xrightarrow{} 400 \xrightarrow{} 200 \xrightarrow{} 100 \xrightarrow{} 10)$ and $(784 \xrightarrow{}600 \xrightarrow{} 400 \xrightarrow{} 200 \xrightarrow{} 100 \xrightarrow{} 50 \xrightarrow{} 10)$ in order from top to bottom, all with Leaky ReLU activations. The first three columns correspond to MNIST, and the last three correspond to KMNIST. The first and fourth columns show training loss; second and fifth columns show test accuracy; third and sixth columns show test loss. For all the plots, X axis is the number of weights updated.}
    \label{tab:K/MNIST_l_MSE}
\end{figure*}

Both MNIST and KMNIST datasets consist of grayscale images of size $28 \times 28$. The input layer hence has dimension 784. For both these datasets, we use one-hidden layer networks having 100, 250, 400 and 600 neurons in the hidden layer, as well as a two hidden-layer network $(784 \xrightarrow{} 400 \xrightarrow{} 200 \xrightarrow{} 10)$, a three hidden-layer network $(784 \xrightarrow{} 400 \xrightarrow{} 200 \xrightarrow{} 100 \xrightarrow{} 10)$ and a five hidden-layer network $(784 \xrightarrow{}600 \xrightarrow{} 400 \xrightarrow{} 200 \xrightarrow{} 100 \xrightarrow{} 50 \xrightarrow{} 10)$. The results are presented in Figure \ref{tab:K/MNIST_l_MSE}.

\begin{figure*}[]
    \setlength\tabcolsep{0pt}
    \centering
    \begin{tabular}{cccccc}
        \hspace*{-3pt} \includegraphics[width=.16\textwidth]{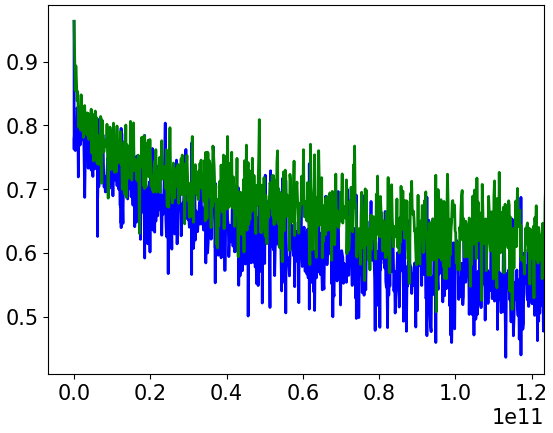} & 
        \hspace*{-3pt} \includegraphics[width=.16\textwidth]{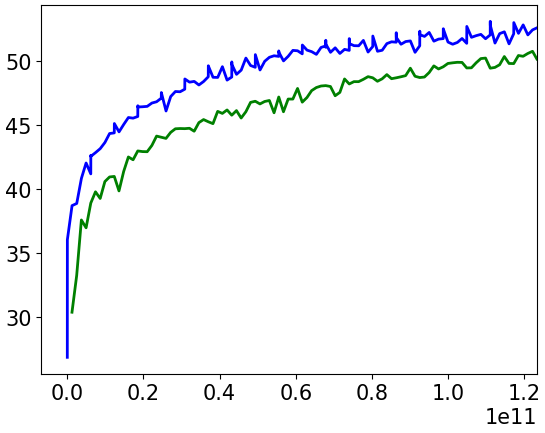} & 
        \hspace*{-3pt} \includegraphics[width=.16\textwidth]{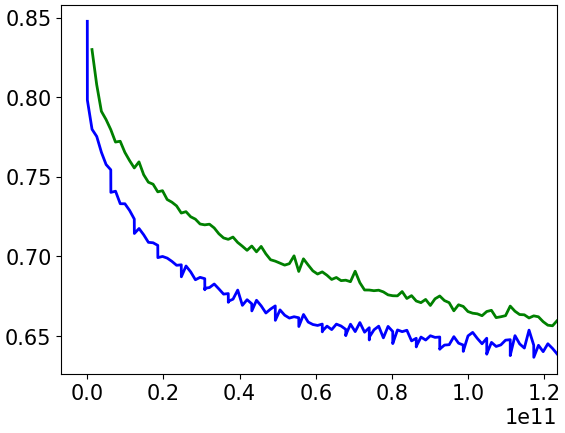} &
        \hspace*{-3pt} \includegraphics[width=.16\textwidth]{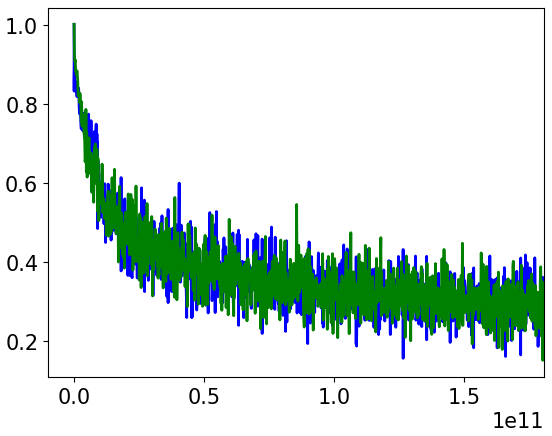} & 
        \hspace*{-3pt} \includegraphics[width=.16\textwidth]{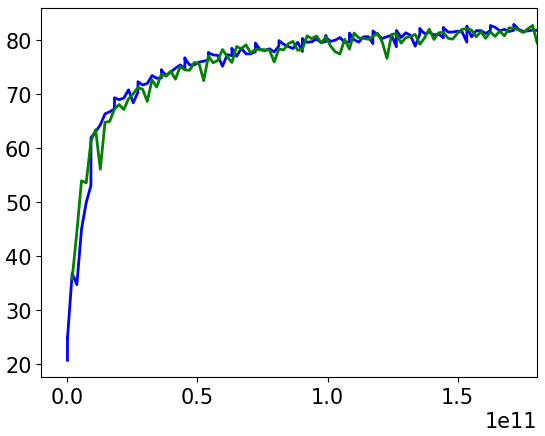} & 
        \hspace*{-3pt} \includegraphics[width=.16\textwidth]{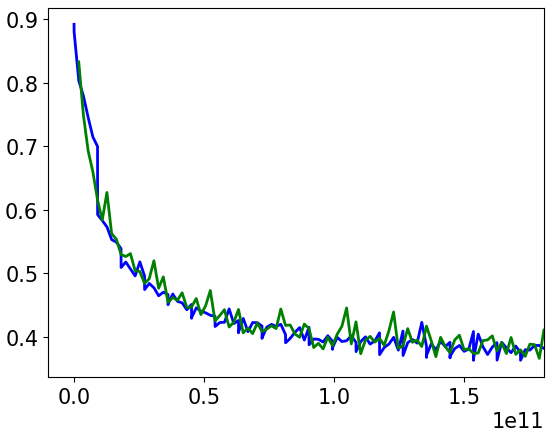} \\
        \hspace*{-3pt} \includegraphics[width=.16\textwidth]{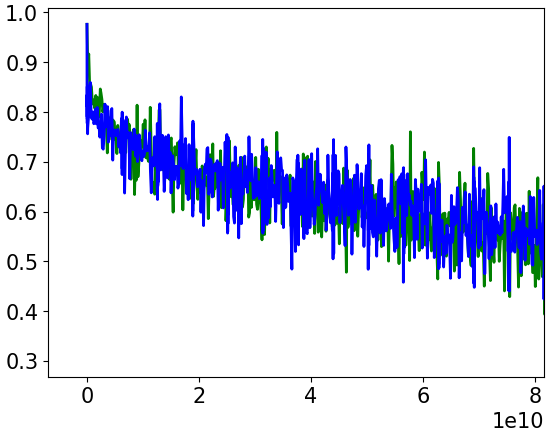} & 
        \hspace*{-3pt} \includegraphics[width=.16\textwidth]{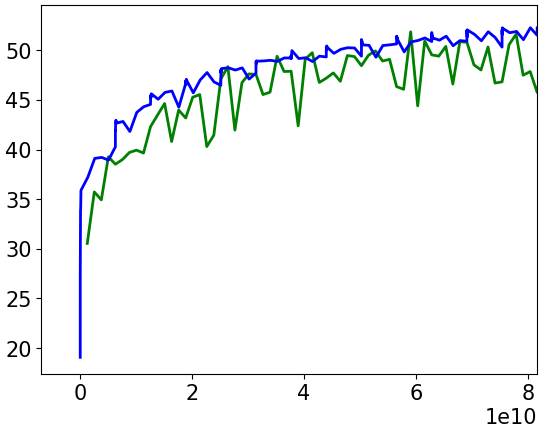} & 
        \hspace*{-3pt} \includegraphics[width=.16\textwidth]{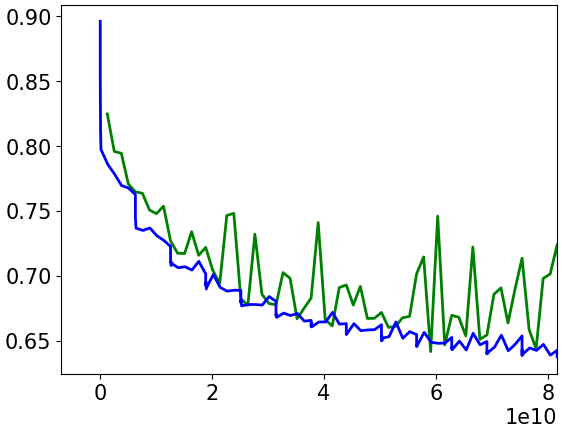} &
        \hspace*{-3pt} \includegraphics[width=.16\textwidth]{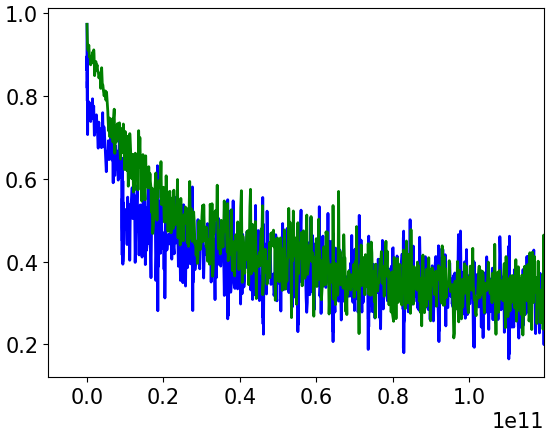} & 
        \hspace*{-3pt} \includegraphics[width=.16\textwidth]{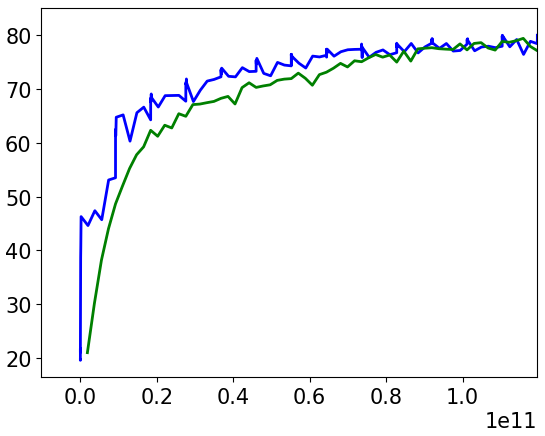} & 
        \hspace*{-3pt} \includegraphics[width=.16\textwidth]{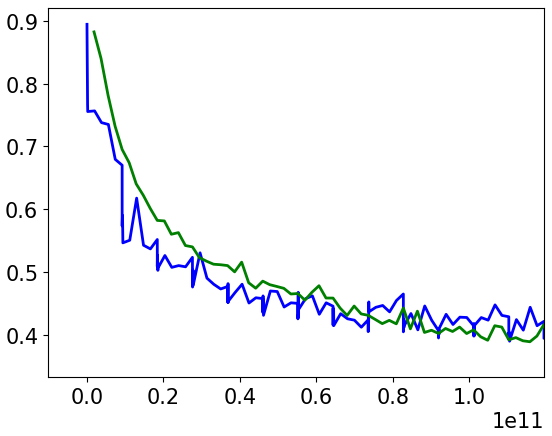} \\
        \hspace*{-3pt} \includegraphics[width=.16\textwidth]{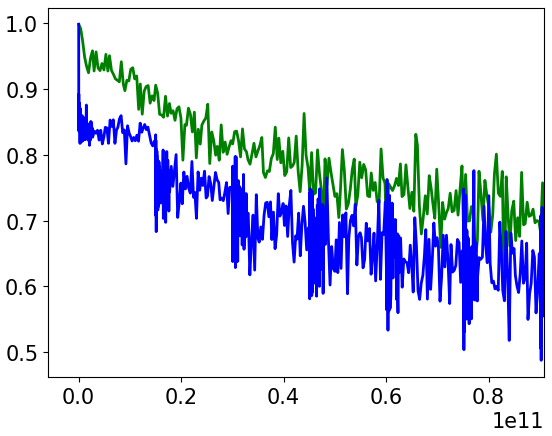} & 
        \hspace*{-3pt} \includegraphics[width=.16\textwidth]{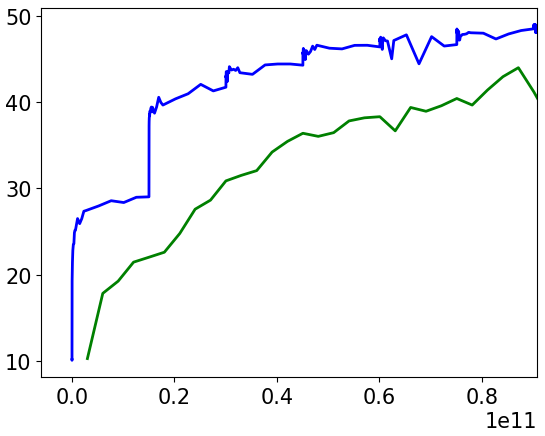} & 
        \hspace*{-3pt} \includegraphics[width=.16\textwidth]{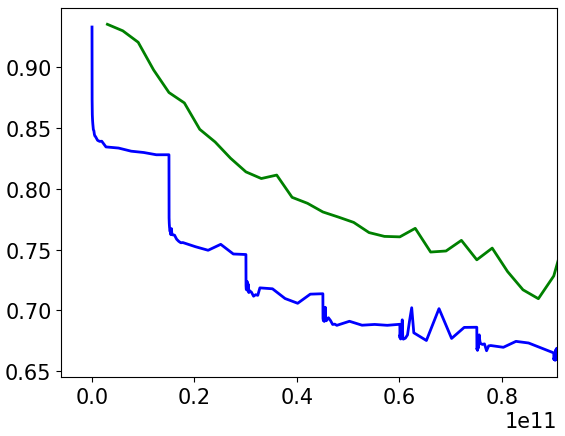} &
        \hspace*{-3pt} \includegraphics[width=.16\textwidth]{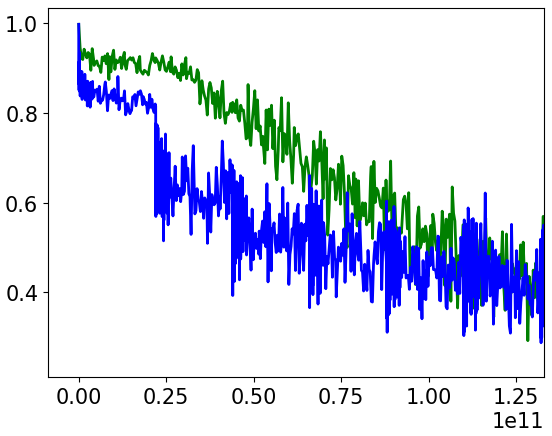} & 
        \hspace*{-3pt} \includegraphics[width=.16\textwidth]{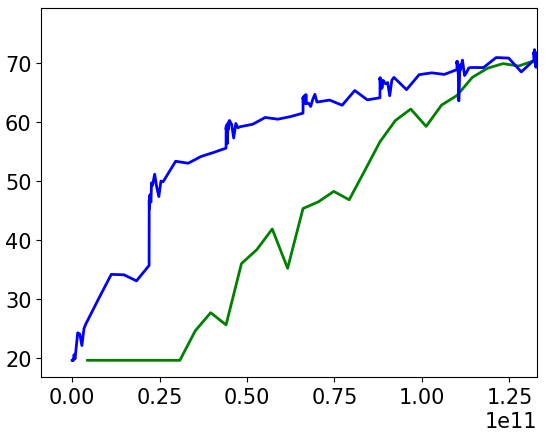} & 
        \hspace*{-3pt} \includegraphics[width=.16\textwidth]{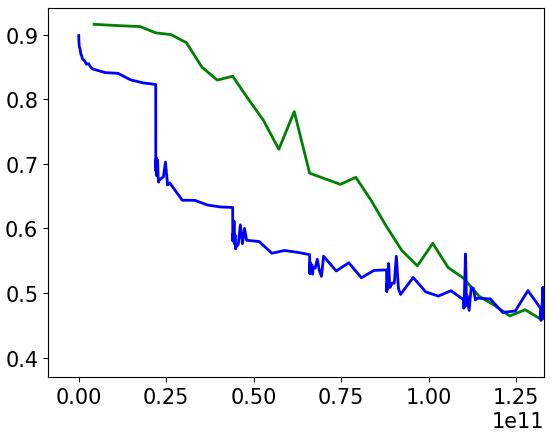} \\
    \end{tabular}
    \caption{\textit{(Best viewed in color)} Comparative performance of SGD (Green) and DANTE (Blue) on CIFAR-10 and SVHN datasets. The rows correspond to networks $(3072 \xrightarrow{} 512 \xrightarrow{} 10)$, $(3072 \xrightarrow{} 512 \xrightarrow{} 64 \xrightarrow{} 10)$ and $(3072 \xrightarrow{} 1024 \xrightarrow{} 512 \xrightarrow{} 256 \xrightarrow{} 128 \xrightarrow{} 64 \xrightarrow{} 10)$ in order from top to bottom, all with Leaky ReLU activations. The first three columns correspond to CIFAR-10, and the last three correspond to SVHN. The first and fourth columns show training loss; second and fifth columns show test accuracy; third and sixth columns show test loss. For all the plots, X axis is the number of weights updated.}
    \label{tab:CIFAR10-SVHN_l_MSE}
\end{figure*}

Both CIFAR-10 and SVHN datasets have colored (3-channel) images of size 3 x 32 x 32, thus the input layer for these is of dimension 3072. For these datasets, we use a one hidden-layer network $(3072 \xrightarrow{} 512 \xrightarrow{} 10)$, a two-hidden layer network $(3072 \xrightarrow{} 512 \xrightarrow{} 64 \xrightarrow{} 10)$. and a five hidden-layer network $(3072 \xrightarrow{} 1024 \xrightarrow{} 512 \xrightarrow{} 256 \xrightarrow{} 128 \xrightarrow{} 64 \xrightarrow{} 10)$. Figure \ref{tab:CIFAR10-SVHN_l_MSE} shows the results.

\begin{figure}[]
    \setlength\tabcolsep{0pt}
    \centering
    \begin{tabular}{ccc}
        \hspace*{-3pt} \includegraphics[width=.16\textwidth]{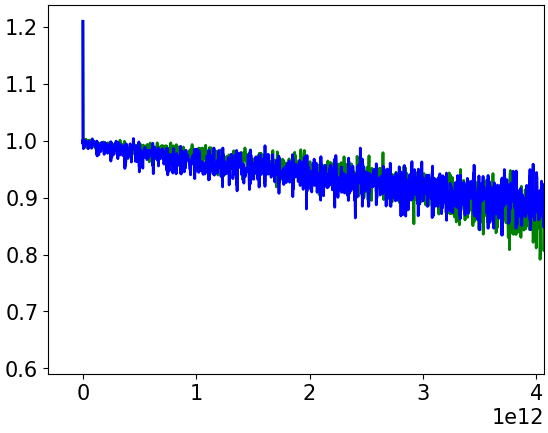} & 
        \hspace*{-3pt} \includegraphics[width=.16\textwidth]{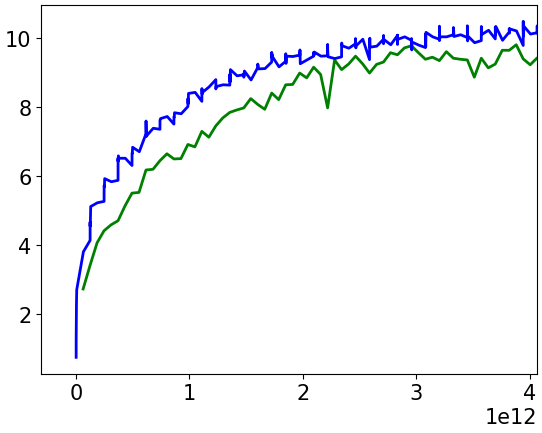} & 
        \hspace*{-3pt} \includegraphics[width=.16\textwidth]{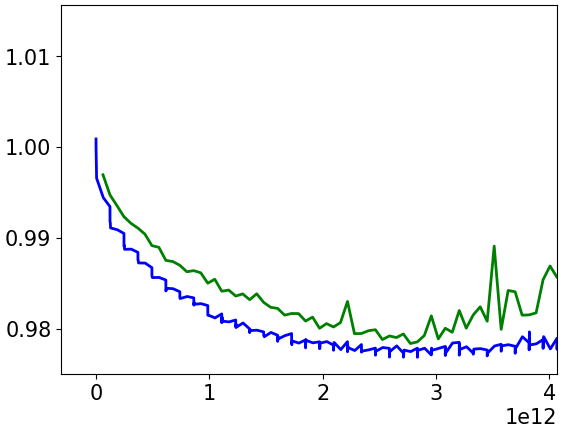} \\
        \hspace*{-3pt} \includegraphics[width=.16\textwidth]{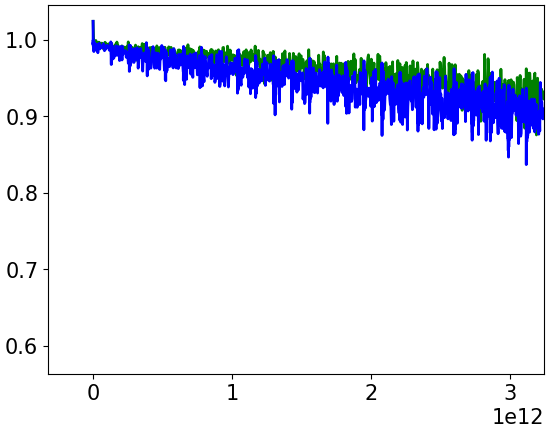} & 
        \hspace*{-3pt} \includegraphics[width=.16\textwidth]{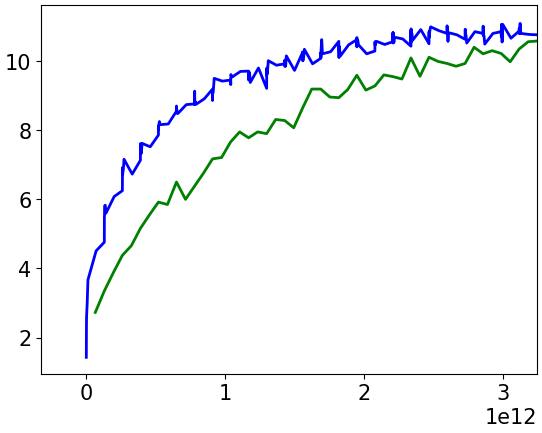} & 
        \hspace*{-3pt} \includegraphics[width=.16\textwidth]{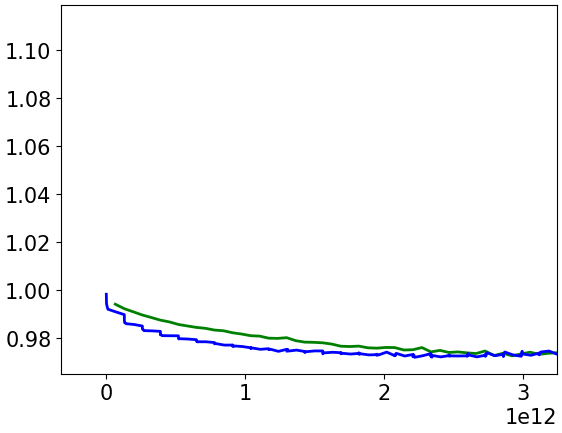} \\
        \hspace*{-3pt} \includegraphics[width=.16\textwidth]{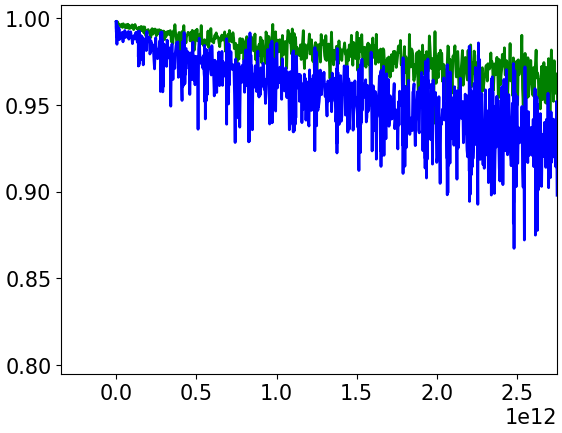} & 
        \hspace*{-3pt} \includegraphics[width=.16\textwidth]{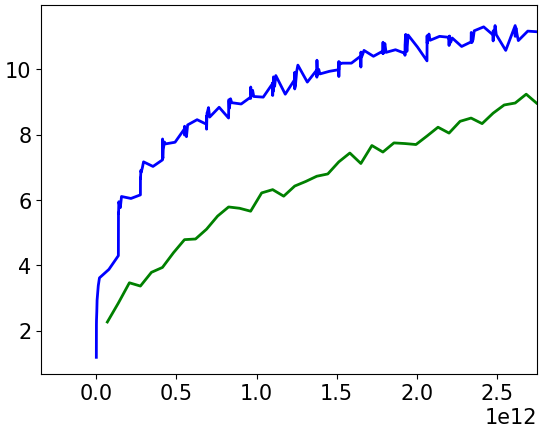} & 
        \hspace*{-3pt} \includegraphics[width=.16\textwidth]{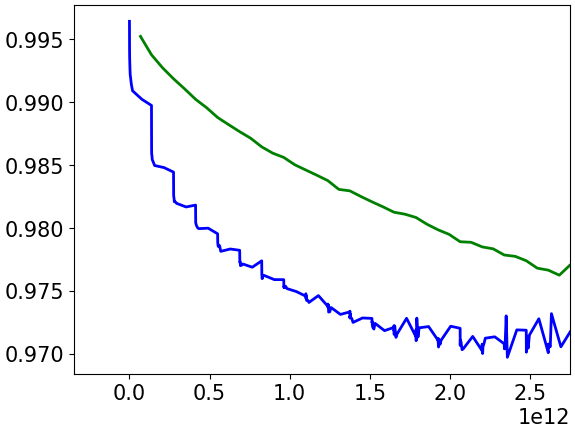} \\
    \end{tabular}
    \caption{\textit{(Best viewed in color)} Comparative performance of SGD (Green) and DANTE (Blue) on Tiny Imagenet. The rows correspond to networks $(12288 \xrightarrow{} 3072 \xrightarrow{} 512 \xrightarrow{} 200)$, $(12288 \xrightarrow{} 3072 \xrightarrow{} 1024 \xrightarrow{} 512 \xrightarrow{} 200)$ and $(12288 \xrightarrow{} 3072 \xrightarrow{} 1536 \xrightarrow{} 768 \xrightarrow{} 384 \xrightarrow{} 200)$ in order from top to bottom, all with Leaky ReLU activations. The first column shows training loss, second shows validation accuracy and third shows validation loss. For all plots, X-axis is the number of weights updated.}
    \label{tab:TinyImgNet_l_MSE}
\end{figure}

Tiny-Imagenet is a widely used subset of the original Imagenet dataset having 200 classes, and 500 images of each class in the training set. We train three networks ($12288 \xrightarrow{} 3072 \xrightarrow{} 512 \xrightarrow{} 200$, $12288 \xrightarrow{} 3072 \xrightarrow{} 1024 \xrightarrow{} 512 \xrightarrow{} 200$, and $12288 \xrightarrow{} 3072 \xrightarrow{} 1536 \xrightarrow{} 768 \xrightarrow{} 384 \xrightarrow{} 200$) using DANTE and SGD on this dataset to compare the performance on a more difficult task. Since the labels of the test set are not available, we report the performance on the standard validation set (which has 50 images of each class) in Figure \ref{tab:TinyImgNet_l_MSE}. (Both SGD and DANTE do not achieve high accuracies on this dataset as the network considered is a simple MLP. Considering our objective in this work was to prove the feasibility of this approach with MLPs, studying extension of DANTE on convolutional layers, LSTMs and other variants are important directions of our future work.)

The results clearly show the effectiveness of using \myalgo for training neural networks - \myalgo obtains lower training/test loss and higher test accuracy. Even in cases where the final losses of DANTE and SGD are almost equal, DANTE always minimizes the loss faster than SGD. 

\begin{figure*}[]
    \setlength\tabcolsep{0pt}
    \centering
    \begin{tabular}{cccccc}
        \hspace*{-5pt} \includegraphics[width=.16\textwidth]{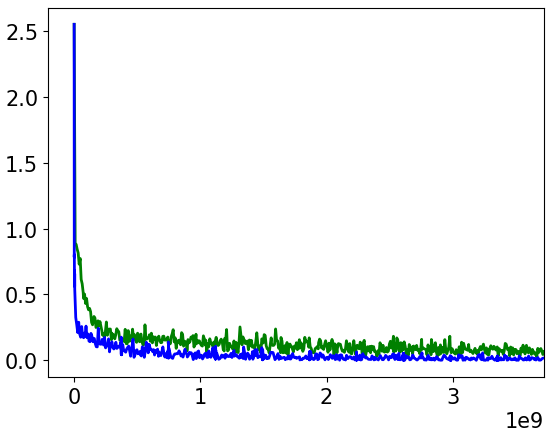} & 
        \hspace*{-3pt} \includegraphics[width=.16\textwidth]{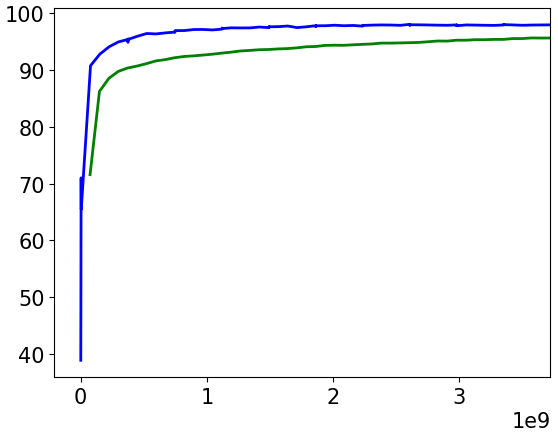} & 
        \hspace*{-3pt} \includegraphics[width=.16\textwidth]{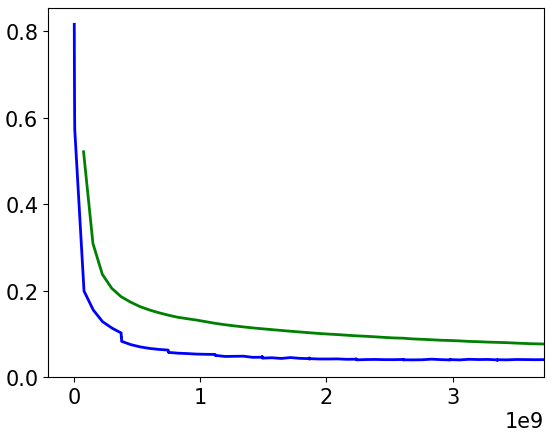} &
        \hspace*{-3pt} \includegraphics[width=.16\textwidth]{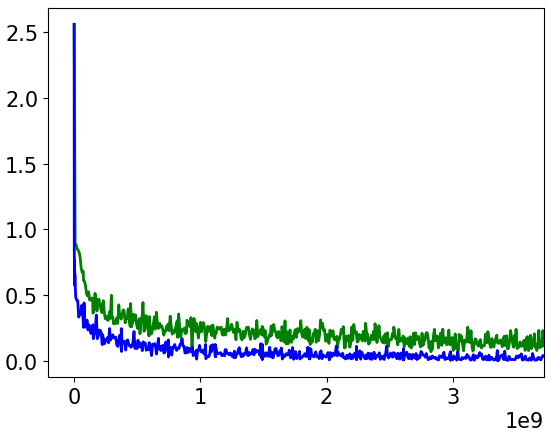} & 
        \hspace*{-3pt} \includegraphics[width=.16\textwidth]{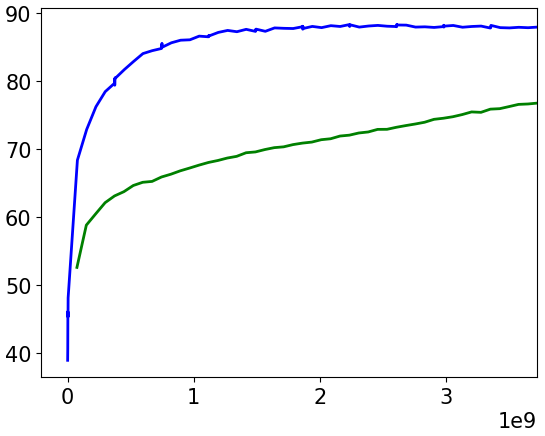} & 
        \hspace*{-3pt} \includegraphics[width=.16\textwidth]{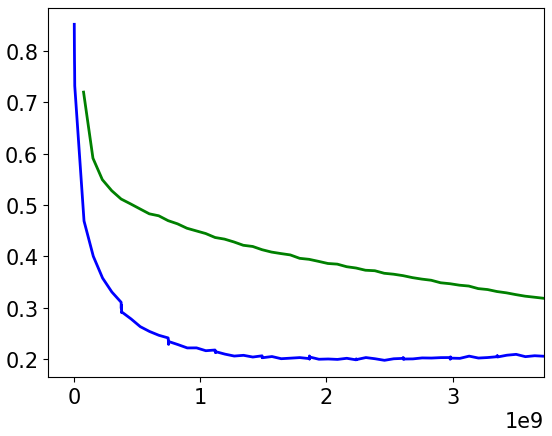} \\
        \hspace*{-5pt} \includegraphics[width=.16\textwidth]{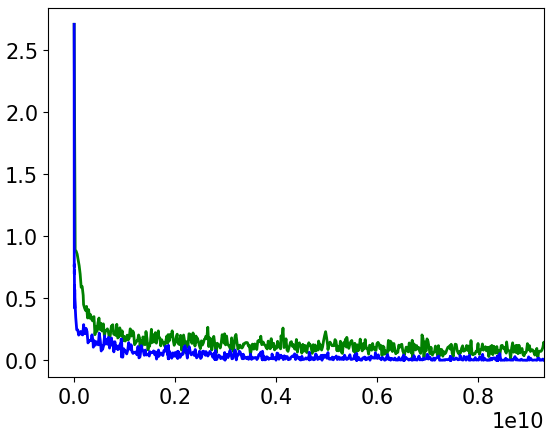} & 
        \hspace*{-3pt} \includegraphics[width=.16\textwidth]{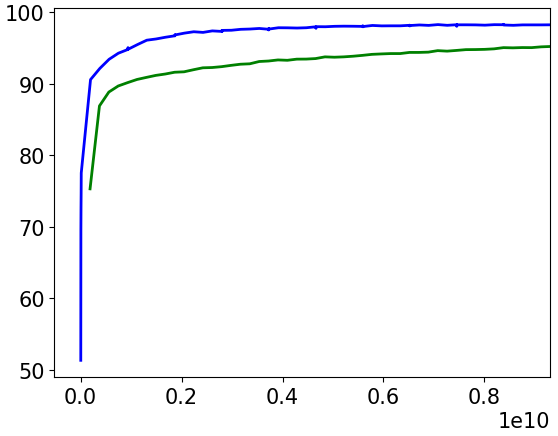} & 
        \hspace*{-3pt} \includegraphics[width=.16\textwidth]{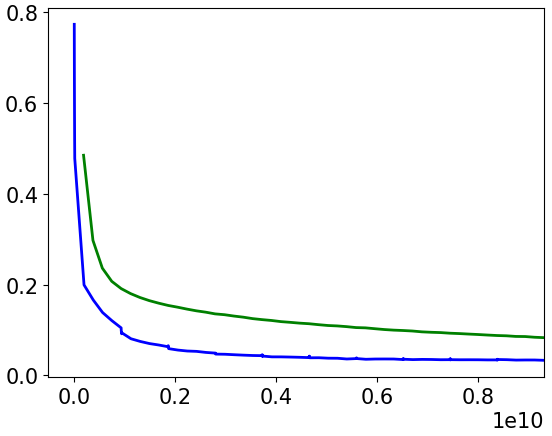} &
        \hspace*{-3pt} \includegraphics[width=.16\textwidth]{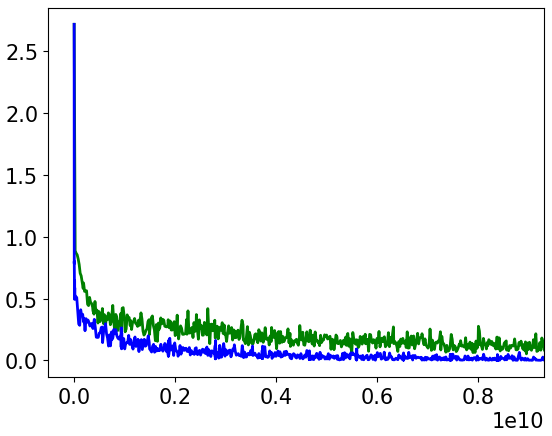} & 
        \hspace*{-3pt} \includegraphics[width=.16\textwidth]{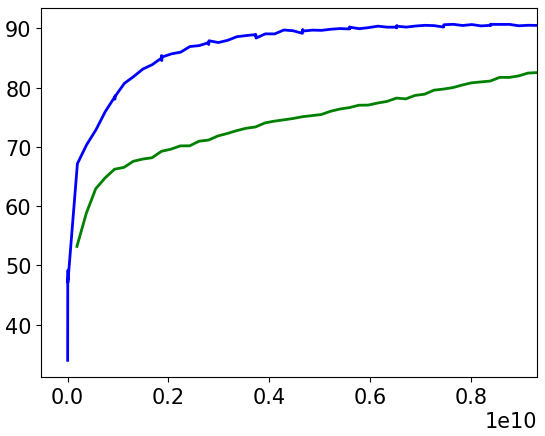} & 
        \hspace*{-3pt} \includegraphics[width=.16\textwidth]{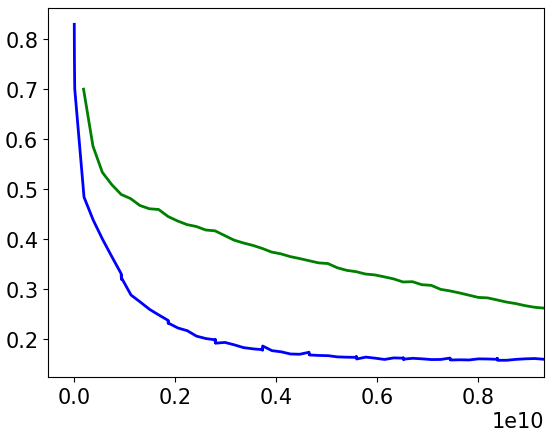} \\
        \hspace*{-5pt} \includegraphics[width=.16\textwidth]{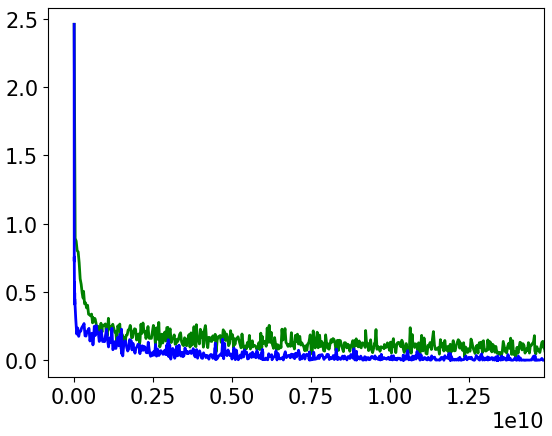} & 
        \hspace*{-3pt} \includegraphics[width=.16\textwidth]{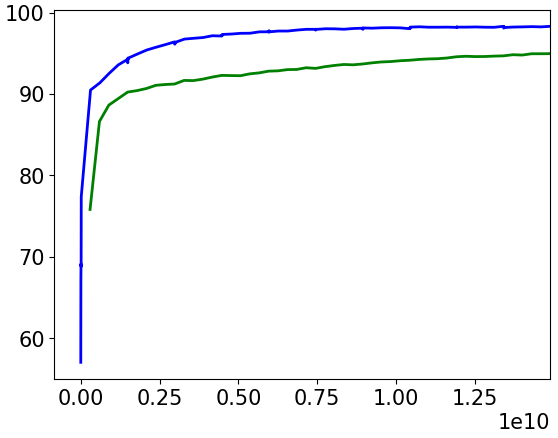} & 
        \hspace*{-3pt} \includegraphics[width=.16\textwidth]{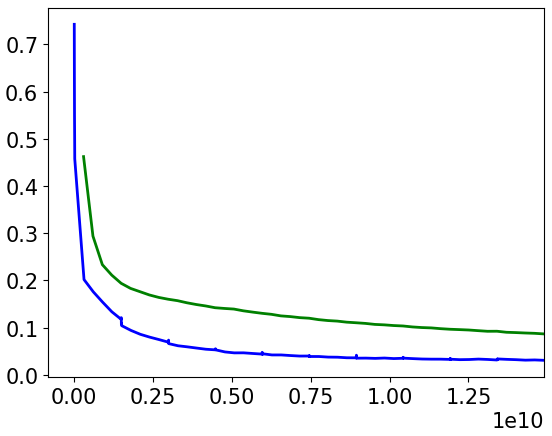} &
        \hspace*{-3pt} \includegraphics[width=.16\textwidth]{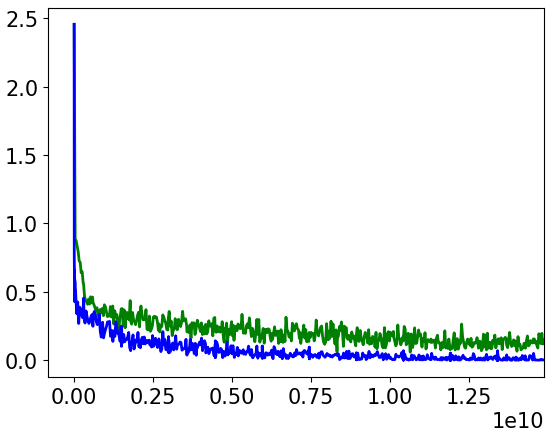} & 
        \hspace*{-3pt} \includegraphics[width=.16\textwidth]{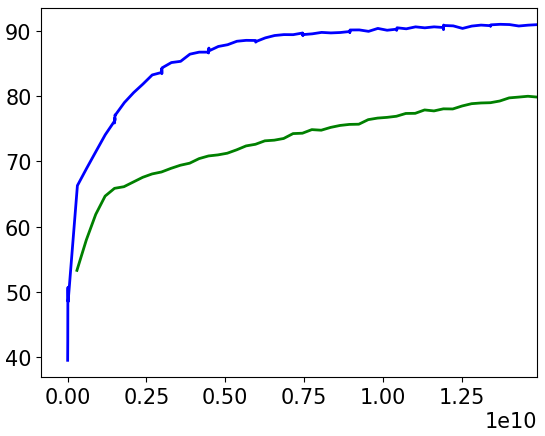} & 
        \hspace*{-3pt} \includegraphics[width=.16\textwidth]{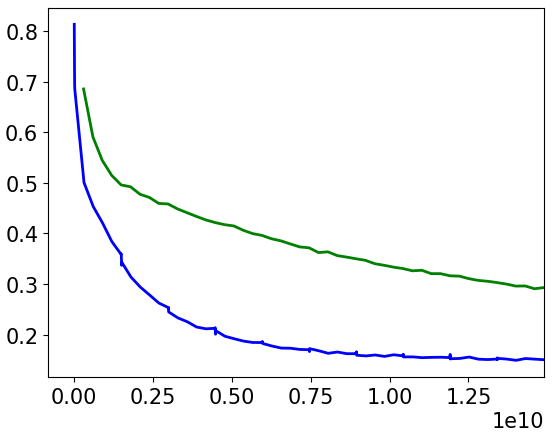} \\
        \hspace*{-5pt} \includegraphics[width=.16\textwidth]{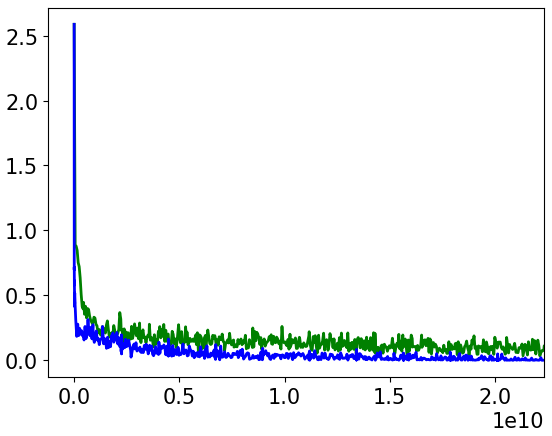} & 
        \hspace*{-3pt} \includegraphics[width=.16\textwidth]{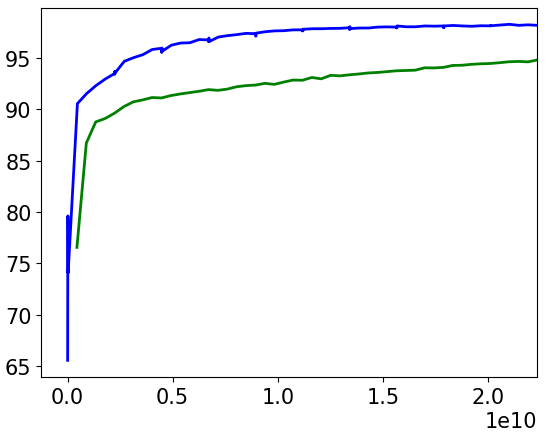} & 
        \hspace*{-3pt} \includegraphics[width=.16\textwidth]{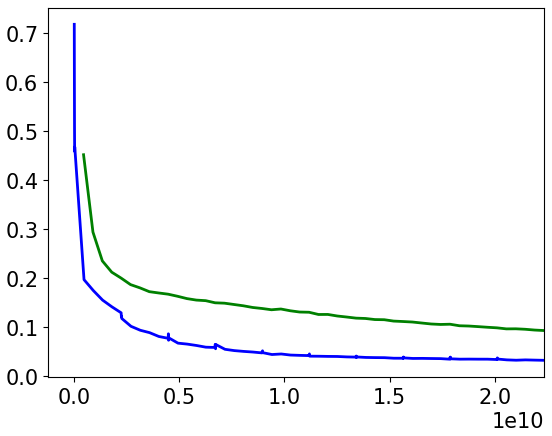} &
        \hspace*{-3pt} \includegraphics[width=.16\textwidth]{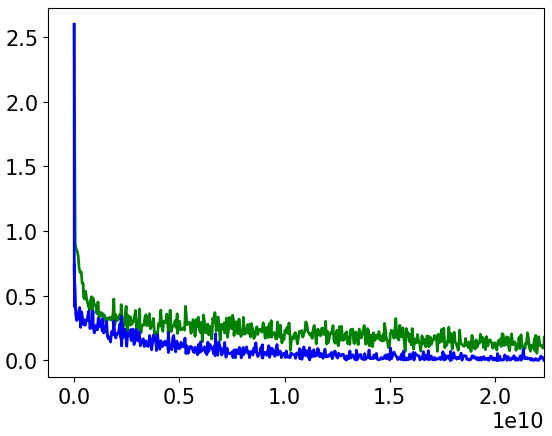} & 
        \hspace*{-3pt} \includegraphics[width=.16\textwidth]{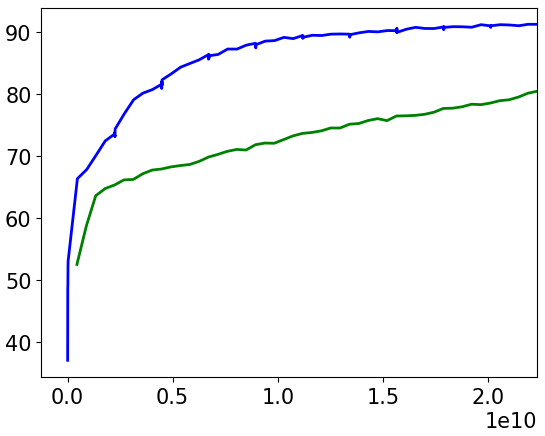} & 
        \hspace*{-3pt} \includegraphics[width=.16\textwidth]{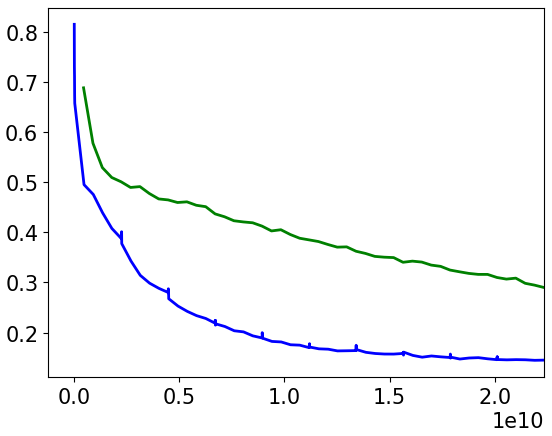} \\
        \hspace*{-5pt} \includegraphics[width=.16\textwidth]{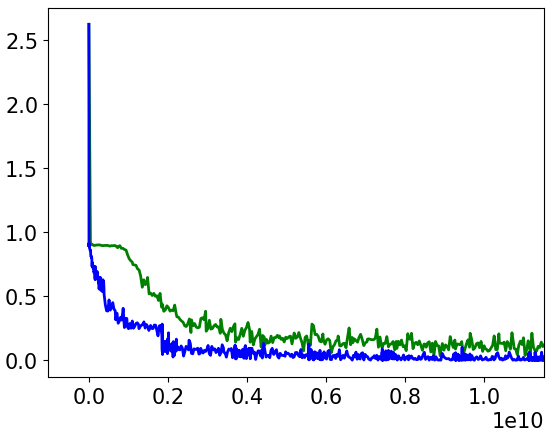} & 
        \hspace*{-3pt} \includegraphics[width=.16\textwidth]{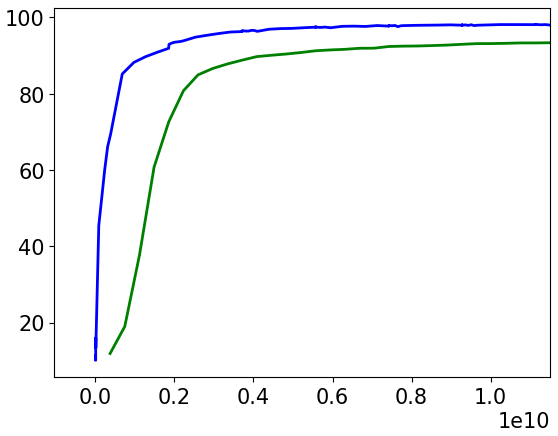} & 
        \hspace*{-3pt} \includegraphics[width=.16\textwidth]{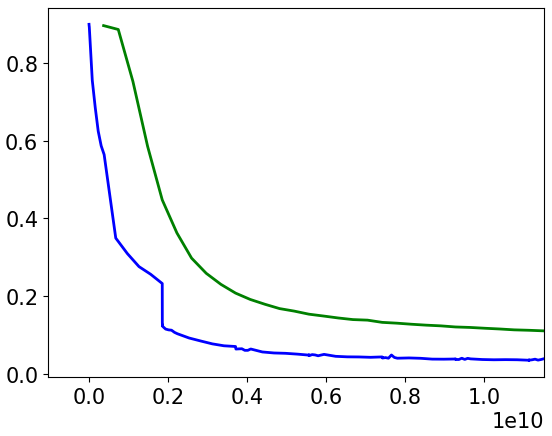}&
        \hspace*{-3pt} \includegraphics[width=.16\textwidth]{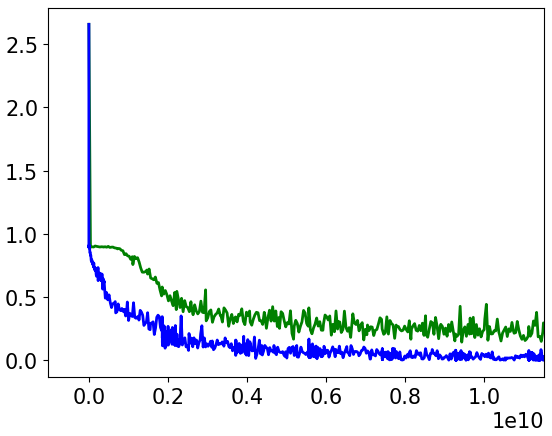} & 
        \hspace*{-3pt} \includegraphics[width=.16\textwidth]{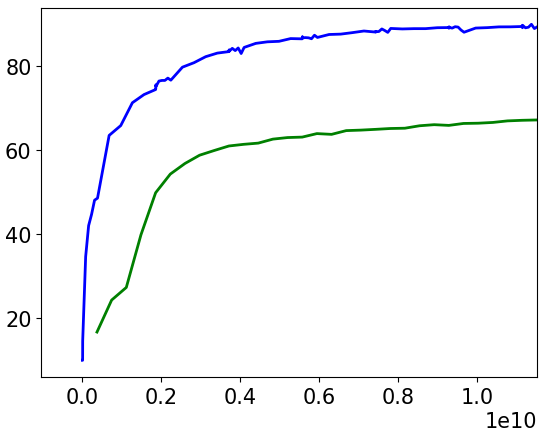} & 
        \hspace*{-3pt} \includegraphics[width=.16\textwidth]{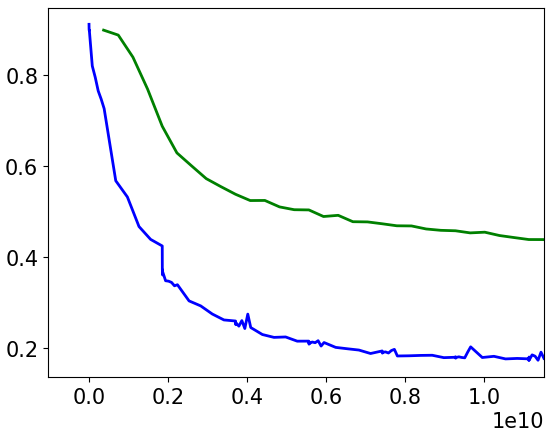} \\
        \hspace*{-5pt} \includegraphics[width=.16\textwidth]{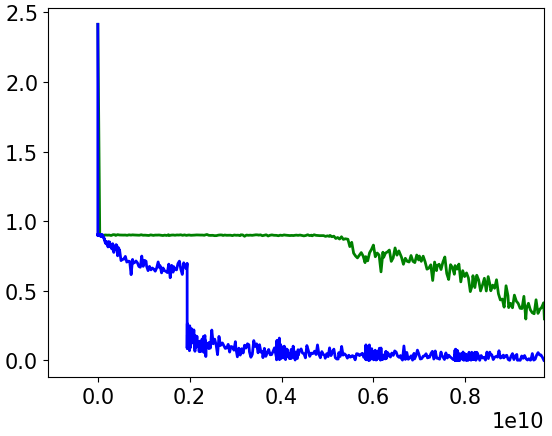} & 
        \hspace*{-3pt} \includegraphics[width=.16\textwidth]{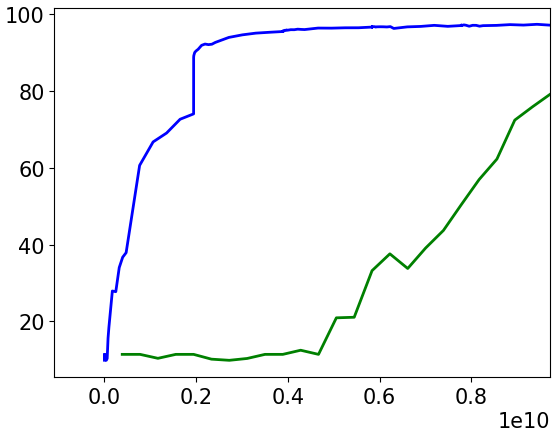} & 
        \hspace*{-3pt} \includegraphics[width=.16\textwidth]{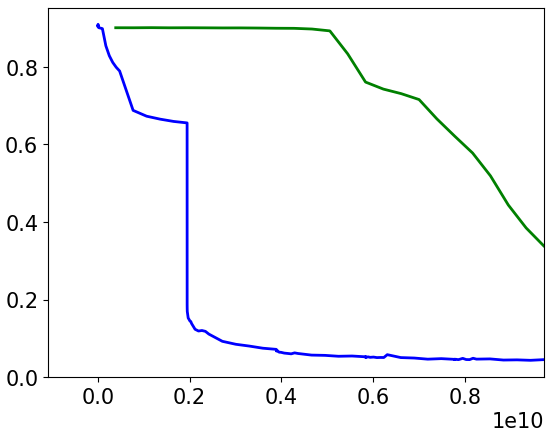} &
        \hspace*{-3pt} \includegraphics[width=.16\textwidth]{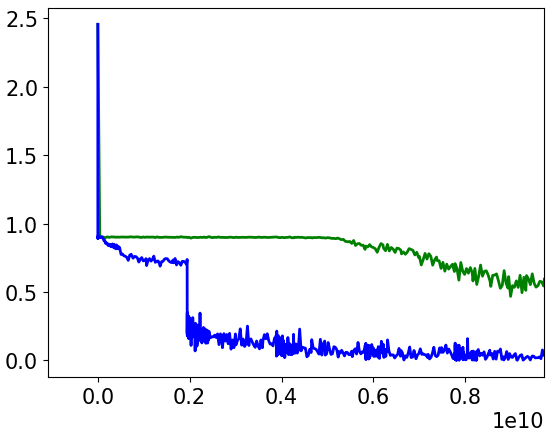} & 
        \hspace*{-3pt} \includegraphics[width=.16\textwidth]{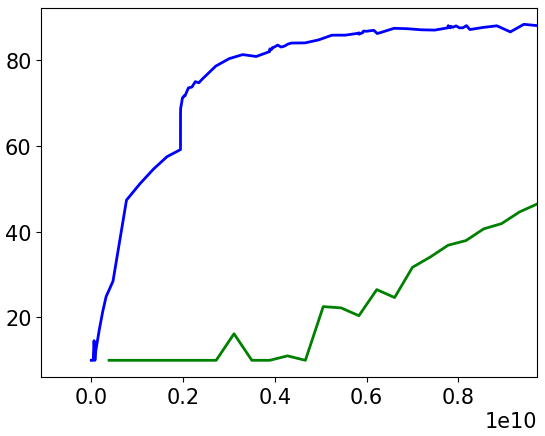} & 
        \hspace*{-3pt} \includegraphics[width=.16\textwidth]{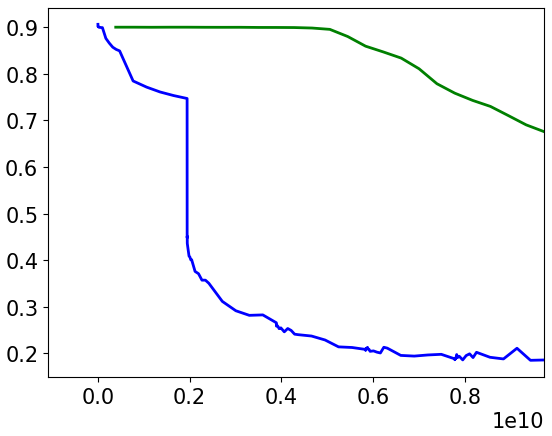} \\
    \end{tabular}
    \caption{\textit{(Best viewed in color)} Comparative performance of SGD (Green) and DANTE (Blue) on MNIST and KMNIST datasets. The rows correspond to networks (in order) $(784 \xrightarrow{} 100 \xrightarrow{} 10)$, $(784 \xrightarrow{} 250 \xrightarrow{} 10)$, $(784 \xrightarrow{} 400 \xrightarrow{} 10)$, $(784 \xrightarrow{} 600 \xrightarrow{} 10)$, $(784 \xrightarrow{} 400 \xrightarrow{} 200 \xrightarrow{} 10)$ and $(784 \xrightarrow{} 400 \xrightarrow{} 200 \xrightarrow{} 100 \xrightarrow{} 10)$ having sigmoid activations. The first three columns correspond to MNIST, and the last three correspond to KMNIST. The first and fourth columns show training loss; second and fifth columns show test accuracy; third and sixth columns show test loss. For all the plots, X-axis is the number of weights updated.}
    \label{tab:K/MNIST_s_MSE}
\end{figure*}

\noindent \textbf{Feedforward Neural Networks with Sigmoid Activations.} We now present the comparative performance of SGD and \myalgo on feedforward neural networks with sigmoid activations and Mean Square Error loss function. We show our results with MNIST and KMNIST datasets. We use the same architectures as in the previous subsection, except that we use sigmoid activation instead of Leaky ReLU. The results are presented in Figure \ref{tab:K/MNIST_s_MSE}. It is apparent from the results that DANTE performs better than SGD in this case too.

\begin{figure*}[]
    \setlength\tabcolsep{0pt}
    \centering
    \begin{tabular}{cccccc}
        \hspace*{-3pt} \includegraphics[width=.16\textwidth]{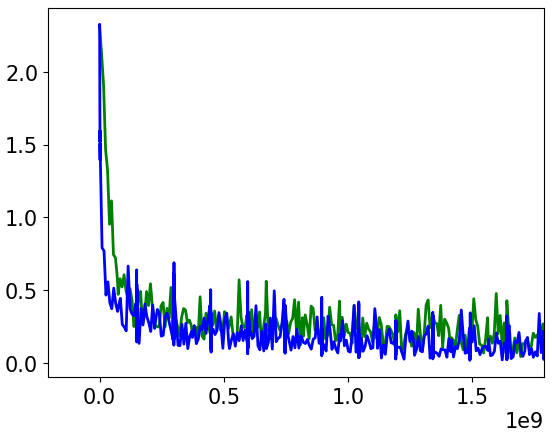} & 
        \hspace*{-3pt} \includegraphics[width=.16\textwidth]{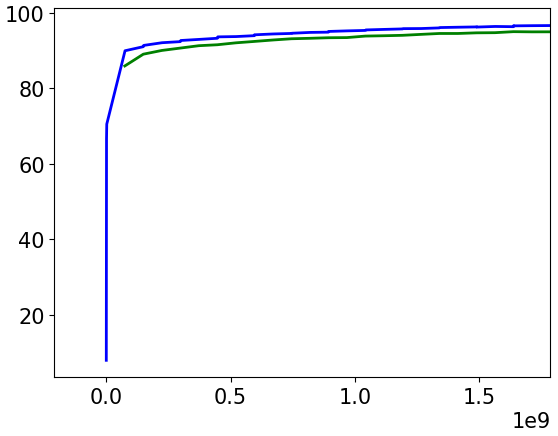} & 
        \hspace*{-3pt} \includegraphics[width=.16\textwidth]{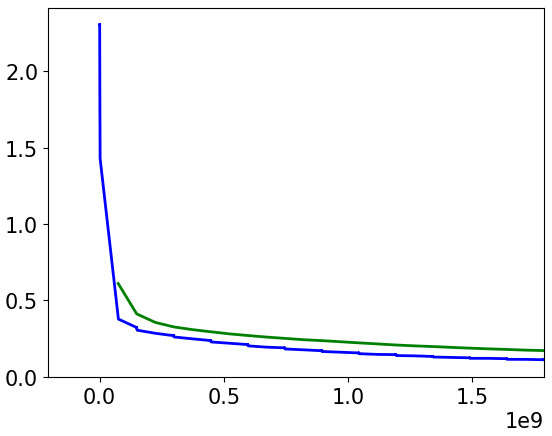} &
        \hspace*{-3pt} \includegraphics[width=.16\textwidth]{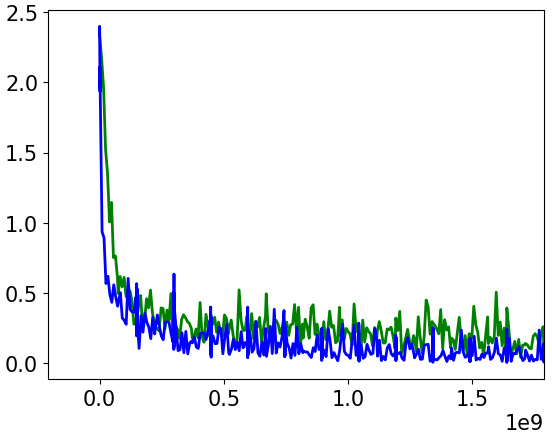} & 
        \hspace*{-3pt} \includegraphics[width=.16\textwidth]{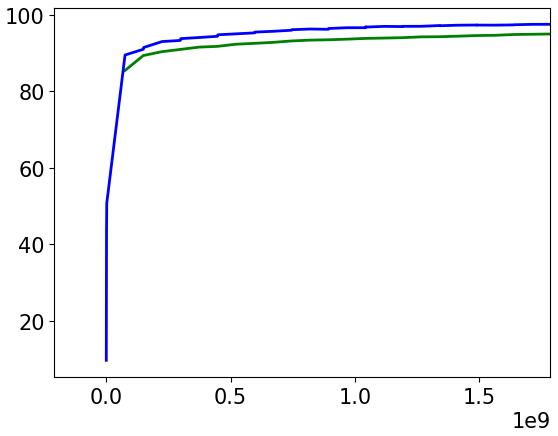} & 
        \hspace*{-3pt} \includegraphics[width=.16\textwidth]{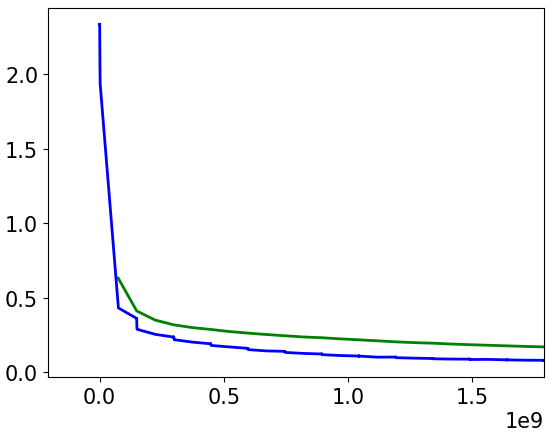} \\
        \hspace*{-3pt} \includegraphics[width=.16\textwidth]{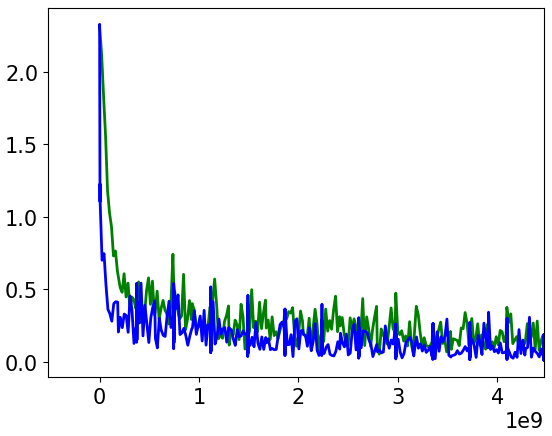} & 
        \hspace*{-3pt} \includegraphics[width=.16\textwidth]{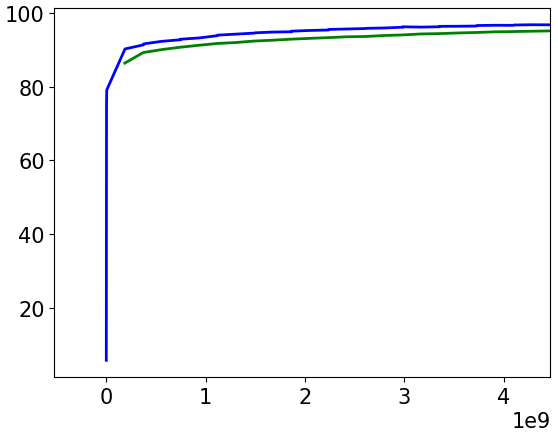} & 
        \hspace*{-3pt} \includegraphics[width=.16\textwidth]{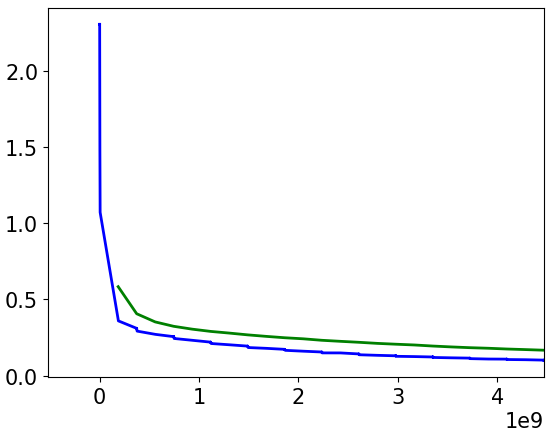} &
        \hspace*{-3pt} \includegraphics[width=.16\textwidth]{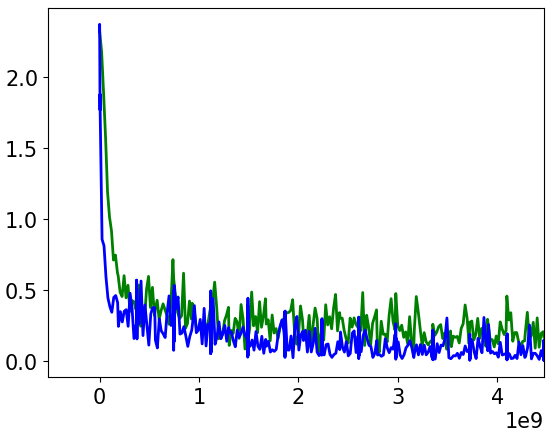} & 
        \hspace*{-3pt} \includegraphics[width=.16\textwidth]{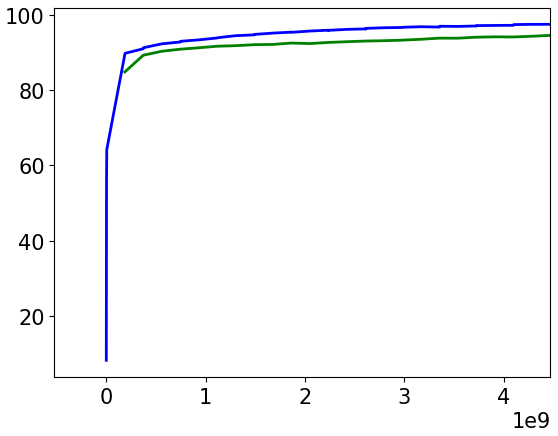} & 
        \hspace*{-3pt} \includegraphics[width=.16\textwidth]{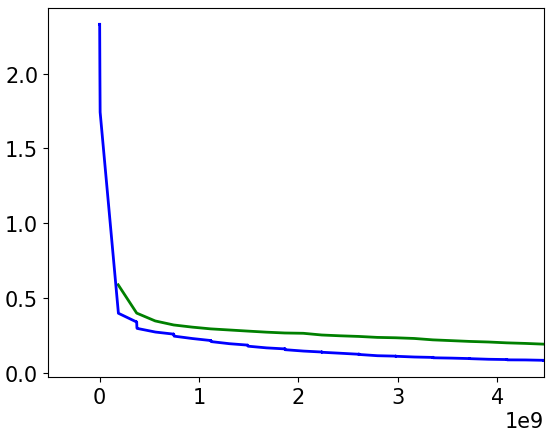} \\
        \hspace*{-3pt} \includegraphics[width=.16\textwidth]{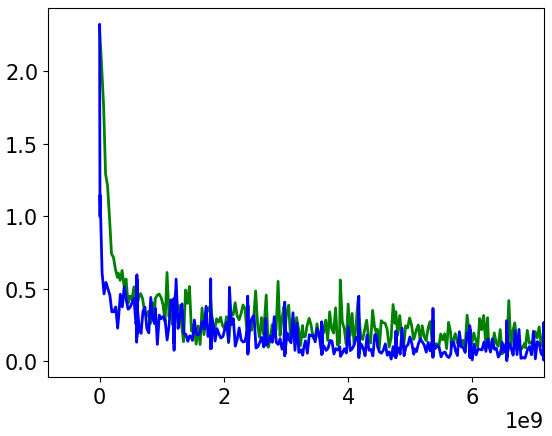} & 
        \hspace*{-3pt} \includegraphics[width=.16\textwidth]{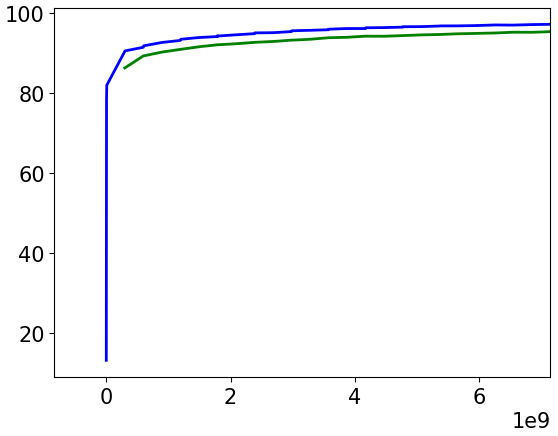} & 
        \hspace*{-3pt} \includegraphics[width=.16\textwidth]{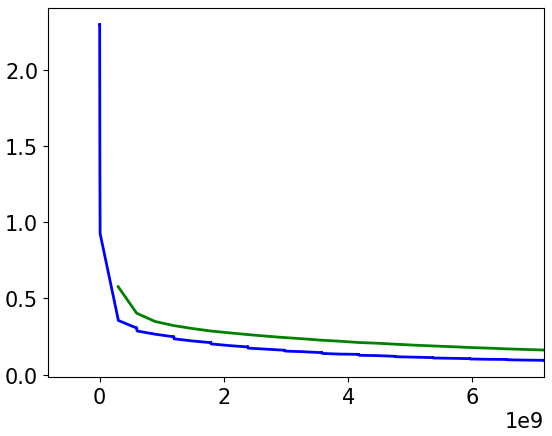} &
        \hspace*{-3pt} \includegraphics[width=.16\textwidth]{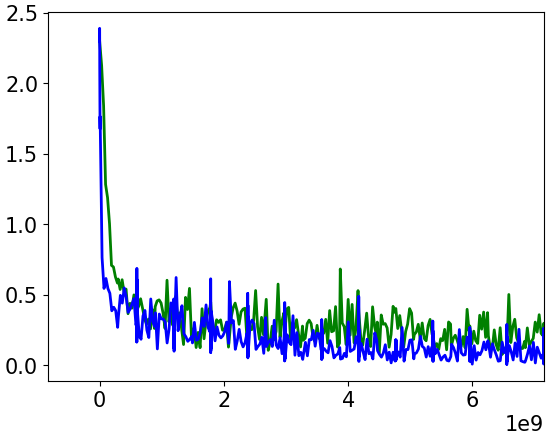} & 
        \hspace*{-3pt} \includegraphics[width=.16\textwidth]{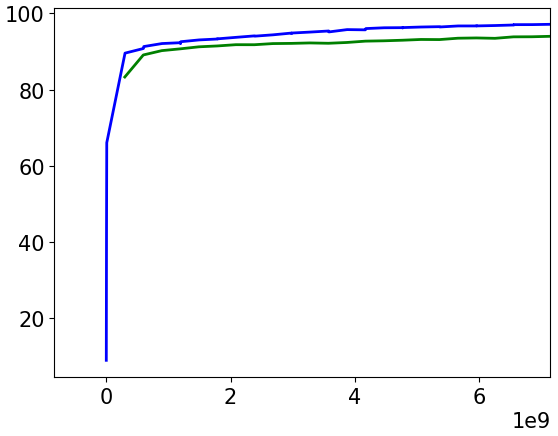} & 
        \hspace*{-3pt} \includegraphics[width=.16\textwidth]{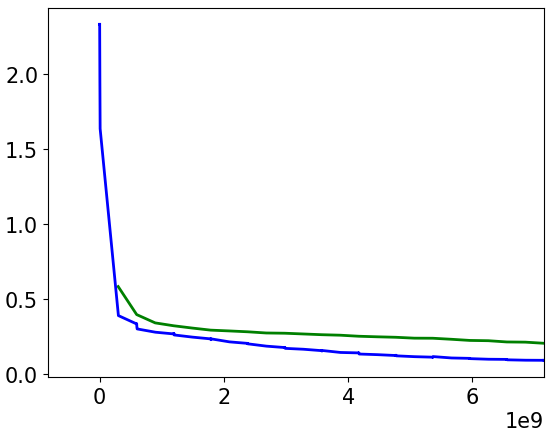} \\
        \hspace*{-3pt} \includegraphics[width=.16\textwidth]{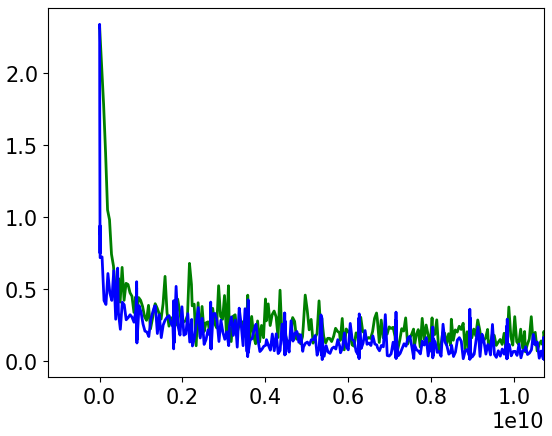} & 
        \hspace*{-3pt} \includegraphics[width=.16\textwidth]{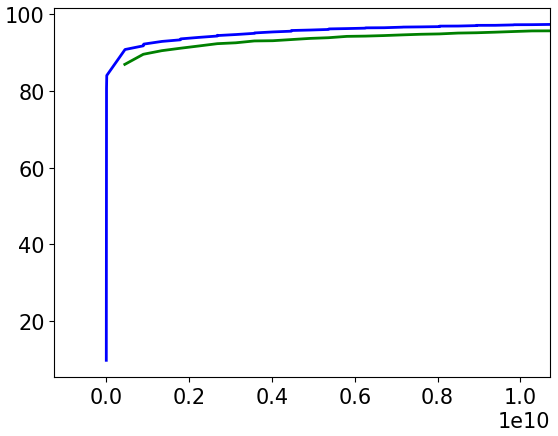} & 
        \hspace*{-3pt} \includegraphics[width=.16\textwidth]{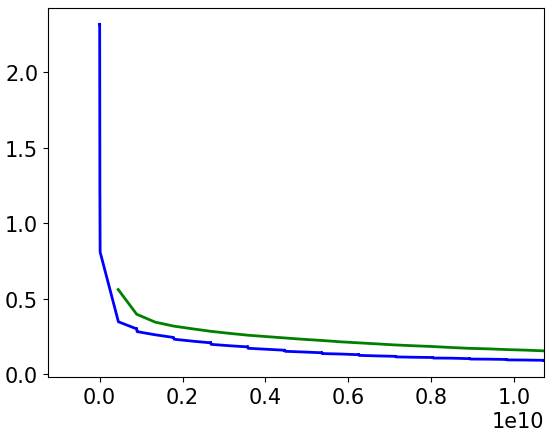} &
        \hspace*{-3pt} \includegraphics[width=.16\textwidth]{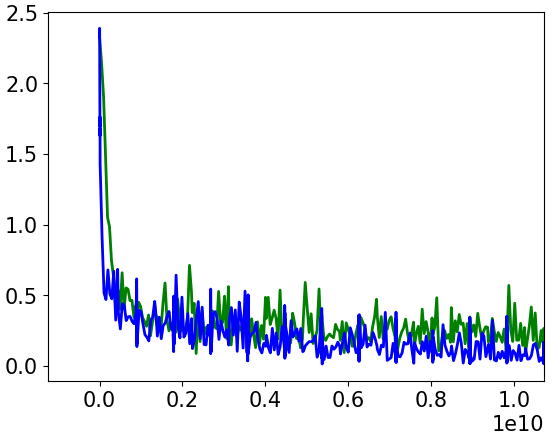} & 
        \hspace*{-3pt} \includegraphics[width=.16\textwidth]{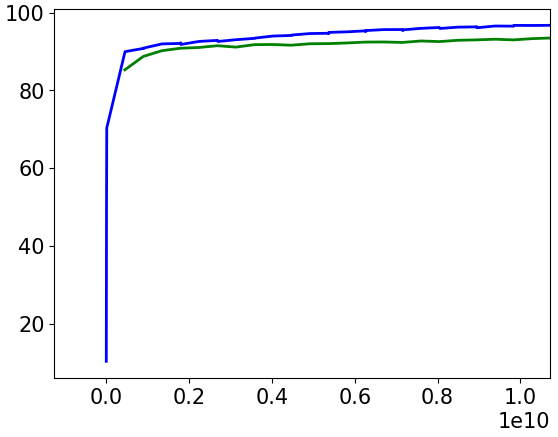} & 
        \hspace*{-3pt} \includegraphics[width=.16\textwidth]{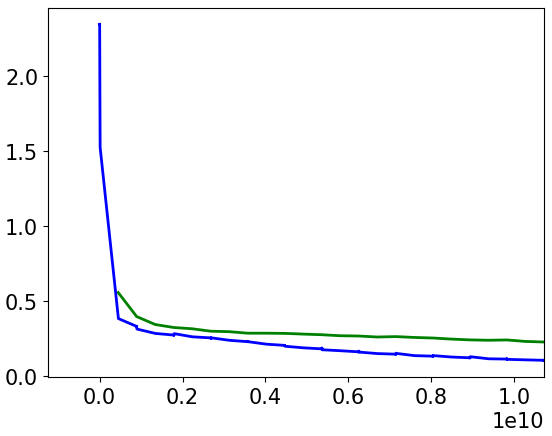} \\
        \hspace*{-3pt} \includegraphics[width=.16\textwidth]{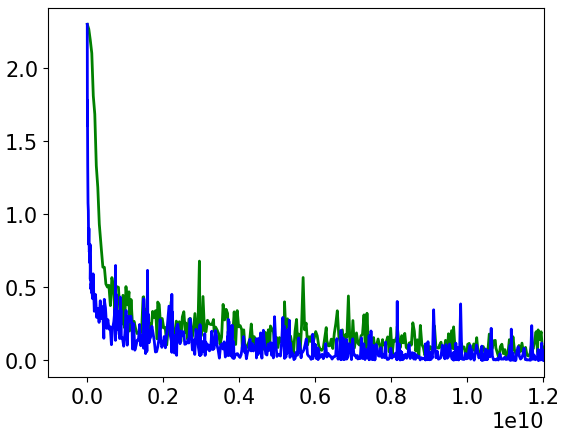} & 
        \hspace*{-3pt} \includegraphics[width=.16\textwidth]{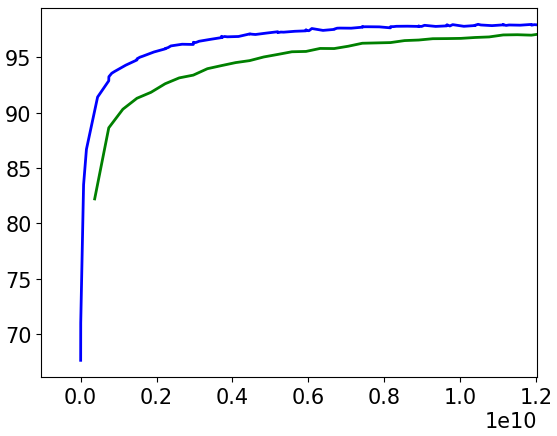} & 
        \hspace*{-3pt} \includegraphics[width=.16\textwidth]{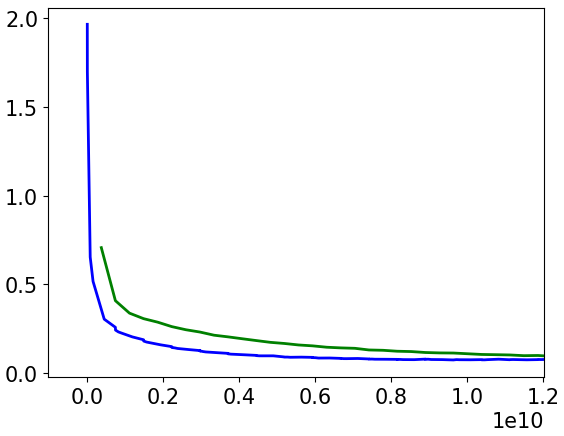}&
        \hspace*{-3pt} \includegraphics[width=.16\textwidth]{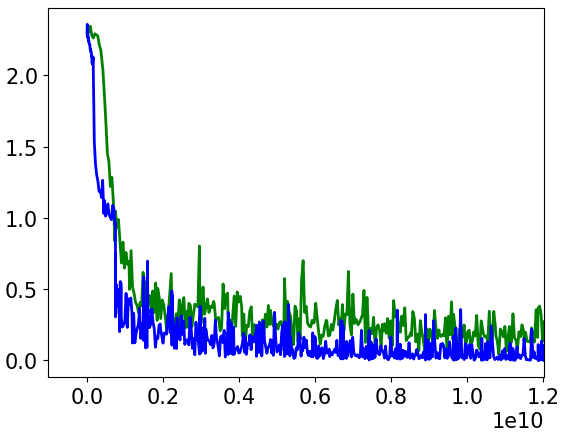} & 
        \hspace*{-3pt} \includegraphics[width=.16\textwidth]{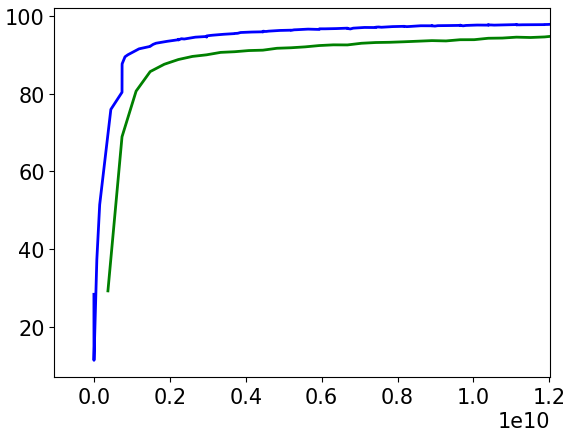} & 
        \hspace*{-3pt} \includegraphics[width=.16\textwidth]{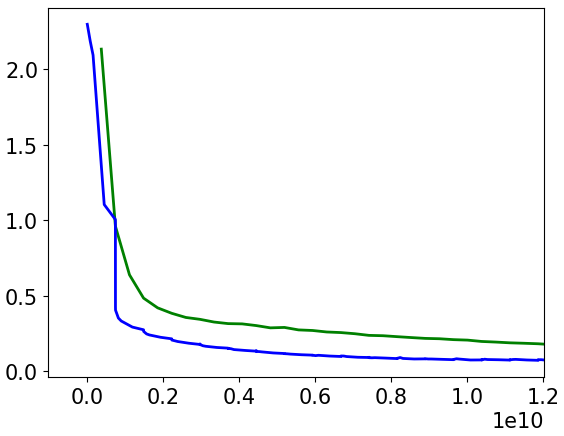} \\
        \hspace*{-3pt} \includegraphics[width=.16\textwidth]{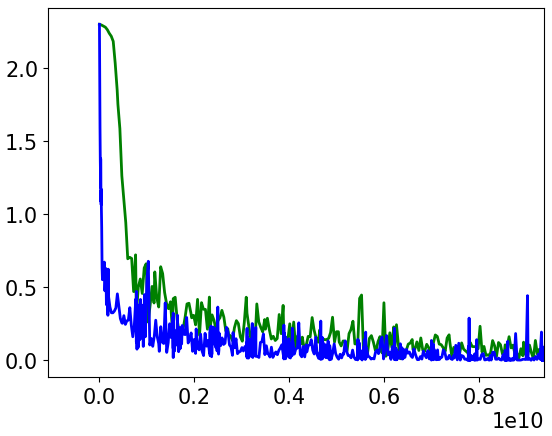} & 
        \hspace*{-3pt} \includegraphics[width=.16\textwidth]{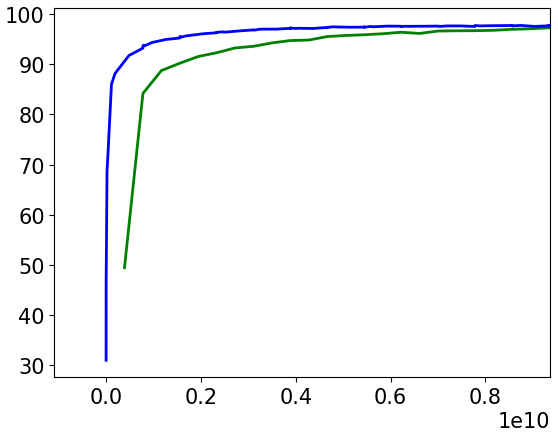} & 
        \hspace*{-3pt} \includegraphics[width=.16\textwidth]{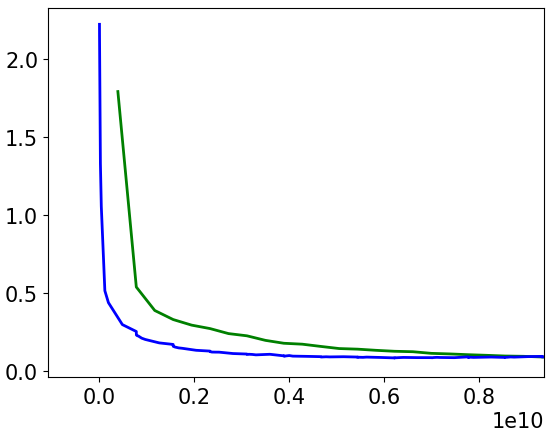} &
        \hspace*{-3pt} \includegraphics[width=.16\textwidth]{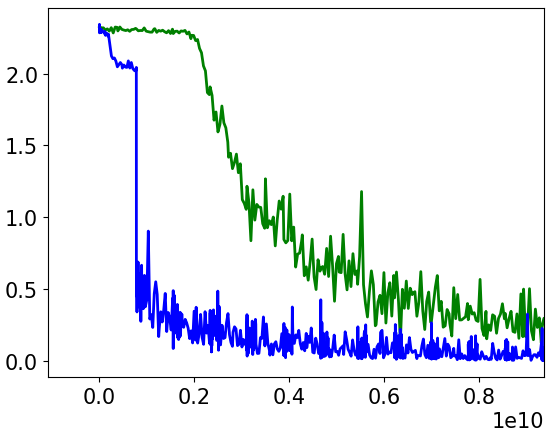} & 
        \hspace*{-3pt} \includegraphics[width=.16\textwidth]{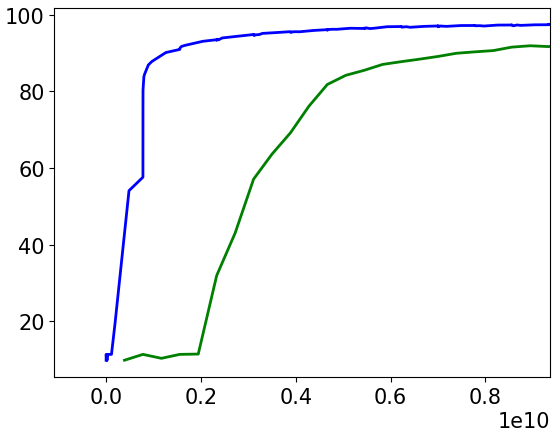} & 
        \hspace*{-3pt} \includegraphics[width=.16\textwidth]{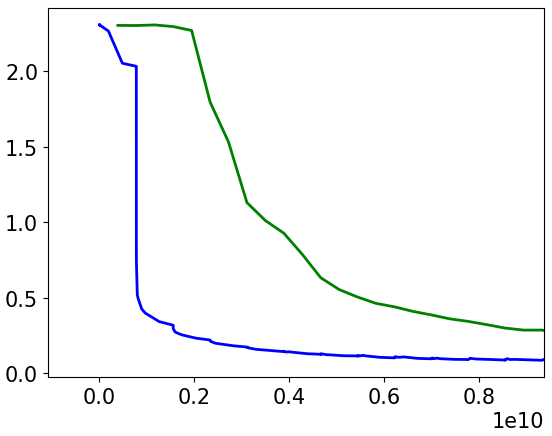} \\
        \hspace*{-3pt} \includegraphics[width=.16\textwidth]{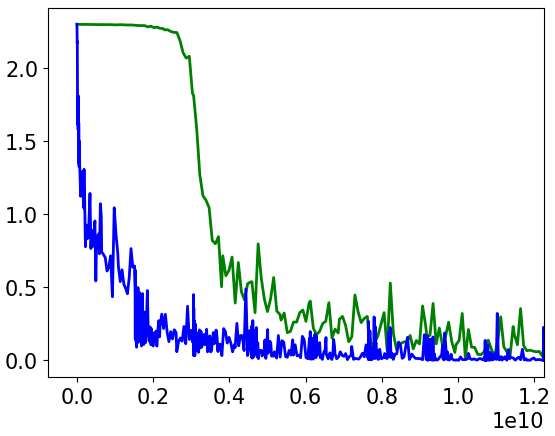} & 
        \hspace*{-3pt} \includegraphics[width=.16\textwidth]{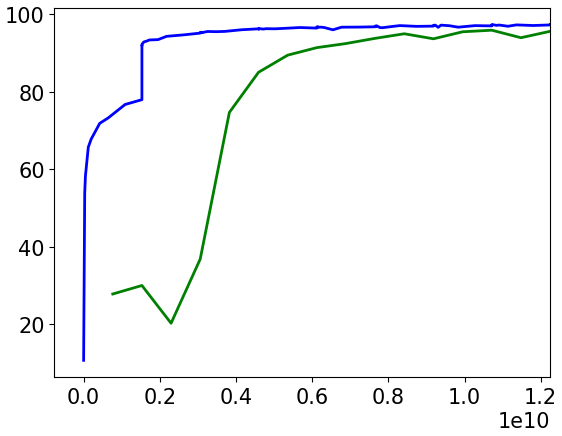} & 
        \hspace*{-3pt} \includegraphics[width=.16\textwidth]{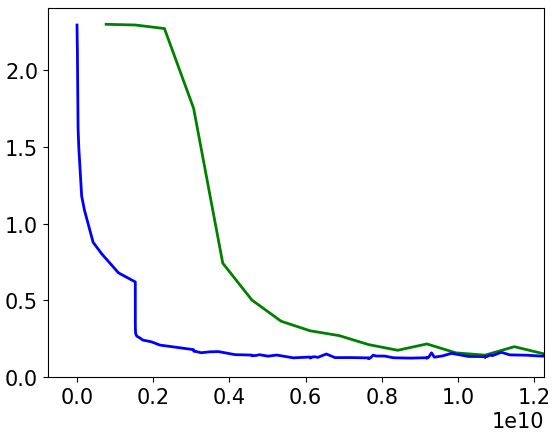} &
        \hspace*{-3pt} \includegraphics[width=.16\textwidth]{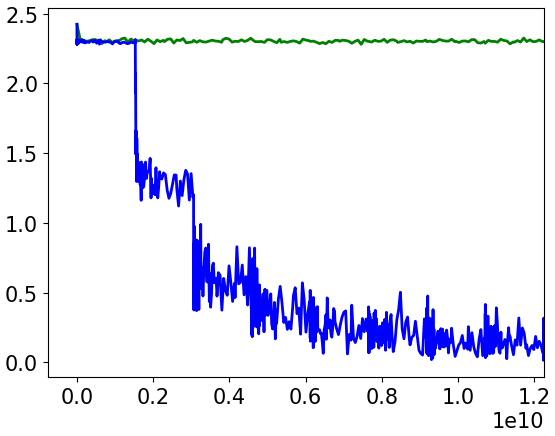} & 
        \hspace*{-3pt} \includegraphics[width=.16\textwidth]{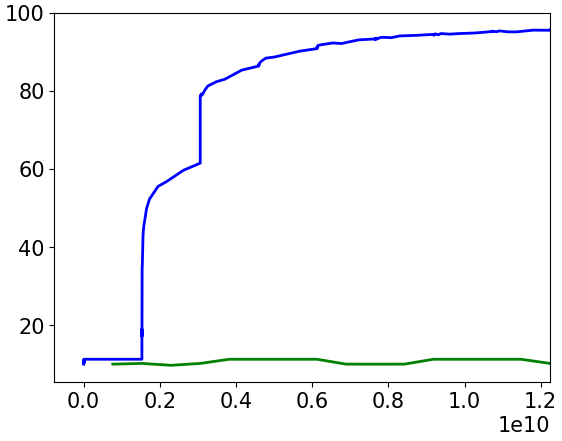} & 
        \hspace*{-3pt} \includegraphics[width=.16\textwidth]{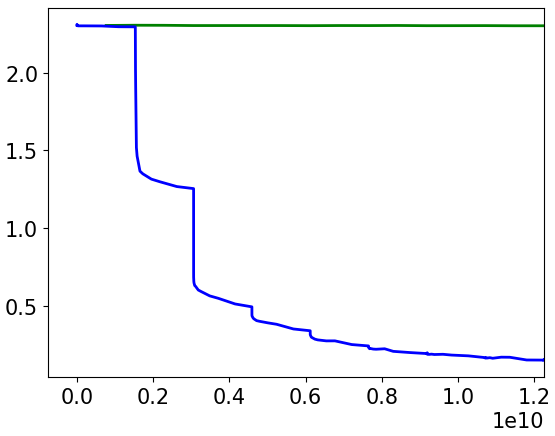} \\
    \end{tabular}
    \caption{\textit{(Best viewed in color)} Comparative performance of SGD (Green) and DANTE (Blue) on the MNIST dataset with cross-entropy loss. The rows correspond to networks (in order) $(784 \xrightarrow{} 100 \xrightarrow{} 10)$, $(784 \xrightarrow{} 250 \xrightarrow{} 10)$, $(784 \xrightarrow{} 400 \xrightarrow{} 10)$, $(784 \xrightarrow{} 600 \xrightarrow{} 10)$, $(784 \xrightarrow{} 400 \xrightarrow{} 200 \xrightarrow{} 10)$, $(784 \xrightarrow{} 400 \xrightarrow{} 200 \xrightarrow{} 100 \xrightarrow{} 10)$ and $(784 \xrightarrow{}600 \xrightarrow{} 400 \xrightarrow{} 200 \xrightarrow{} 100 \xrightarrow{} 50 \xrightarrow{} 10)$. The first three columns correspond to networks with Leaky ReLU activations; the last three correspond to networks with sigmoid activations. The first and fourth columns show training loss; second and fifth columns show test accuracy; third and sixth columns show test loss. For all the plots, X-axis is the number of weights updated.}
    \label{tab:ablation_CE}
\end{figure*}

\noindent \textbf{Feedforward Neural Networks with Cross-Entropy Loss.}
To compare \myalgo with SGD on networks with cross-entropy loss, we experiment with sigmoid and Leaky ReLU actiavtions on the MNIST dataset with the same network architectures as before, but with cross-entropy as loss function. Figure \ref{tab:ablation_CE} presents the results of these experiments.

\begin{figure}[]
    \setlength\tabcolsep{0pt}
    \centering
    \begin{tabular}{cccccc}
        \hspace*{-3pt} \includegraphics[width=.16\textwidth]{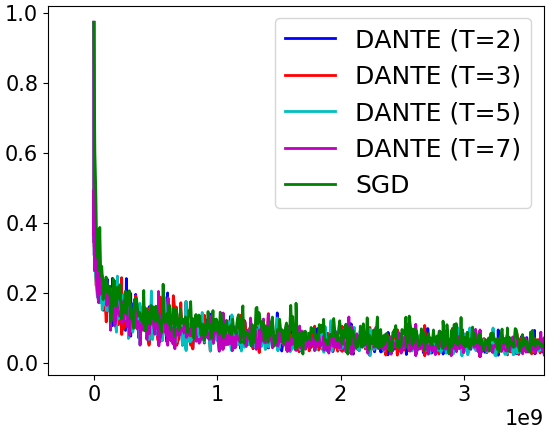} & 
        \hspace*{-3pt} \includegraphics[width=.16\textwidth]{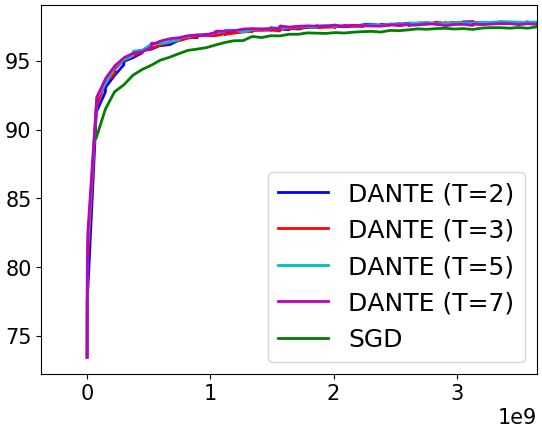} & 
        \hspace*{-3pt} \includegraphics[width=.16\textwidth]{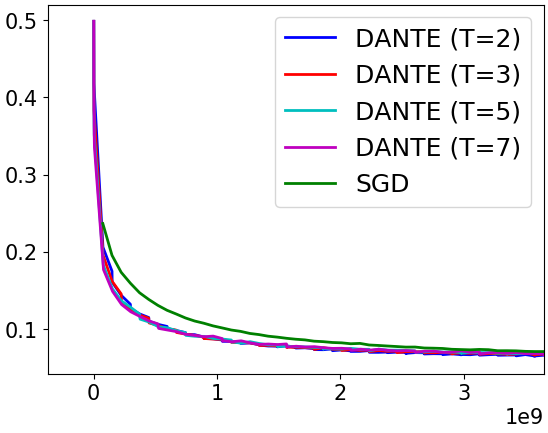} \\
        \hspace*{-3pt} \includegraphics[width=.16\textwidth]{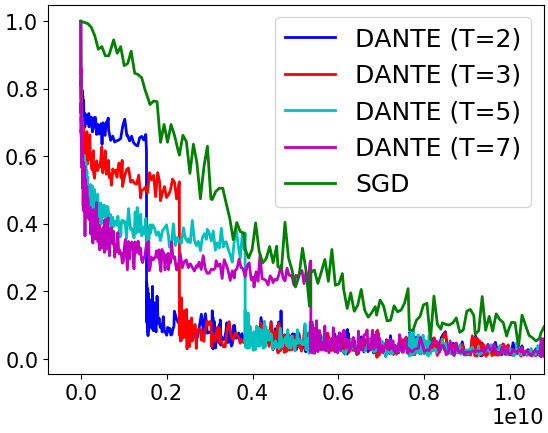} & 
        \hspace*{-3pt} \includegraphics[width=.16\textwidth]{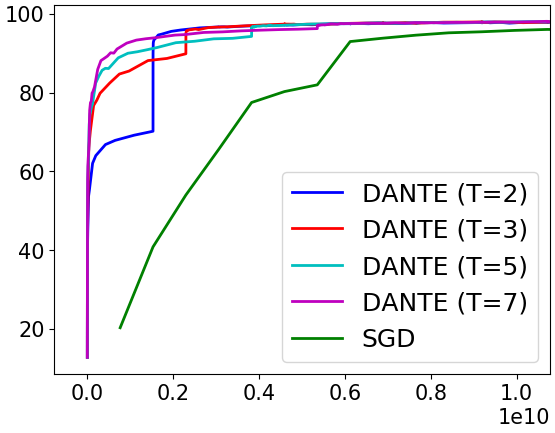} & 
        \hspace*{-3pt} \includegraphics[width=.16\textwidth]{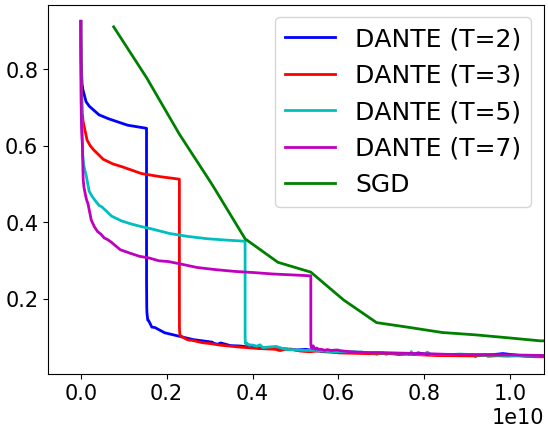}
    \end{tabular}
    \caption{\textit{(Best viewed in color)} Comparative performance of SGD (Green) and DANTE (Blue) with varying parameter $T$ on MNIST dataset. The network used in the top three plots is $(784 \xrightarrow{} 100 \xrightarrow{} 10)$ and the bottom three plots is $(784 \xrightarrow{}600 \xrightarrow{} 400 \xrightarrow{} 200 \xrightarrow{} 100 \xrightarrow{} 50 \xrightarrow{} 10)$. The first column shows training loss; the second shows test accuracy; and third shows test loss. For all plots, X-axis is the number of weights updated.}
    \label{tab:ablation_T}
\end{figure}

\subsection{Ablation Studies}
\subsubsection{Impact of Adaptive Learning Rate Methods} 
As stated earlier, in all of the abovementioned experiments, we chose the best learning rates for both SGD and \myalgo in each experiment. To go further, we also studied the use of several adaptive learning schemes with both SGD and \myalgo. The results, the final test loss at the end of training, in these studies on the MNIST dataset are presented in Table \ref{tab:adaptive}. \myalgo with some momentum is able to outperform SGD with all the popular adaptive learning rate schemes.
\begin{table}
    \centering
       \begin{tabular}{|c|c|c|} \hline
        Algorithm & Parameter  & Loss \\ \hline \hline
        \myalgo & LR = 0.001 & 0.030775  \\ \hline
         SGD & LR = 0.001 &  0.031126 \\ \hline
         SGD+Adam  & 0.0001 &  0.021704 \\ \hline
         SGD+Adagrad  & 0.0001 &  0.021896	 \\ \hline
         SGD+RMSProp  & 0.0001 &  0.021953 \\ \hline
         SGD+Momentum  & MP=0.9 &  0.021497	 \\ \hline
         \myalgo+Momentum & MP=0.0005 &  0.020816 \\ \hline
    \end{tabular}
    \caption{Loss on using various adaptive learning schemes with \myalgo and SGD. (LR = Learning Rate; MP = Momentum Parameter)}
        \label{tab:adaptive}
\end{table}

\subsubsection{Using SGD in Alternating Minimization} 
A natural question one could ask is the relevance of SNGD to train each layer of the proposed methodology. To study this empirically, we compared our algorithm to an analogous algorithm that uses SGD for each inner loop of \myalgo. Table \ref{tab:sgd_altmin} presents these results, the test loss at the end of training. Although we allowed different learning rates for the SGD variant, DANTE provides a better performance than any of these variants.
\begin{table}[]
    \centering
       \begin{tabular}{|c|c|c|} \hline
        Algorithm & Learning Rate  & Loss \\ \hline \hline
        \myalgo & 0.001 & 0.030775  \\ \hline
         SGD & 0.001 &  0.031126 \\ \hline
         SGD+AltMin  & 0.001 &  0.032912 \\ \hline
         SGD+AltMin  & 0.0001 &  0.044911	 \\ \hline
         SGD+AltMin  & 0.0005 &  0.035567 \\ \hline
    \end{tabular}
    \caption{Loss on using SGD+AltMin to learn the MNIST dataset.}
    \label{tab:sgd_altmin}
\end{table}
\subsubsection{Effect of the T Parameter}

\myalgo alternatively optimizes over each layer using SNGD. An important parameter for SNGD which can affect performance is the number of epochs for which SNGD algorithm runs for each layer (parameter $T$ in Algorithm \ref{alg_sngd}). We vary $T$ and compare how the performance of \myalgo varies when compared to SGD. The results are presented in Figure \ref{tab:ablation_T}. We observe that \myalgo is fairly robust to changes in $T$. The network used for the presented result was $(784 \xrightarrow{} 100 \xrightarrow{} 10)$ with the MNIST dataset. Observing the performance of this experiment, we chose $T=5$ for all our experiments.

\subsection{Other Empirical Studies}

\subsubsection{Comparative Study}

\begin{figure}[]
    \setlength\tabcolsep{0pt}
    \centering
    \begin{tabular}{cccccc}
        \hspace*{-3pt} \includegraphics[width=.16\textwidth]{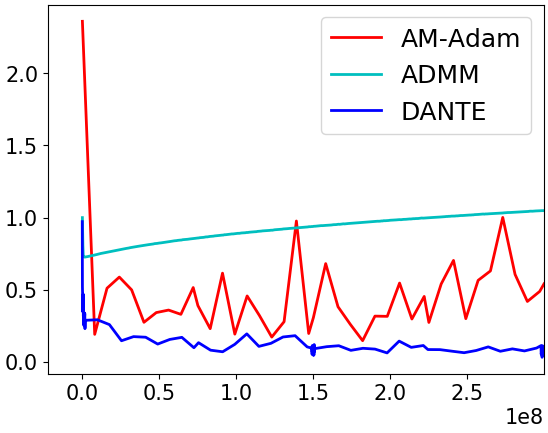} & 
        \hspace*{-3pt} \includegraphics[width=.16\textwidth]{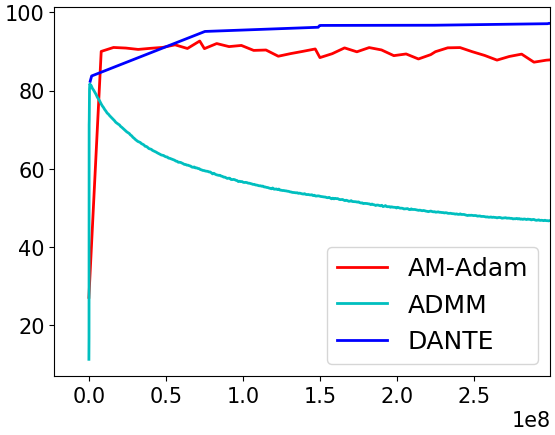} & 
        \hspace*{-3pt} \includegraphics[width=.16\textwidth]{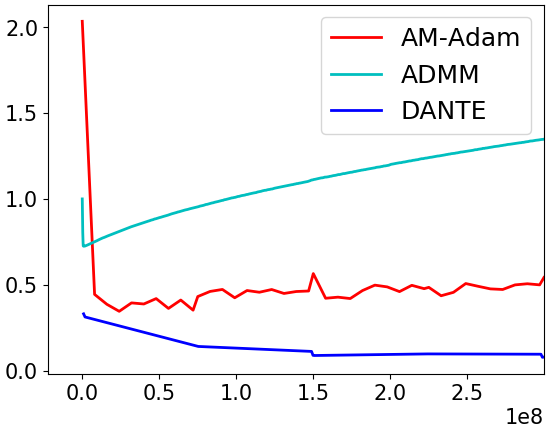} \\
        \hspace*{-3pt} \includegraphics[width=.16\textwidth]{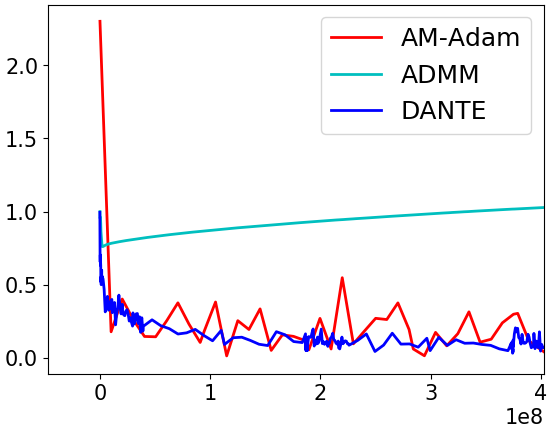} & 
        \hspace*{-3pt} \includegraphics[width=.16\textwidth]{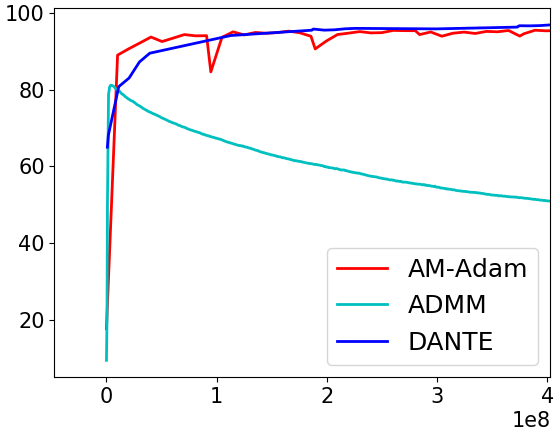} & 
        \hspace*{-3pt} \includegraphics[width=.16\textwidth]{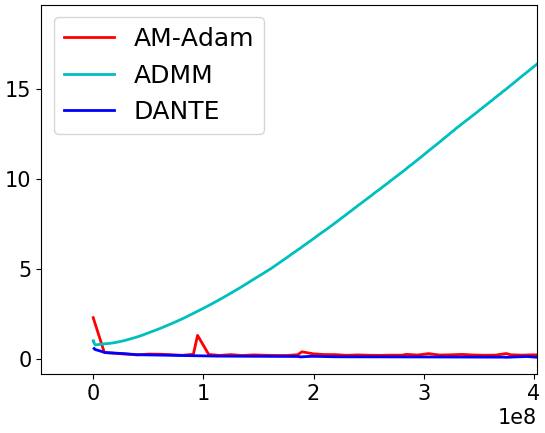}
    \end{tabular}
    \caption{\textit{(Best viewed in color)} Comparative performance of DANTE (Blue), Choromanska's AM-Adam (red) and Taylor's ADMM based algorithm (cyan) on MNIST dataset. The network used in the top three plots is $(784 \xrightarrow{} 100 \xrightarrow{} 10)$ and the bottom three plots is $(784 \xrightarrow{}100 \xrightarrow{} 100 \xrightarrow{} 100 \xrightarrow{} 10)$. The first column shows training loss; the second shows test accuracy; and third shows test loss. For all plots, X-axis is the number of parameters updated.}
    \label{tab:comparision}
\end{figure}

We have compared \myalgo to other Alternating-Minimization approaches for training neural networks: Choromanska's \cite{pmlr-v97-choromanska19a} AM-Adam (which was their best performing variant) and Taylor's \cite{taylor2016training} ADMM approach, from the codes provided by the corresponding authors. The results of comparison between \myalgo, AM-Adam and ADMM on MNIST are presented in figure \ref{tab:comparision}. Taylor's ADMM algorithm peaked at an accuracy of about 81\% for both the networks. Note that Taylor's method seems to show significant instability when trained on well-known datasets over a longer period. Our proposed method does not suffer from this issue. As is clear of the graphs and results, \myalgo outperforms both the AM-Adam and Taylor's ADMM algorithm.

\subsubsection{Regression Tasks}

All the above-mentioned experiments were done for the classification task. We hence also studied the performance of \myalgo on standard regression datasets from the UCI repository. Table \ref{tab:regression} presents the final test error values at the end of training. We followed The standard benchmark evaluation setup of each of the datasets, as specified in the repository. These results further support the promise of the proposed method.
\begin{table}[h!]
    \centering
        \begin{tabular}{|c|c|c|} \hline
        & \myalgo  & SGD \\ \hline \hline
        Air-Foil & \textbf{0.066852} & 0.069338 \\	\hline 
        Fires  & \textbf{0.024988} & 0.029008 \\ \hline 
        CCPP & 	\textbf{0.000283} & 0.0003009 \\	\hline 
        \end{tabular}
      \captionof{table}{Test error on UCI regression datasets with \myalgo and SGD. \vspace{5ex}}
    \label{tab:regression}
\end{table}

\subsubsection{Training Autoencoder Models}

Going further, we conducted experiments to study the effectiveness of the feature representations learned using the autoencoder models trained using \myalgo and SGD. After training, we passed the datasets (from UCI repository) through the autoencoder, extracted the hidden layer representations, and then trained a linear SVM. The classification accuracy results using the hidden representations are given in Table \ref{tab:student}. The table clearly highlights the improved performance of \myalgo on this task. In case of the SVMGuide4 dataset, \myalgo showed a significant improvement of over 17\% on the classification accuracy.
\begin{table}[]
    \centering
        \begin{tabular}{|c|c|c|} \hline
        & \myalgo  & SGD \\ \hline \hline
        MNIST & \textbf{93.6\%} &  92.44\% \\ \hline
        Ionosphere & 92.45\% &  \textbf{96.22\%} \\ \hline
        SVMGuide4 & \textbf{87.65\%} & 70.37\%\\ \hline
	    USPS & \textbf{90.43\%} &  89.49\% \\ \hline
   	    Vehicle & \textbf{77.02\%} & 74.80\% \\ \hline
        \end{tabular}
      \caption{Classification accuracies using ReLU autoencoder features on different datasets.}
    \label{tab:student}
\end{table}

\begin{figure}[]
    \centering
    \includegraphics[width=.30\textwidth]{{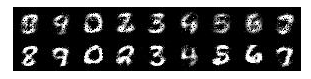}}
    \caption{Reconstructions using the autoencoder models with ReLU activation. \textit{Top}: Model trained using SGD; \textit{Bottom}: Model trained using DANTE.}
    \label{fig:reconstruction}
\end{figure}

Figure \ref{fig:reconstruction} shows some of the best reconstructions obtained by trained models for the autoencoder with the ReLU activation on MNIST in both cases (SGD and \myalgo). The model trained using DANTE shows qualitatively better reconstructions, when compared to reconstructions obtained using a model trained by SGD under the same settings.

\section{Proofs}
\subsection{Proof of Theorem \ref{theorem_iglm_relu}}
\label{appendix_idealglm}

\begin{proof}
Consider $\mathbf{w} \in \mathbb{B}(0, W)$, $\|\textbf{w}\| \le W$ such that $\hat{err}_m(\textbf{w)}= \frac{1}{m} \sum_{i=1}^m (y_i - \phi \langle \textbf{w}, \textbf{x}_i \rangle)^2 \ge \epsilon$, where $m$ is the total number of samples. Also let $\textbf{v}$ be a point $\epsilon/\kappa$-close to minima $\textbf{w}^*$ with $\kappa = \frac{2b^{3}W}{a}$. Let $g$ be the subgradient of the generalized ReLU activation and $G$ be the subgradient of $\hat{err}_m(\textbf{w)}$. (Note that as before, $g \langle .,. \rangle$ denotes $g (\langle .,. \rangle)$.) Then:
\begin{align}
&\langle G(\textbf{w}),\textbf{w}-\textbf{v}\rangle & \notag \\
&= \frac{2}{m}\sum_{i=1}^{m}g\langle\mathbf{w},\textbf{x}_{i}\rangle\left(\phi\langle\mathbf{w},\textbf{x}_{i}\rangle-y_{i}\right)\langle \textbf{x}_{i},(\mathbf{w}-\mathbb{\mathbf{v}})\rangle & \notag \\
\begin{split}
& = \frac{2}{m}\sum_{i=1}^{m}g\langle\mathbf{w},\textbf{x}_{i}\rangle\left(\phi\langle\mathbf{w},\textbf{x}_{i}\rangle-\phi\langle\mathbf{w^{*},}\textbf{x}_{i}\rangle\right)  \\ & \qquad \left[\langle \textbf{x}_{i},\mathbf{w}-\mathbf{w^{*}}\rangle+\langle \textbf{x}_{i},\mathbf{w^{*}}-\mathbf{v}\rangle\right] \end{split} \tag{Step 1}\\
\begin{split}
&  \ge \frac{2}{m}\sum_{i=1}^{m}g\langle\mathbf{w},\textbf{x}_{i}\rangle \Bigl[ b^{-1}\left(\phi\langle\mathbf{w},\textbf{x}_{i}\rangle-\phi\langle\mathbf{w^{*},}\textbf{x}_{i}\rangle\right)^{2} \\ & \qquad +  \left(\phi\langle\mathbf{w},\textbf{x}_{i}\rangle-\phi\langle\mathbf{w^{*},}\textbf{x}_{i}\rangle\right)\langle \textbf{x}_{i},\mathbf{w^{*}}-\mathbf{v}\rangle \Bigr] \end{split} \tag{Step 2}\\
\begin{split}
& \ge \frac{2}{m}\sum_{i=1}^{m}g\langle\mathbf{w},\textbf{x}_{i}\rangle \Bigl[b^{-1}\left(\phi\langle\mathbf{w},\textbf{x}_{i}\rangle-\phi\langle\mathbf{w^{*},}\textbf{x}_{i}\rangle\right)^{2} \\ & \qquad -   |\phi\langle\mathbf{w},\textbf{x}_{i}\rangle-\phi\langle\mathbf{w^{*},}\textbf{x}_{i}\rangle| \| \textbf{x}_{i}\| \|\mathbf{w^{*}}-\mathbf{v}\| \Bigr] \end{split} \notag \\
\begin{split}
& \ge \frac{2}{m}\sum_{i=1}^{m}ab^{-1}\left(\phi\langle\mathbf{w},\textbf{x}_{i}\rangle-\phi\langle\mathbf{w}^{*},\textbf{x}_{i}\rangle\right)^{2} \\ & \qquad \\ & \qquad -  \frac{2}{m}\sum_{i=1}^{m} b |\phi\langle\mathbf{w},\textbf{x}_{i}\rangle-\phi\langle\mathbf{w^{*},}\textbf{x}_{i}\rangle| \| \textbf{x}_{i}\| \|\mathbf{w^{*}}-\mathbf{v}\| \end{split} \tag{Step 3}\\
\begin{split}
&  \ge \frac{2}{m} \sum^m_{i=1} a b^{-1}\left(\phi\langle\mathbf{w},\textbf{x}_{i}\rangle-\phi\langle\mathbf{w}^{*},\textbf{x}_{i}\rangle\right)^{2} \\ & \qquad - \frac{2}{m}\sum_{i=1}^{m} b^2 \| \langle\mathbf{w},\textbf{x}_{i}\rangle-\langle\mathbf{w^{*},}\textbf{x}_{i}\rangle \| \frac{\epsilon}{\kappa}\|\textbf{x}_i\| \end{split} \tag{Step 4}\\
&  \ge  2a b^{-1}\epsilon  - \frac{a \epsilon}{bWm} \sum_{i=1}^{m} \| \langle\mathbf{w},\textbf{x}_{i}\rangle-\langle\mathbf{w^{*},}\textbf{x}_{i}\rangle \| \|\textbf{x}_i\| & \notag \\
&  \ge  2a b^{-1}\epsilon  - \frac{a \epsilon}{bWm} \sum_{i=1}^{m} \| \textbf{w} - \textbf{w}^* \| \| \textbf{x}_i \|^2 & \notag \\
&  \ge a b^{-1} \epsilon (2  - \frac{1}{W} \| \textbf{w} - \textbf{w}^* \| ) & \tag{Step 5} \\
& \ge 0  
\end{align}

\end{proof}

In the above proof, we first use the fact (in Step 1) that in the GLM, there is some $\textbf{w}^*$ such that $\phi \langle \textbf{w}^{*}, \textbf{x}_i \rangle = y_i$. Then, we use the fact (in Steps 2 and 4) that the generalized ReLU function is $b$-Lipschitz, and the fact that the minimum value of the quasigradient of $g$ is $a$ (Step 3). Subsequently, in Step 5, we simply use the given bounds on the variables $\textbf{x}_i, \textbf{w},\textbf{w}^{*}$ due to the setup of the problem ($\textbf{w}\in \mathbb{B}(0,W)$, and $\textbf{x}_i\in \mathbb{B}(0,1)$, the unit $d$-dimensional ball, as defined earlier in this section).

\subsection{Proof of Corollary \ref{cor_iglm_relu}}
\label{appendix_idealglm_cor}

\begin{proof}
Similar to the previous proof, consider $\mathbf{w} \in \mathbb{B}(0, W)$, $\|\textbf{w}\| \le W$ such that $\frac{1}{m} \sum_{i=1, \langle \textbf{w}, \textbf{x}_i \rangle > 0}^m (y_i - \phi \langle \textbf{w}, \textbf{x}_i \rangle)^2 \ge \epsilon$, where $m$ is the total number of samples. Also let $\textbf{v}$ be a point $\epsilon/\kappa$-close to minima $\textbf{w}^*$ with $\kappa = 2b^{2}W$. Let $g$ be the subgradient of the generalized ReLU activation and $G$ be the subgradient of $\hat{err}_m(\textbf{w)}$.

Note here that since $\phi$ is ReLU, if $\langle \textbf{w}, \textbf{x}_i \rangle \le 0$, then $\phi \langle \textbf{w}, \textbf{x}_i \rangle = 0$ and $g \langle \textbf{w}, \textbf{x}_i \rangle = 0$. Then:

\begin{align*}
&\langle G(\textbf{w}),\textbf{w}-\textbf{v}\rangle & \notag \\
&= \frac{2}{m}\sum_{i=1}^{m}g\langle\mathbf{w},\textbf{x}_{i}\rangle\left(\phi\langle\mathbf{w},\textbf{x}_{i}\rangle-y_{i}\right)\langle \textbf{x}_{i},(\mathbf{w}-\mathbb{\mathbf{v}})\rangle & \notag \\
\begin{split}
& = \frac{2}{m}\sum_{i=1}^{m}g\langle\mathbf{w},\textbf{x}_{i}\rangle\left(\phi\langle\mathbf{w},\textbf{x}_{i}\rangle-\phi\langle\mathbf{w^{*},}\textbf{x}_{i}\rangle\right)  \\ & \qquad \left[\langle \textbf{x}_{i},\mathbf{w}-\mathbf{w^{*}}\rangle+\langle \textbf{x}_{i},\mathbf{w^{*}}-\mathbf{v}\rangle\right] \end{split} \tag{Step 1}\\
\begin{split}
&  \ge \frac{2}{m}\sum_{i=1}^{m}g\langle\mathbf{w},\textbf{x}_{i}\rangle \Bigl[ b^{-1}\left(\phi\langle\mathbf{w},\textbf{x}_{i}\rangle-\phi\langle\mathbf{w^{*},}\textbf{x}_{i}\rangle\right)^{2} \\ & \qquad +  \left(\phi\langle\mathbf{w},\textbf{x}_{i}\rangle-\phi\langle\mathbf{w^{*},}\textbf{x}_{i}\rangle\right)\langle \textbf{x}_{i},\mathbf{w^{*}}-\mathbf{v}\rangle \Bigr] \end{split} \tag{Step 2}\\
\begin{split}
&  = \frac{2}{m}\sum_{\substack{i=1\\ \langle \textbf{w}, \textbf{x}_i \rangle > 0}}^{m}g\langle\mathbf{w},\textbf{x}_{i}\rangle \Bigl[ b^{-1}\left(\phi\langle\mathbf{w},\textbf{x}_{i}\rangle-\phi\langle\mathbf{w^{*},}\textbf{x}_{i}\rangle\right)^{2} \\ & \qquad +  \left(\phi\langle\mathbf{w},\textbf{x}_{i}\rangle-\phi\langle\mathbf{w^{*},}\textbf{x}_{i}\rangle\right)\langle \textbf{x}_{i},\mathbf{w^{*}}-\mathbf{v}\rangle \Bigr] \end{split} \tag{Step 3}\\
\begin{split}
& \ge \frac{2}{m}\sum_{\substack{i=1\\ \langle \textbf{w}, \textbf{x}_i \rangle > 0}}^{m}b\Bigl[b^{-1}\left(\phi\langle\mathbf{w},\textbf{x}_{i}\rangle-\phi\langle\mathbf{w^{*},}\textbf{x}_{i}\rangle\right)^{2} \\ & \qquad -   |\phi\langle\mathbf{w},\textbf{x}_{i}\rangle-\phi\langle\mathbf{w^{*},}\textbf{x}_{i}\rangle| \| \textbf{x}_{i}\| \|\mathbf{w^{*}}-\mathbf{v}\| \Bigr] \end{split} \notag \\
\begin{split}
& \ge \frac{2}{m}\sum_{\substack{i=1\\ \langle \textbf{w}, \textbf{x}_i \rangle > 0}}^{m}b\Bigl[b^{-1}\left(\phi\langle\mathbf{w},\textbf{x}_{i}\rangle-\phi\langle\mathbf{w}^{*},\textbf{x}_{i}\rangle\right)^{2} \\ & \qquad \\ & \qquad -  b \| \langle\mathbf{w},\textbf{x}_{i}\rangle-\langle\mathbf{w^{*},}\textbf{x}_{i}\rangle \| \frac{\epsilon}{\kappa}\|\textbf{x}_i\| \Bigr] \end{split} \tag{Step 4}\\
\begin{split}
&  = \frac{2}{m} \sum^m_{\substack{i=1\\ \langle \textbf{w}, \textbf{x}_i \rangle > 0}} \left(\phi\langle\mathbf{w},\textbf{x}_{i}\rangle-\phi\langle\mathbf{w}^{*},\textbf{x}_{i}\rangle\right)^{2} - \\ & \qquad  \frac{2}{m} \sum^m_{\substack{i=1\\ \langle \textbf{w}, \textbf{x}_i \rangle > 0}} b^2 \| \langle\mathbf{w},\textbf{x}_{i}\rangle-\langle\mathbf{w^{*},}\textbf{x}_{i}\rangle \| \frac{\epsilon}{\kappa}\|\textbf{x}_i\| \end{split} \tag{Step 5}\\
&  \ge  2\epsilon - \frac{2}{m} \sum^m_{\substack{i=1\\ \langle \textbf{w}, \textbf{x}_i \rangle > 0}} b^2 \| \textbf{w} - \textbf{w}^* \| \frac{\epsilon}{\kappa}\| \textbf{x}_i \|^2 & \tag{Step 6}\\
&  =  2\epsilon - \frac{2}{m} m b^2 \| \textbf{w} - \textbf{w}^* \| \frac{\epsilon}{\kappa} & \tag{Step 7}\\
&  \ge \epsilon (2  - \frac{1}{W} \| \textbf{w} - \textbf{w}^* \| ) & \notag \\
& \ge 0  
\end{align*}
\end{proof}

The proof uses similar arguments as the proof in Theorem \ref{theorem_iglm_relu}. In Step 3, we use the fact that $g \langle \textbf{w}, \textbf{x}_i \rangle = 0$ if $\langle \textbf{w}, \textbf{x}_i \rangle \le 0$ and $b$ otherwise. For Step 7, we observe that there at most $m$ $i$'s.

\subsection{Proof of Theorem \ref{theorem_noisyglm_relu}}
\label{appendix_noisyglm}



\begin{proof}
Here, $\forall i, y_i \in [0,1]$, the following holds:

\begin{equation}
\label{eq_noisyglm_appendix}
y_i = \phi \langle \textbf{w}^*, \textbf{x} \rangle + \xi_{i}
\end{equation}

\noindent where $\{\xi_i\}^m_{i=1}$ are zero mean, independent and bounded random variables, i.e. $\forall i \in [m], || \xi_i || \le 1$. Then, $\hat{err}_m(\textbf{w})$ may be written as follows (expanding $y_i$ as in Eqn \ref{eq_noisyglm_appendix}):
	
\begin{align*}
\hat{err}_m(\textbf{w}) &= \frac{1}{m} \sum_{i=1}^m (y_i - \phi \langle \textbf{w}, \textbf{x}_i \rangle)^2 \\
&= \frac{1}{m} \Big( \sum_{i=1}^m (\phi \langle \textbf{w}^*, \textbf{x}_i \rangle - \phi \langle \textbf{w}, \textbf{x}_i \rangle)^2 \\
&+ \sum_{i=1}^m 2 \xi_i (\phi \langle \textbf{w}^*, \textbf{x}_i \rangle - \phi \langle \textbf{w}, \textbf{x}_i \rangle) + \sum_{i=1}^m \xi_{i}^2 \Big)
\end{align*}

\noindent Therefore, we also have (by definition of noisy GLM in Defn \ref{defn:nglm}):
\begin{align*}
\hat{err}_m(\textbf{w)} - \hat{err}_m(\textbf{w}^*) &= \frac{1}{m} \sum_{i=1}^m (\phi \langle \textbf{w}^*, \textbf{x}_i \rangle - \phi \langle \textbf{w}, \textbf{x}_i \rangle)^2 \\
& + \frac{1}{m}\sum_{i=1}^m 2 \xi_i (\phi \langle \textbf{w}^*, \textbf{x}_i \rangle - \phi \langle \textbf{w}, \textbf{x}_i \rangle)
\end{align*}

\noindent Consider $|| \textbf{w} || \le W$ such that $\hat{err}_m(\textbf{w)} - \hat{err}_m(\textbf{w}^*) \ge \epsilon$. Also, let $\textbf{v}$ be a point $\epsilon/\kappa$-close to minima $\textbf{w}^*$ with $\kappa = \frac{2b^{3}W}{a}$. Let $g$ be the subgradient of the generalized ReLU activation and $G$ be the subgradient of $\hat{err}_m(\textbf{w)}$, as before. Then:

\begin{align*}
& \langle G(\textbf{w}),\textbf{w}-\textbf{v}\rangle & \notag\\  
&= \frac{2}{m}\sum_{i=1}^{m}g\langle\mathbf{w},\textbf{x}_{i}\rangle\left(\phi\langle\mathbf{w},\textbf{x}_{i}\rangle-y_{i}\right)\langle \textbf{x}_{i},(\mathbf{w}-\mathbb{\mathbf{v}})\rangle & \notag\\
\begin{split}&= \frac{2}{m}\sum_{i=1}^{m} g\langle\mathbf{w},\textbf{x}_{i}\rangle\left ( \phi\langle\mathbf{w},\textbf{x}_{i}\rangle   -  \phi\langle\mathbf{w}^{*},\textbf{x}_{i}\rangle - \xi_i \right) \\ & \qquad \qquad \left[\langle \textbf{x}_{i},\mathbf{w}-\mathbf{w}^{*}\rangle+\langle \textbf{x}_{i},\mathbf{w}^{*}-\mathbf{v}\rangle\right] \end{split} \tag{Step 1}\\ 
\begin{split}
& \ge \frac{2b^{-1}}{m} \sum_{i=1}^m g\langle\mathbf{w},\textbf{x}_{i}\rangle (\phi \langle \textbf{w}^*, \textbf{x}_i \rangle - \phi \langle \textbf{w}, \textbf{x}_i \rangle)^2 \\ & \qquad \qquad - \frac{2}{m} \sum_{i=1}^m  g\langle\mathbf{w},\textbf{x}_{i}\rangle \xi_i (\langle \textbf{w}, \textbf{x}_i \rangle - \langle \textbf{w}^*, \textbf{x}_i \rangle) \\ & \qquad \qquad + \frac{2}{m} \sum_{i=1}^m g\langle\mathbf{w},\textbf{x}_{i}\rangle \\ & \qquad \qquad \qquad \cdot (\phi \langle \textbf{w}, \textbf{x}_i \rangle - \phi \langle \textbf{w}^*, \textbf{x}_i \rangle - \xi_i) \langle \textbf{w}^*-\textbf{v},\textbf{x}_i \rangle \end{split} \tag{Step 2}\\
\begin{split}& \ge \frac{2b^{-1}}{m} \sum_{i=1}^m g\langle\mathbf{w},\textbf{x}_{i}\rangle (\phi \langle \textbf{w}^*, \textbf{x}_i \rangle - \phi \langle \textbf{w}, \textbf{x}_i \rangle)^2 \\ & \qquad \qquad - \frac{2}{m} \sum_{i=1}^m g\langle\mathbf{w},\textbf{x}_{i}\rangle \xi_i (\langle \textbf{w}, \textbf{x}_i \rangle -  \langle \textbf{w}^*, \textbf{x}_i \rangle) \\ & \qquad \qquad - 2 \frac{\epsilon b^2}{\kappa} (||\textbf{w} - \textbf{w*}|| + \frac{1}{m} \sum_{i=1}^m |\xi_i|)  \end{split} \tag{Step 3}\\
\begin{split} & = \frac{2b^{-1}}{m} \sum_{i=1}^m  a [(\phi \langle \textbf{w}^*, \textbf{x}_i \rangle - \phi \langle \textbf{w}, \textbf{x}_i \rangle)^2
\\ & \qquad \qquad - 2 \xi_i (\phi \langle \textbf{w}, \textbf{x}_i \rangle - \phi \langle \textbf{w}^*, \textbf{x}_i \rangle)] 
\\ & \qquad \qquad - \frac{2}{m} \sum_{i=1}^m [g\langle\mathbf{w},\textbf{x}_{i}\rangle ( \xi_i (\langle \textbf{w}, \textbf{x}_i \rangle - \langle \textbf{w}^*, \textbf{x}_i \rangle))  \\ &  \qquad \qquad - 2ab^{-1} \xi_i (\phi \langle \textbf{w}, \textbf{x}_i \rangle - \phi \langle \textbf{w}^*, \textbf{x}_i \rangle)] \\ & \qquad \qquad - 2 \frac{\epsilon b^2}{\kappa}(||\textbf{w} - \textbf{w*}|| + \frac{1}{m} \sum_{i=1}^m|\xi_i|) 
\end{split} \tag{Step 4}\\
 & \ge 2ab^{-1}\epsilon - 2 \frac{\epsilon b^2}{\kappa} (||\textbf{w} - \textbf{w*}|| + \frac{1}{m} \sum_{i=1}^m|\xi_i|) \\ & \qquad \qquad + \frac{1}{m}\sum^m_{i=1} \xi_i \lambda_i(\textbf{w}) & \tag{Step 5}\\
 & \ge 2ab^{-1}\epsilon - ab^{-1}W^{-1}\epsilon(||\textbf{w} - \textbf{w*}|| + \frac{1}{m} \sum_{i=1}^m |\xi_i|) \\ & \qquad \qquad +  \frac{1}{m}\sum^m_{i=1} \xi_i \lambda_i(\textbf{w}) \tag{Step 6} \\
 & \ge 2ab^{-1}\epsilon - ab^{-1}\epsilon(1 + W^{-1}) +  \frac{1}{m}\sum^m_{i=1} \xi_i \lambda_i(\textbf{w}) \tag{Step 7} \\
 & \ge - ab^{-1}\epsilon W^{-1} +  \frac{1}{m}\sum^m_{i=1} \xi_i \lambda_i(\textbf{w}) \tag{Step 8}
\end{align*}
\vspace{2pt}
Here, $\lambda_i(\textbf{w}) = 2g\langle\mathbf{w},\textbf{x}_{i}\rangle (\langle \textbf{w}, \textbf{x}_i \rangle - \langle \textbf{w}^*, \textbf{x}_i \rangle) - 4ab^{-1} (\phi \langle \textbf{w}, \textbf{x}_i \rangle - \phi \langle \textbf{w}^*, \textbf{x}_i \rangle)$, and \\
$ |\xi_i \lambda_i(\textbf{w})| \le 2b(|\langle \textbf{w}, \textbf{x}_i \rangle - \langle \textbf{w}^*, \textbf{x}_i \rangle| + 4ab^{-1}|\phi \langle \textbf{w}, \textbf{x}_i \rangle - \phi \langle \textbf{w}^*, \textbf{x}_i \rangle|)  \le 2b (3|\langle \textbf{w}, \textbf{x}_i \rangle - \langle \textbf{w}^*, \textbf{x}_i \rangle|) \le  2b(6W) = 12bW $

The above proof uses arguments similar to the proof for the idealized GLM (please see the lines after the proof of Theorem \ref{theorem_iglm_relu}, viz. the $b$-Lipschitzness of the generalized ReLU, and the problem setup). Now, when 
\[\frac{1}{m}\sum^m_{i=1} \xi \lambda_i(\textbf{w}) \ge ab^{-1} W^{-1} \epsilon \]
our model is SLQC. By simply using the Hoeffding's bound, we get that the theorem statement holds for $m \ge \frac{288 b^4 W^4 }{a^2 } log(1/\delta)/ \epsilon^2 $.
\end{proof}

\subsection{Viewing the Outer Layer of a Neural Network as a Set of GLMs}
\label{sec_appendix_set}
Given an (unknown) distribution $\mathcal{D}$, let the layer be characterized by a linear operator $W \in \mathbb{R}^{d \times d'}$ and a non-linear activation function defined by $\phi:\mathbb{R} \rightarrow \mathbb{R}$. Let the layer output be defined by $\phi \langle W, \textbf{x}\rangle$, where $\textbf{x} \in \mathbb{R}^d$ is the input, and $\phi$ is used element-wise in this function.

Consider the mean squared error loss, commonly used in neural networks, given by:
\begin{align*}
\underset{W}\min \quad err(W) & = \underset{W}\min \quad \mathbb E_{\mathbf x \sim \cal D} \|\phi \langle W, \mathbf x \rangle - \mathbf y\|_2^2 \\
& = \underset{W}\min \quad \mathbb E_{\mathbf x \sim \cal D} \|\sum^{d'}_{i=1} \phi \langle W_{:,i}, \mathbf x \rangle - \mathbf y_i \|_2^2 \\
& = \underset{W}\min \quad  \sum^{d'}_{i=1} \mathbb E_{\mathbf x \sim \cal D} \| \phi \langle W_{:,i}, \mathbf x \rangle - \mathbf y_i \|_2^2 \\
& = \sum^{d'}_{i=1} \underset{W}\min \quad  \mathbb E_{\mathbf x \sim \cal D} \| \phi \langle W_{:,i}, \mathbf x \rangle - \mathbf y_i \|_2^2
\end{align*}

Each of these sub-problems above is a GLM, which can be solved effectively using SNGD as seen in Theorem \ref{theorem_sngd_hazan}, which we leverage in this work.

\subsection{Proof of Theorem \ref{theorem_relu_outer_layer}}
\label{appendix_relu_outer_layer}

\begin{proof} 
Consider $\mathbf{W} \in \mathbb{B}(0, W)$, $\|\textbf{W}\| \le W$ such that $\hat{err}_m(\textbf{W}) = \frac{1}{m} \sum_{i=1}^m ( \textbf{y}_i - \phi \langle \textbf{W}, \textbf{x}_i \rangle)^2 \ge d' \epsilon$, where $m$ is the total number of samples. 
Also let $\textbf{V} = [\textbf{v}_1 \enskip \textbf{v}_2 \cdots \enskip \textbf{v}_{d'}]$ be a point $\epsilon/\kappa$-close to minima $\textbf{W}^*$ with $\kappa = \frac{2b^{3}W}{a}$. Let $g$ be the subgradient of the generalized ReLU activation, $G(\textbf{W})$ be the subgradient of $\hat{err}_m(\textbf{W})$ and $G(\textbf{w}_j)$ be the subgradient of $\hat{err}_m(\textbf{w}_j )$. 
(Note that as before, $g \langle .,. \rangle$ denotes $g (\langle .,. \rangle)$.) Then:
\begin{align*}
& \langle G(\textbf{W}),\textbf{W}-\textbf{V}\rangle & \notag\\  
& = \sum^{d'}_{j=1}\langle G(\mathbf{w_j}),\mathbf{w_j}-\mathbf{v_j}\rangle_F & \tag{By definition of Frobenius inner product}\\
& = \frac{2}{m} \sum_{i=1}^{m} \sum^{d'}_{j=1}  \left(\phi \langle \mathbf{w_j},\mathbf{x_i} \rangle  - y_{ij}\right) \langle \frac{\partial (\phi \langle \mathbf{w_j},\mathbf{x_i} \rangle )}{\partial \mathbf{w_j}},(\mathbf{w_j}-\mathbb{\mathbf{v_j}})\rangle & \tag{Step 1}\\
\begin{split} &= \frac{2}{m} \sum_{i=1}^{m} \sum^{d'}_{j=1} g(\mathbf{w_j}, \textbf{x}_i) \left(\phi \langle \mathbf{w_j},\mathbf{x_i} \rangle  - y_{ij}\right) \\ & \qquad [\langle   \textbf{x}_i , \mathbf{w_j}-\mathbf{w^*_j}\rangle  + \langle   \textbf{x}_i , \mathbf{w^*_j}-\mathbf{v_j}\rangle  ]
\end{split} \tag{Step 2}\\
\begin{split}
&  \ge \frac{2}{m} \sum_{i=1}^{m} \sum^{d'}_{j=1} g\langle\mathbf{w_j},\textbf{x}_{i}\rangle \Bigl[ b^{-1}\left(\phi\langle\mathbf{w_j},\textbf{x}_{i}\rangle-\phi\langle\mathbf{w^{*}_j},\textbf{x}_{i}\rangle\right)^{2} \\ & \qquad + \left(\phi\langle\mathbf{w_j},\textbf{x}_{i}\rangle-\phi\langle\mathbf{w^{*}_j} \textbf{x}_{i}\rangle\right)\langle \textbf{x}_{i},\mathbf{w^{*}_j}-\mathbf{v_j}\rangle \Bigr] \end{split} \tag{Step 3}\\
\begin{split}
& \ge \frac{2}{m} \sum_{i=1}^{m} \sum^{d'}_{j=1} g\langle\mathbf{w_j},\textbf{x}_{i}\rangle \Bigr[b^{-1}\left(\phi\langle\mathbf{w_j},\textbf{x}_{i}\rangle-\phi\langle\mathbf{w^{*}_j,}\textbf{x}_{i}\rangle\right)^{2} \\ & \qquad -   |\phi\langle\mathbf{w_j},\textbf{x}_{i}\rangle-\phi\langle\mathbf{w^{*}_j,}\textbf{x}_{i}\rangle| \| \textbf{x}_{i}\| \|\mathbf{w^{*}_j}-\mathbf{v_j}\| \Bigl] \end{split} \tag{Step 4} \\
\begin{split}
& \ge \frac{2}{m} \sum_{i=1}^{m} \sum^{d'}_{j=1} \Bigr[ab^{-1}\left(\phi\langle\mathbf{w_j},\textbf{x}_{i}\rangle-\phi\langle\mathbf{w^{*}_j},\textbf{x}_{i}\rangle\right)^{2} \\ & \qquad -  b |\phi\langle\mathbf{w_j},\textbf{x}_{i}\rangle-\phi\langle\mathbf{w^{*}_j,}\textbf{x}_{i}\rangle| \| \textbf{x}_{i}\| \|\mathbf{w^{*}_j}-\mathbf{v_j}\| \Bigl] \end{split} \notag\\
\begin{split}
& \ge \frac{2}{m} \sum_{i=1}^{m} \sum^{d'}_{j=1} \Bigr[a b^{-1}\left(\phi\langle\mathbf{w_j},\textbf{x}_{i}\rangle-\phi\langle\mathbf{w^{*}_j},\textbf{x}_{i}\rangle\right)^{2} \\ & \qquad -  b^2 \| \langle\mathbf{w_j},\textbf{x}_{i}\rangle-\langle\mathbf{w^{*}_j,}\textbf{x}_{i}\rangle \| \frac{\epsilon}{\kappa}\|\textbf{x}_i\| \Bigl] \end{split} \tag{Step 5}\\
& \ge  2a b^{-1} d' \epsilon  - \frac{a d' \epsilon}{bWm} \sum_{i=1}^{m} \| \langle\mathbf{w},\textbf{x}_{i}\rangle-\langle\mathbf{w^{*},}\textbf{x}_{i}\rangle \| \|\textbf{x}_i\| & \tag{Step 6}\\
& \ge a b^{-1} d' \epsilon (2  - \frac{1}{Wm} \sum_{i=1}^{m}\| \textbf{w} - \textbf{w}^* \| \| \textbf{x}_i \|^2 ) & \tag{Step 7} \\
& \ge a b^{-1} d' \epsilon (2  - \frac{1}{W}\| \textbf{w} - \textbf{w}^* \| ) \ge 0
\end{align*}

In Step 6, $\| \langle\mathbf{w},\textbf{x}_{i}\rangle-\langle\mathbf{w^{*},}\textbf{x}_{i}\rangle \| = \underset{j}{\max} \| \langle\mathbf{w_j},\textbf{x}_{i}\rangle-\langle\mathbf{w^{*}_j,}\textbf{x}_{i}\rangle \| $ To simplify from Step 7 we use the fact that $\| \mathbf{W^*} \| \le W \implies \| \mathbf{w^*} \| \le W $. The remainder of the proof proceeds precisely as in Theorem \ref{theorem_iglm_relu}.
\end{proof}

\subsection{Proof of Corollary \ref{cor_relu_outer_layer}}
\label{appendix_cor_relu_outer_layer}
\begin{proof}
Let all the variables be the same as in the proof for Theorem \ref{theorem_relu_outer_layer} except that $\frac{1}{m} \sum_{i=1}^m \sum_{j=1, \langle \textbf{w}_j, \textbf{x}_i \rangle > 0}^{d'} ( y_{ij} - \phi \langle \textbf{w}_j, \textbf{x}_i \rangle)^2 \ge d' \epsilon$ and $\kappa = 2b^{2}W$. Again note here that since $\phi$ is ReLU, if $\langle \textbf{w}, \textbf{x}_i \rangle \le 0$, then $\phi \langle \textbf{w}, \textbf{x}_i \rangle = 0$ and $g \langle \textbf{w}, \textbf{x}_i \rangle = 0$. Using the results from previous proof, we continue from Step 4,
\begin{align*}
& \langle G(\textbf{W}),\textbf{W}-\textbf{V}\rangle & \notag\\ 
\begin{split}
& \ge \frac{2}{m} \sum_{i=1}^{m} \sum^{d'}_{j=1} g\langle\mathbf{w_j},\textbf{x}_{i}\rangle \Bigr[b^{-1}\left(\phi\langle\mathbf{w_j},\textbf{x}_{i}\rangle-\phi\langle\mathbf{w^{*}_j,}\textbf{x}_{i}\rangle\right)^{2} \\ & \qquad -   |\phi\langle\mathbf{w_j},\textbf{x}_{i}\rangle-\phi\langle\mathbf{w^{*}_j,}\textbf{x}_{i}\rangle| \| \textbf{x}_{i}\| \|\mathbf{w^{*}_j}-\mathbf{v_j}\| \Bigl] \end{split} \tag{Step 4, Borrowed} \\
\begin{split}
& \ge \frac{2}{m} \sum_{i=1}^{m} \sum^{d'}_{\substack{j=1\\ \langle \mathbf{w_j}, \mathbf{x}_i \rangle > 0}} b \Bigr[ b^{-1}\left(\phi\langle\mathbf{w_j},\textbf{x}_{i}\rangle-\phi\langle\mathbf{w^{*}_j},\textbf{x}_{i}\rangle\right)^{2} \Bigl] \\ & \qquad - \frac{2}{m} \sum_{i=1}^{m} \sum^{d'}_{j=1} \Bigr[ b |\phi\langle\mathbf{w_j},\textbf{x}_{i}\rangle-\phi\langle\mathbf{w^{*}_j,}\textbf{x}_{i}\rangle| \| \textbf{x}_{i}\| \|\mathbf{w^{*}_j}-\mathbf{v_j}\| \Bigl] \end{split} \tag{Step 5}\\
& \ge d' \epsilon (2  - \frac{1}{W}\| \textbf{w} - \textbf{w}^* \| ) \ge 0
\end{align*}
Simplification from Step 5 to last step follows from similar arguments as last proof.
\end{proof}

\subsection{Proof of Theorem \ref{theorem_relu_inner_layer_single_op}}
\label{appendix_relu_inner_layer_single_op}
\begin{proof}
In this case, the prediction of the network on $\mathbf{x}$ is $f(\mathbf{W_1}; \mathbf{w_2};\mathbf{x})$.\\
Consider $\mathbf{W_1} \in \mathbb{B}(0, W_1)$, $\| \mathbf{W_1} \| \le W_1$, $\| \mathbf{w_2} \| \le W_2$ such that $\hat{err}(\mathbf{W_1}, \mathbf{w_2}) \ge \epsilon $. Let $\mathbf{V_1}$ be a point $\frac{\epsilon}{\kappa}$ close to minima $\mathbf{W}_1$, where $\kappa = \left( \frac{a}{4b^5W_2^2W_1} - \frac{W_1}{\epsilon}\right)^{-1}$.\\
Let $\|f(\mathbf{W_1}; \mathbf{w_2};\mathbf{x}) - \mathbf y\|_2^2 = \|\phi_2 \langle \textbf{w}_2,\phi_{1} \langle \textbf{W}_{1}, \mathbf{x} \rangle \rangle - \mathbf y\|_2^2$ and $\langle \cdot \rangle_F$ be the Frobenius inner product. 
\begin{align*}
& \langle \nabla_{\mathbf{W_1}} \hat{err}(\mathbf{W_1}, \mathbf{w_2}),\textbf{W}_1-\textbf{V}_1\rangle_F & \notag\\  
\begin{split}
&= \frac{2}{m}\sum_{i=1}^{m} \left(\phi_2 \langle \textbf{w}_2,\phi_{1} \langle \textbf{W}_{1}, \mathbf{x_i} \rangle \rangle - y_{i}\right) \\ & \qquad \qquad \langle \frac{\partial (\phi_2 \langle \textbf{w}_2,\phi_{1} \langle \textbf{W}_{1}, \mathbf{x_i} \rangle \rangle)}{\partial \textbf{W}_1},(\mathbf{W_1}-\mathbb{\mathbf{V_1}})\rangle_F 
\end{split}\tag{Step 1}\\
\end{align*}
\vspace{-2mm}
Using chain rule, we can simplify $\frac{\partial (\phi_2 \langle \textbf{w}_2,\phi_{1} \langle \textbf{W}_{1}, \mathbf{x} \rangle \rangle)}{\partial \textbf{W}_1}$ as
\begin{align*}
& \left[\frac{\partial (\phi_2 \langle \textbf{w}_2,\phi_{1} \langle \textbf{W}_{1}, \mathbf{x} \rangle \rangle)}{\partial \textbf{W}_1}\right]^T = \frac{\partial (\phi_2 \langle \textbf{w}_2,\phi_{1} \langle \textbf{W}_{1}, \mathbf{x} \rangle \rangle)}{\partial \langle \textbf{w}_2,\phi_{1} \langle \textbf{W}_{1}, \mathbf{x} \rangle \rangle}  \\
& \qquad \cdot \left[\frac{\partial \langle \textbf{w}_2,\phi_{1} \langle \textbf{W}_{1}, \mathbf{x} \rangle \rangle}{\partial \phi_{1} \langle \textbf{W}_{1}, \mathbf{x} \rangle}^T  \cdot \frac{\partial \phi_{1} \langle \textbf{W}_{1}, \mathbf{x} \rangle}{\partial \langle \textbf{W}_{1}, \mathbf{x_i} \rangle}^T\right]^T \cdot \left[\frac{\partial \langle \textbf{W}_{1}, \mathbf{x}  \rangle}{\partial \textbf{W}_{1}}\right]^T \\
& \qquad = g_2(\textbf{W}_1, \textbf{w}_2, x) \cdot g_1(\textbf{W}_1, x) \cdot \textbf{w}_2    \cdot \mathbf{x}^T  &\tag{Let}\\
\end{align*}
Continuing from Step 1:
\begin{align*}
&= \frac{2}{m}\sum_{i=1}^{m} g_2(\textbf{W}_1, \textbf{w}_2, \textbf{x}_i) \left(\phi_2 \langle \textbf{w}_2,\phi_{1} \langle \textbf{W}_1, \mathbf{x_i} \rangle \rangle - y_{i}\right) \\ & \qquad \langle   \textbf{x}_i \textbf{w}^T_2 g_1(\textbf{W}_1, \textbf{x}_i)^T  ,(\mathbf{W_1}-\mathbf{V_1})\rangle_F & \notag \\
\begin{split} &= \frac{2}{m}\sum_{i=1}^{m} g_2(\textbf{W}_1, \textbf{w}_2, \textbf{x}_i) \left(\phi_2 \langle \textbf{w}_2,\phi_{1} \langle \textbf{W}_1, \mathbf{x_i} \rangle \rangle - y_{i}\right) \\ & \qquad [\langle   \textbf{x}_i \textbf{w}^T_2 g_1(\textbf{W}_1, \textbf{x}_i)^T  , \mathbf{W_1} \rangle_F \\ & \qquad - \langle \textbf{x}_i (\textbf{w}^*_2)^T g_1(\textbf{W}^*_1, \textbf{x}_i)^T  , \mathbf{W^*_1} \rangle_F \\ & \qquad + \langle \textbf{x}_i (\textbf{w}^*_2)^T g_1(\textbf{W}^*_1, \textbf{x}_i)^T  , \mathbf{W^*_1} \rangle_F \\ & \qquad - \langle   \textbf{x}_i \textbf{w}^T_2 g_1(\textbf{W}_1, \textbf{x}_i)^T  , \mathbf{V_1}\rangle_F  ]
\end{split}
\end{align*}
\vspace{-10pt}
\begin{align*}
\begin{split} &= \frac{2}{m}\sum_{i=1}^{m} g_2(\textbf{W}_1, \textbf{w}_2, \textbf{x}_i) \left(\phi_2 \langle \textbf{w}_2,\phi_{1} \langle \textbf{W}_1, \mathbf{x_i} \rangle \rangle - y_{i}\right)  \\ & \qquad [Tr( g_1(\textbf{W}_1, \textbf{x}_i) \textbf{w}_2 \textbf{x}^T_i \mathbf{W_1}) - Tr( g_1(\textbf{W}^*_1, \textbf{x}_i) \textbf{w}^*_2 \textbf{x}^T_i \mathbf{W^*_1}) \\ & \qquad + Tr( g_1(\textbf{W}^*_1, \textbf{x}_i) \textbf{w}^*_2 \textbf{x}^T_i \mathbf{W^*_1}) - Tr( g_1(\textbf{W}_1, \textbf{x}_i) \textbf{w}_2 \textbf{x}^T_i  \mathbf{V_1})]
\end{split} \tag{Step 2}
\end{align*}
In order to convert the above terms into a more familiar form, we begin with the following observation:
\begin{align*}
 \langle \textbf{w}_2,\phi_{1} \langle  \mathbf{W}, \mathbf{x} \rangle \rangle_F = Tr(g_1(\textbf{W}, \textbf{x})  \textbf{w}_2 \textbf{x}^T   (\mathbf{W})) 
\end{align*}
Also, note that $g_1(\mathbf{W}, \mathbf{x})$ is a diagonal $d' \times d'$ matrix consisting of $a$'s and $b$'s on the diagonal:
\begin{align*}
 & \langle \textbf{w}_2,\phi_{1} \langle \textbf{W}_1, \mathbf{x} \rangle \rangle - \langle \textbf{w}^*_2,\phi_{1} \langle  \mathbf{W}, \mathbf{x} \rangle \rangle \\ & = Tr(g_1(\mathbf{W_1}, \textbf{x})  \textbf{w}_2 \textbf{x}^T   \mathbf{W_1}) - Tr(g_1(\textbf{W}, \textbf{x})  \textbf{w}^*_2 \textbf{x}^T   \mathbf{W})
 \label{eq:trace_dotproduct}
\end{align*}

Therefore, on setting $\mathbf{W} = \mathbf{W^*_1}$ and using the fact that the generalized ReLU is $b$-Lipschitz and monotonically increasing, we have:
\begin{align*}
& ( \phi_2 \langle \textbf{w}_2,\phi_{1} \langle \textbf{W}_1, \mathbf{x} \rangle \rangle - \phi_2 \langle \textbf{w}^*_2,\phi_{1} \langle \ \mathbf{W^*_1}, \mathbf{x} \rangle \rangle )^2 \\
& \le b ( \phi_2 \langle \textbf{w}_2,\phi_{1} \langle \textbf{W}_1, \mathbf{x} \rangle \rangle - \phi_2 \langle \textbf{w}^*_2,\phi_{1} \langle \ \mathbf{W^*_1}, \mathbf{x} \rangle \rangle )
\\ & \qquad \cdot ( \langle \textbf{w}_2,\phi_{1} \langle \textbf{W}_1, \mathbf{x} \rangle \rangle - \langle \textbf{w}^*_2,\phi_{1} \langle  \mathbf{W^*_1}, \mathbf{x} \rangle \rangle ) \\
& = b ( \phi_2 \langle \textbf{w}_2,\phi_{1} \langle \textbf{W}_1, \mathbf{x} \rangle \rangle - \phi_2 \langle \textbf{w}^*_2,\phi_{1} \langle \ \mathbf{W^*_1}, \mathbf{x} \rangle \rangle )
\\ & \qquad \cdot (Tr(g_1(\mathbf{W_1}, \textbf{x})  \textbf{w}_2 \textbf{x}^T   \mathbf{W_1}) - Tr(g_1(\textbf{W}^*_1, \textbf{x})  \textbf{w}^*_2 \textbf{x}^T   \mathbf{W^*_1}))
\notag
\end{align*}
Plugging this result into Step 2:
\vspace{-2mm}
\begin{align*}
\begin{split} & \ge \frac{2}{m}\sum_{i=1}^{m} g_2(\textbf{W}_1, \textbf{w}_2, \textbf{x}_i) \\ & \qquad [b^{-1} ( \phi_2 \langle \textbf{w}_2,\phi_{1} \langle \textbf{W}_1, \mathbf{x_i} \rangle \rangle - \phi_2 \langle \textbf{w}^*_2,\phi_{1} \langle \ \mathbf{W^*_1}, \mathbf{x_i} \rangle \rangle )^2 
\\ & \qquad + \left(\phi_2 \langle \textbf{w}_2,\phi_{1} \langle \textbf{W}_1, \mathbf{x_i} \rangle \rangle - y_{i}\right) \\ & \qquad \cdot ( Tr( g_1(\textbf{W}^*_1, \textbf{x}_i) \textbf{w}^*_2 \textbf{x}^T_i \mathbf{W^*_1}) 
- Tr( g_1(\textbf{W}_1, \textbf{x}_i) \textbf{w}_2 \textbf{x}^T_i  \mathbf{V_1}))]
\end{split}\\
\begin{split} & \ge 2ab^{-1} \epsilon + \frac{2}{m}\sum_{i=1}^{m} g_2(\textbf{W}_1, \textbf{w}_2, \textbf{x}_i)  \left(\phi_2 \langle \textbf{w}_2,\phi_{1} \langle \textbf{W}_1, \mathbf{x_i} \rangle \rangle - y_{i}\right) 
\\ & \qquad \cdot [Tr( g_1(\textbf{W}^*_1, \textbf{x}_i) \textbf{w}^*_2 \textbf{x}^T_i \textbf{W}^*_1) - Tr( g_1(\textbf{W}_1, \textbf{x}_i) \textbf{w}_2 \textbf{x}^T_i   \mathbf{V_1})]
\end{split}\\
\begin{split} & \ge \frac{2a}{b} \epsilon - \frac{2}{m}\sum_{i=1}^{m} g_2(\textbf{W}_1, \textbf{w}_2, \textbf{x}_i) \cdot | \left(\phi_2 \langle \textbf{w}_2,\phi_{1} \langle \textbf{W}_1, \mathbf{x_i} \rangle \rangle - y_{i}\right)|
\\ & \qquad | Tr( g_1(\textbf{W}^*_1, \textbf{x}_i) \textbf{w}^*_2 \textbf{x}^T_i \textbf{W}^*_1) - Tr( g_1(\textbf{W}_1, \textbf{x}_i) \textbf{w}_2 \textbf{x}^T_i   \mathbf{V_1}) |
\end{split} \tag{Step 3}\\
\vspace{-2mm}
\end{align*}
First consider the term $ | Tr( g_1(\textbf{W}^*_1, \textbf{x}_i) \textbf{w}^*_2 \textbf{x}^T_i \textbf{W}^*_1) - Tr( g_1(\textbf{W}_1, \textbf{x}_i) \textbf{w}_2 \textbf{x}^T_i   \mathbf{V_1}) | $. Note that $ \lVert \textbf{V}_1 - \textbf{W}_1^* \rVert \le \frac{\epsilon}{\kappa}$. From triangle inequality, $ \lVert \textbf{V}_1 \rVert \le \lVert \textbf{V}_1 - \textbf{W}_1^* \rVert + \lVert \textbf{W}_1^* \rVert$.
\begin{align*}
\begin{split}
& | Tr( g_1(\textbf{W}^*_1, \textbf{x}_i) \textbf{w}^*_2 \textbf{x}^T_i \textbf{W}^*_1) - Tr( g_1(\textbf{W}_1, \textbf{x}_i) \textbf{w}_2 \textbf{x}^T_i   \mathbf{V_1}) | 
\\ & \qquad \le | Tr( g_1(\textbf{W}^*_1, \textbf{x}_i) \textbf{w}^*_2 \textbf{x}^T_i \textbf{W}^*_1)| + | Tr( g_1(\textbf{W}_1, \textbf{x}_i) \textbf{w}_2 \textbf{x}^T_i   \mathbf{V_1}) |
\\ & \qquad \le b \cdot \lVert \textbf{w}_2^* \rVert \lVert \textbf{x}_i \rVert \lVert \textbf{W}_1^* \rVert + b \cdot \lVert \textbf{w}_2 \rVert \lVert \textbf{x}_i \rVert \lVert \textbf{V}  \rVert
\\ & \qquad \le 1 \cdot b \cdot W_2 [ \lVert \textbf{W}_1^* \rVert + \lVert \textbf{V}_1 \rVert ]
\\ & \qquad \le b \cdot W_2 [ 2 \cdot \lVert \textbf{W}_1^* \rVert + \lVert \textbf{V}_1 - \textbf{W}_1^* \rVert]
\\ & \qquad \le b \cdot W_2 [ 2 \cdot \lVert \textbf{W}_1^* \rVert + \frac{\epsilon}{\kappa}]
\end{split}
\end{align*}
Now consider the term $| \phi_2 \langle \textbf{w}_2,\phi_{1} \langle \textbf{W}_1, \mathbf{x_i} \rangle \rangle - \phi_2 \langle \textbf{w}^*_2,\phi_{1} \langle \textbf{W}^*_1, \mathbf{x_i} \rangle \rangle |$ appearing in Step 3. We have,
\vspace{-2mm}
\begin{align*}
\vspace{-2mm}
& | \phi_2 \langle \textbf{w}_2,\phi_{1} \langle \textbf{W}_1, \mathbf{x_i} \rangle \rangle - \phi_2 \langle \textbf{w}^*_2,\phi_{1} \langle \textbf{W}^*_1, \mathbf{x_i} \rangle \rangle | \\ &  \qquad \le | \phi_2 \langle \textbf{w}_2,\phi_{1} \langle \textbf{W}_1, \mathbf{x_i} \rangle \rangle | + | \phi_2 \langle \textbf{w}^*_2,\phi_{1} \langle \textbf{W}^*_1, \mathbf{x_i} \rangle \rangle |\\
& \qquad \le b \bigr[ | \langle \textbf{w}_2,\phi_{1} \langle \textbf{W}_1, \mathbf{x_i} \rangle \rangle | + | \langle \textbf{w}^*_2,\phi_{1} \langle \textbf{W}^*_1, \mathbf{x_i} \rangle \rangle | \bigr]\\
& \qquad \le b \bigr[ \| \textbf{w}_2 \| \cdot \| \phi_{1} \langle \textbf{W}_1, \mathbf{x_i} \rangle \| + \| \textbf{w}^*_2 \| \cdot \| \phi_{1} \langle \textbf{W}^*_1, \mathbf{x_i} \rangle \| \bigr]\\
& \qquad \le b \cdot W_2 \cdot \bigr[ \| \phi_{1} \langle \textbf{W}_1, \mathbf{x_i} \rangle \| + \| \phi_{1} \langle \textbf{W}^*_1, \mathbf{x_i} \rangle \| \bigr]\\
& \qquad \le b \cdot W_2 \cdot \bigr[ b \| \langle \textbf{W}_1, \mathbf{x_i} \rangle \| + b \| \langle \textbf{W}^*_1, \mathbf{x_i} \rangle \| \bigr]\\
& \qquad \le b^2 \cdot W_2 \cdot \bigr[\| \textbf{W}_1 \| \| \mathbf{x_i} \| + \| \textbf{W}^*_1 \| \| \mathbf{x_i} \| \bigr]\\
& \qquad \le b^2 W_2 \cdot [ \| \mathbf{W}_1 \| + \| \mathbf{W}_1^*\| ] \\
& \qquad \le 2 \cdot b^2 \cdot W_2 \cdot W_1
\vspace{-2mm}
\end{align*}
\vspace{-2mm}
Using these and the fact that $ |g_2(\textbf{W}_1, \textbf{w}_2, \textbf{x}_i)| \le b $ in Step 3
\begin{align*}
\begin{split} & \ge \frac{2a}{b} \epsilon - 2b \cdot 2b^2W_2W_1 \cdot b W_2[2 \cdot \lVert \textbf{W}_1^* \rVert + \frac{\epsilon}{\kappa}] 
\end{split}\\
\begin{split} & \ge \frac{2a}{b} \epsilon - 4b^4W_2^2W_1 \cdot W_1 - 4b^4W_2^2W_1 \frac{\epsilon}{\kappa}
\end{split} \\
\begin{split} & = \frac{2a}{b} \epsilon - 4b^4W_2^2W_1 \cdot W_1 - 4b^4W_2^2W_1 \epsilon \left( \frac{a}{4b^5W_2^2W_1} - \frac{W_1}{\epsilon} \right)
\end{split} \\
\begin{split} & = \frac{a}{b} \epsilon \ge 0
\end{split}
\vspace{-4mm}
\end{align*}
The idea of the proof is similar to that of previous theorems. The proof uses the fact that the minimum value of the quasigradient of $g$ is $a$.
\end{proof}
\vspace{-8mm}
\subsection{Proof of Theorem \ref{theorem_relu_inner_layer_multi_op}}
\label{appendix_relu_inner_layer_multi_op}
\begin{proof}
Consider $\mathbf{W_1} \in \mathbb{B}(0, W_1)$, $\| \mathbf{W_1} \| \le W_1$, $\| \mathbf{W_2} \| \le W_2$ such that $\hat{err}(\mathbf{W_1}, \mathbf{W_2}) \ge \epsilon $. Let $\mathbf{V_1}$ be a point $\frac{\epsilon}{\kappa}$ close to minima $\mathbf{W}_1$, where $\kappa = \left( \frac{a}{4b^5W_2^2W_1} - \frac{W_1}{\epsilon}\right)^{-1}$\\
Let $\mathbf{W_2} \in \mathbb{R}^{d' \times d^{''}} = [\mathbf{w^2_1}, \mathbf{w^2_2}, \dots \mathbf{ w^2_{d^{''}}} ] $.\\
Note here that,
\begin{align*}
& \nabla_{\mathbf{W_1}} \hat{err}(\mathbf{W_1}, \mathbf{W_2}) \\
& \qquad = \nabla_{\mathbf{W_1}} \frac{1}{m} \sum_{i=1}^{m} \| \mathbf{y_i} - \phi_2 \langle \mathbf{W_2}, \phi_1 \langle \mathbf{W_1}, \mathbf{x_i} \rangle \rangle \|^2 \\
& \qquad = \nabla_{\mathbf{W_1}} \frac{1}{m} \sum_{j=1}^{d''} \sum_{i=1}^{m} \| \mathbf{y_{ij}} - \phi_{2} \langle \mathbf{w_{j}^2}, \phi_1 \langle \mathbf{W_1}, \mathbf{x_i} \rangle \rangle \|^2 \\
& \qquad = \sum_{j=1}^{d''} \nabla_{\mathbf{W_1}} \frac{1}{m} \sum_{i=1}^{m} \| \mathbf{y_{ij}} - \phi_{2} \langle \mathbf{w_{j}^2}, \phi_1 \langle \mathbf{W_1}, \mathbf{x_i} \rangle \rangle \|^2 \\
& \qquad = \sum_{j=1}^{d''} \nabla_{\mathbf{W_1}} \hat{err}(\mathbf{W_1}, \mathbf{w_j^2})
\end{align*}
Now,
\begin{align*}
& \langle \nabla_{\mathbf{W_1}} \hat{err}(\mathbf{W_1}, \mathbf{W_2}), \mathbf{W_1} - \mathbf{V_1} \rangle_F\\
& \qquad = \sum_{j=1}^{d''} \langle \nabla_{\mathbf{W_1}} \hat{err}(\mathbf{W_1}, \mathbf{w_j^2}), \mathbf{W_1} - \mathbf{V_1} \rangle_F
\end{align*}
Observe here that $\| \mathbf W_2 \| \le W_2 \implies \| \mathbf w_j^2 \| \le W_2 \forall j$. Using this and the result from theorem \ref{theorem_relu_inner_layer_single_op} we get that each term $\langle \nabla_{\mathbf{W_1}} \hat{err}(\mathbf{W_1}, \mathbf{w_j^2}), \mathbf{W_1} - \mathbf{V_1} \rangle_F \ge \frac{a}{b} \epsilon$. Hence, we get that:
\begin{align*}
& \langle \nabla_{\mathbf{W_1}} \hat{err}(\mathbf{W_1}, \mathbf{W_2}), \mathbf{W_1} - \mathbf{V_1} \rangle_F \ge \frac{a}{b} \epsilon d'' \ge 0
\end{align*}
\end{proof}
\vspace{-8mm}

\subsection{Proof of Theorem \ref{thm_iglm_ce}}
\label{appendix_iglm_ce}
\begin{proof}
Consider $\mathbf{w} \in \mathbb{B}(0, W)$, $\|\textbf{w}\| \le W$ such that $\hat{err}_i(\textbf{w)}= - (y_i \log (\phi \langle \textbf{w}, \textbf{x}_i \rangle )) + (1 - y_i) (1 - \log (\phi \langle \textbf{w}, \textbf{x}_i \rangle)) \ge \epsilon $ ( $ \implies \hat{err}_m(\textbf{w)}= \frac{1}{m} \sum_{i=1}^m - (y_i \log (\phi \langle \textbf{w}, \textbf{x}_i \rangle ) + (1 - y_i) (1 - \log (\phi \langle \textbf{w}, \textbf{x}_i \rangle))) \ge \epsilon$, where $m$ is the total number of samples). Also let $\textbf{v}$ be a point $\epsilon/\kappa$-close to minima $\textbf{w}^*$ with $\kappa = \frac{\epsilon}{(1 - e^{-\epsilon})^2}$.

Consider the case when $y_i = 1$. In this case $\hat{err}_i(\textbf{w)}= - \log (\phi \langle \textbf{w}, \textbf{x}_i \rangle ) \ge \epsilon $. Using $- \log p \ge \epsilon$. $\implies (1 - p)^2 \ge (1 - e^{-\epsilon})^2$, we get that $ (y_i - \phi \langle \textbf{w}, \textbf{x}_i \rangle)^2 \ge (1 - e^{-\epsilon})^2$. In the other case when $y_i = 0$, $\hat{err}_i(\textbf{w)}= - \log (1- \phi \langle \textbf{w}, \textbf{x}_i \rangle ) \ge \epsilon $. Here using $- \log (1-p) \ge \epsilon$ $\implies (p)^2 \ge (1 - e^{-\epsilon})^2$, we get that $\implies (y_i - \phi \langle \textbf{w}, \textbf{x}_i \rangle)^2 \ge (1 - e^{-\epsilon})^2$. Combining these we get, $ (y_i - \phi \langle \textbf{w}, \textbf{x}_i \rangle)^2 \ge (1 - e^{-\epsilon})^2$ for all $i$. Then:

\begin{align*}
    & \langle \nabla err(\mathbf{w}), \mathbf{w} - \mathbf{v} \rangle \\
    & = \frac{1}{m} \sum_{i=1}^{m} (\phi \langle \mathbf{w}, \mathbf{x}_i \rangle - y_i) \langle \mathbf{x}_i, \mathbf{w} - \mathbf{v} \rangle \tag{Step 1}\\
    & = \frac{1}{m} \sum_{i=1}^{m} (\phi \langle \mathbf{w}, \mathbf{x}_i \rangle - \phi \langle \mathbf{w}^*, \mathbf{x}_i \rangle) ( \langle \mathbf{x}_i, \mathbf{w} - \mathbf{w}^* \rangle + \langle \mathbf{x}_i, \mathbf{w}^* - \mathbf{v} \rangle ) \tag{Step 2}\\
    \begin{split}
    & \ge \frac{1}{m} \sum_{i=1}^{m} (\phi \langle \mathbf{w}, \mathbf{x}_i \rangle - \phi \langle \mathbf{w}^*, \mathbf{x}_i \rangle) \\ & \qquad [( \langle \mathbf{w}, \mathbf{x}_i \rangle - \langle \mathbf{w}^*, \mathbf{x}_i \rangle ) - \| \mathbf{x}_i \| \| \mathbf{w}^* - \mathbf{v} \| ] 
    \end{split} \tag{Step 3}\\
    \begin{split}
    & \ge \frac{1}{m} \sum_{i=1}^{m}  4 (\phi \langle \mathbf{w}, \mathbf{x}_i \rangle - \phi \langle \mathbf{w}^*, \mathbf{x}_i \rangle)^2 \\ & \qquad - | \phi \langle \mathbf{w}, \mathbf{x}_i \rangle - \phi \langle \mathbf{w}^*, \mathbf{x}_i \rangle | \| \mathbf{x}_i \| \| \mathbf{w}^* - \mathbf{v} \|
    \end{split} \tag{Step 4}\\
    & \ge 4 (1 - e^{-\epsilon})^2 - \frac{\epsilon}{\kappa} \tag{Step 5}\\
    & = 3 (1 - e^{-\epsilon})^2 > 0
\end{align*}

Step 2 uses the fact that $y_i = \phi\langle\mathbf{w^{*}},\textbf{x}_{i}\rangle$. In Step 4 we use the fact that sigmoid is $\frac14$ Lipschitz and so $(\phi(z) - \phi(z'))(z-z') \ge 4 (\phi(z) - \phi(z'))^2$. In Step 5 we use $|\phi\langle\mathbf{w},\textbf{x}_{i}\rangle-\phi\langle\mathbf{w^{*}},\textbf{x}_{i}\rangle| \le 1$ and $\|\mathbf{w^{*}}-\mathbf{v}\| \le \frac{\epsilon}{\kappa}$.
\end{proof}
\vspace{-7mm}
\subsection{Proof of Theorem \ref{theorem_outer_layer_ce}}
\label{appendix_outer_layer_ce}
\begin{proof}
Consider $\mathbf{W} \in \mathbb{B}(0, W)$, $\| \textbf{W}\| \le W$ such that for all $i$,\\
$\hat{err}_i(\textbf{W}) = \sum_{j=1}^{d'} - ( y_{ij} \log ( \phi \langle \textbf{w}_j, \textbf{x}_i \rangle)) \ge \epsilon$ \\ ($\implies \hat{err}_i(\textbf{W}) = \frac{1}{m} \sum_{i=1}^{m} \sum_{j=1}^{d'} - ( y_{ij} \log ( \phi \langle \textbf{w}_j, \textbf{x}_i \rangle)) \ge \epsilon$, where $m$ is the total number of samples.) 
Also let $\textbf{V} = [\textbf{v}_1 \enskip \textbf{v}_2 \cdots \enskip \textbf{v}_{d'}]$ be a point $\epsilon/\kappa$-close to minima $\textbf{W}^*$ with $\kappa = \frac{\epsilon d'}{(1 - e^{-\epsilon})^2}$. Let $G(\textbf{W})$ be the subgradient of $\hat{err}_m(\textbf{W})$ and $G(\textbf{w}_j)$ be the subgradient of $\hat{err}_m(\textbf{w}_j )$.\\
Let for $\mathbf{x}_i$, the correct label be $t$, then $y_{it} = 1$ and $y_{ij} = 0$, for any $j \ne t$. The error for this one data-point would be $\sum_{j=1}^{d'} - ( y_{ij} \log ( \phi \langle \textbf{w}_j, \textbf{x}_i \rangle)) = - \log ( \phi \langle \textbf{w}_t, \textbf{x}_i \rangle) \ge \epsilon$. Using $- \log p \ge \epsilon \implies (1 - p)^2 \ge (1 - e^{-\epsilon})^2$, we get that $(y_{it} - \phi \langle \textbf{w}_t, \textbf{x}_i \rangle)^2 \ge (1 - e^{-\epsilon})^2$. 

Note that for any $\mathbf{x}_i$, $\sum_{j=1}^{d'} \phi \langle \textbf{w}_j, \textbf{x}_i \rangle = 1$. Using this we get that $\sum_{j=1, j \ne t}^{d'} \phi \langle \textbf{w}_j, \textbf{x}_i \rangle = 1 - \phi \langle \textbf{w}_t, \textbf{x}_i \rangle \ge 1 - e^{-\epsilon} $ (The inequality follows as $- \log p \ge \epsilon \implies 1 - p \ge 1 - e^{-\epsilon}$). Now using Cauchy-Schwartz inequality, we get that $\sum_{j=1, j \ne t}^{d'} \phi \langle \textbf{w}_j, \textbf{x}_i \rangle^2 \ge (1 - e^{-\epsilon})^2 / (d' - 1)$. Adding this with $(y_{it} - \phi \langle \textbf{w}_t, \textbf{x}_i \rangle)^2 \ge (1 - e^{-\epsilon})^2$ and recollecting that $y_{it} = 1$ and $y_{ij} = 0$, for any $j \ne t$, we get that $\sum_{j=1}^{d'} (y_{ij} - \phi \langle \textbf{w}_j, \textbf{x}_i \rangle)^2 \ge (1 - e^{-\epsilon})^2 \frac{d'}{d' - 1}$
Then:
\begin{align*}
& \langle G(\textbf{W}),\textbf{W}-\textbf{V}\rangle & \notag\\  
& = \sum^{d'}_{j=1}\langle G(\mathbf{w_j}),\mathbf{w_j}-\mathbf{v_j}\rangle_F & \tag{By definition of Frobenius inner product}\\
& = \frac{1}{m} \sum_{i=1}^{m} \sum^{d'}_{j=1}  \left(\phi \langle \mathbf{w_j},\mathbf{x_i} \rangle  - y_{ij}\right) \langle \mathbf{x}_i,(\mathbf{w_j}-\mathbb{\mathbf{v_j}})\rangle & \tag{Step 1}\\
\begin{split} &= \frac{1}{m} \sum_{i=1}^{m} \sum^{d'}_{j=1} \left(\phi \langle \mathbf{w_j},\mathbf{x_i} \rangle  - \phi\langle\mathbf{w^{*}_j},\textbf{x}_{i}\rangle\right) \\ & \qquad [\langle   \textbf{x}_i , \mathbf{w_j}-\mathbf{w^*_j}\rangle  + \langle   \textbf{x}_i , \mathbf{w^*_j}-\mathbf{v_j}\rangle  ]
\end{split} \tag{Step 2}\\
\begin{split}
&  \ge \frac{1}{m} \sum_{i=1}^{m} \sum^{d'}_{j=1} \Bigl[ 2\left(\phi\langle\mathbf{w_j},\textbf{x}_{i}\rangle-\phi\langle\mathbf{w^{*}_j},\textbf{x}_{i}\rangle\right)^{2} \\ & \qquad + \left(\phi\langle\mathbf{w_j},\textbf{x}_{i}\rangle-\phi\langle\mathbf{w^{*}_j} \textbf{x}_{i}\rangle\right)\langle \textbf{x}_{i},\mathbf{w^{*}_j}-\mathbf{v_j}\rangle \Bigr] \end{split} \tag{Step 3}\\
\begin{split}
& \ge \frac{1}{m} \sum_{i=1}^{m} \sum^{d'}_{j=1} \Bigr[2\left(\phi\langle\mathbf{w_j},\textbf{x}_{i}\rangle-\phi\langle\mathbf{w^{*}_j,}\textbf{x}_{i}\rangle\right)^{2} \\ & \qquad -   |\phi\langle\mathbf{w_j},\textbf{x}_{i}\rangle-\phi\langle\mathbf{w^{*}_j,}\textbf{x}_{i}\rangle| \| \textbf{x}_{i}\| \|\mathbf{w^{*}_j}-\mathbf{v_j}\| \Bigl] \end{split} \tag{Step 4} \\
\begin{split}
& \ge \frac{1}{m} \sum_{i=1}^{m} \sum^{d'}_{j=1} \Bigr[ 2\left(\phi\langle\mathbf{w_j},\textbf{x}_{i}\rangle-\phi\langle\mathbf{w^{*}_j},\textbf{x}_{i}\rangle\right)^{2} - \frac{\epsilon}{\kappa}\|\textbf{x}_i\| \Bigl] \end{split} \tag{Step 5}\\
& \ge 2 (1 - e^{-\epsilon})^2 \frac{d'}{d' - 1}  - \frac{d' \epsilon}{\kappa}& \tag{Step 6}\\
& \ge (1 - e^{-\epsilon})^2 \frac{d' + 1}{d' - 1} > 0
\end{align*}

Step 2 uses the fact that $y_{ij} = \phi\langle\mathbf{w^{*}_j},\textbf{x}_{i}\rangle$. In Step 3, we use the fact that softmax is $\frac12$ Lipschitz and so $(\phi(z) - \phi(z'))(z-z') \ge 2 (\phi(z) - \phi(z'))^2$. In Step 5, we use $|\phi\langle\mathbf{w_j},\textbf{x}_{i}\rangle-\phi\langle\mathbf{w^{*}_j,}\textbf{x}_{i}\rangle| \le 1$ and $\|\mathbf{w^{*}_j}-\mathbf{v_j}\| \le \frac{\epsilon}{\kappa}$.
\end{proof}

\section{Conclusion and Future Work}
\label{sec_discussions}
In this work, we presented a novel methodology, Deep AlterNations for Training nEural networks (DANTE), to effectively train neural networks using alternating minimization, thus providing a competitive alternative to standard backpropagation. We formulated the task of training each layer of a neural network (in particular, an autoencoder without loss of generality) as a Strictly Locally Quasi-Convex (SLQC) problem, and leveraged recent results to use Stochastic Normalized Gradient Descent (SNGD) as an effective method to train each layer of the network. While earlier work \cite{sngd} simply identified the SLQC nature of sigmoidal GLMs, we introduced a new generalized ReLU activation, and showed that a multi-output layer satisfies this SLQC property, thus allowing us to expand the applicability of the proposed method to networks with both sigmoid and ReLU family of activation functions. In particular, we extended the definitions of local quasi-convexity in order to prove that a one hidden-layer neural network with generalized ReLU activation is $\left(\epsilon,\frac{2b^{3}W}{a},\mathbf{W^{*}_2}\right)-SLQC$ in $W_2$ (the same result holds for a GLM) and $\left(\epsilon,\left( \frac{a}{4b^5W_2^2W_1} - \frac{W_1}{\epsilon}\right)^{-1},\mathbf{W^{*}_1}\right)-SLQC$ in $W_1$, which  improves the convergence bound for SLQC in the GLM with the generalized ReLU (as compared to a GLM with sigmoid). We also showed how \myalgo can be extended to train multi-layer neural networks. We empirically validated \myalgo with both sigmoidal and ReLU activations on standard datasets as well as in a multi-layer setting, and observed that it provides a competitive alternative to standard backprop-SGD, as evidenced in the experimental results. 
\vspace{-3pt}
\subsection*{Future Work and Extensions} \myalgo can not only be used to train multi-layer neural networks from scratch, but can also be combined with back-prop SGD, which can be used to finetune the network end-to-end periodically. 
Our future work will involve a more careful study of the proposed method for deeper neural networks, as well as in studying convergence guarantees of the proposed alternating minimization strategy.
In this paper, we focused on validating the feasibility of DANTE for MLPs; however the ideas should work for more advanced networks too. In our future work, we plan to study the extensions of DANTE to convolutional layers, LSTMs and other architectural variants.

\section*{Acknowledgements}
This research was partially supported by the Department of Science and Technology, Govt of India MATRICS program, project MTR/2017/001047.



\bibliographystyle{elsarticle-num} 
\bibliography{dante}




\end{document}